\documentclass[acmsmall]{acmart}
\usepackage{graphicx}
\usepackage{amsmath}
\usepackage{tabularx} 
\usepackage{booktabs} 
\usepackage{cleveref} 
\usepackage{hyperref} 
\usepackage{makecell}
\usepackage{multirow}
\usepackage{url}
\usepackage{subcaption}
\usepackage{adjustbox}

\usepackage[pagewise]{lineno}

    \usepackage{color}
    \usepackage{xcolor}
    
    \newcounter{DaveCommentCounter}
    \newcounter{DongCommentCounter}
\AtBeginDocument{%
  }

\setcopyright{acmcopyright}
\copyrightyear{2018}
\acmYear{2018}
\acmDOI{XXXXXXX.XXXXXXX}

\acmJournal{JACM}
\acmVolume{37}
\acmNumber{4}
\acmArticle{111}
\acmMonth{8}

\begin{document}

\title{Short-Term Electricity-Load Forecasting by Deep Learning: A Comprehensive Survey}

\author{Qi Dong}
\orcid{0009-0008-1737-9324}
\email{3230006098@student.must.edu.mo}
\affiliation{
  \institution{School of Computer Science and Engineering, Macau University of Science and Technology}
  \city{Taipa}
  \state{Macau}
  \country{China}
  \postcode{999078}
}

\author{Rubing Huang}
\email{rbhuang@must.edu.mo}
\orcid{0000-0002-1769-6126}
\affiliation{
  \institution{School of Computer Science and Engineering, Macau University of Science and Technology}
  \city{Taipa}
  \state{Macau}
  \country{China}
  \postcode{999078}
}

\author{Chenhui Cui}
\email{3230002105@student.must.edu.mo}
\orcid{0009-0004-8746-316X}
\affiliation{
  \institution{School of Computer Science and Engineering, Macau University of Science and Technology}
  \city{Taipa}
  \state{Macau}
  \country{China}
  \postcode{999078}
}

\author{Dave Towey}
\email{dave.towey@nottingham.edu.cn}
\orcid{0000-0003-0877-4353}
\affiliation{
  \institution{School of Computer Science, University of Nottingham Ningbo China}
  \city{Ningbo}
  \state{Zhejiang}
  \country{China}
  \postcode{315100}
}

\author{Ling Zhou}
\email{lzhou@must.edu.mo}
\orcid{0000-0002-8313-5749}
\affiliation{
  \institution{School of Computer Science and Engineering, Macau University of Science and Technology}
  \city{Taipa}
  \state{Macau}
  \country{China}
  \postcode{999078}
}

\author{Jinyu Tian}
\email{jytian@must.edu.mo}
\orcid{0000-0002-2449-5277}
\affiliation{
  \institution{School of Computer Science and Engineering, Macau University of Science and Technology}
  \city{Taipa}
  \state{Macau}
  \country{China}
  \postcode{999078}
}

\author{Jianzhou Wang}
\email{jzwang@must.edu.mo}
\orcid{0000-0001-9078-7617}
\affiliation{
  \institution{Department of Engineering Science, Macau University of Science and Technology}
  \city{Taipa}
  \state{Macau}
  \country{China}
  \postcode{999078}
}

\renewcommand{\shortauthors}{Dong et al.}


\begin{abstract}
    Short-Term Electricity-Load Forecasting (STELF) refers to the prediction of the immediate demand (in the next few hours to several days) for the power system.
    Various external factors, such as weather changes and the emergence of new electricity consumption scenarios, can impact electricity demand, causing load data to fluctuate and become non-linear, which increases the complexity and difficulty of STELF.
    In the past decade, deep learning has been applied to STELF, modeling and predicting electricity demand with high accuracy, and contributing significantly to the development of STELF. 
    This paper provides a comprehensive survey on deep-learning-based STELF over the past ten years.
    It examines the entire forecasting process, including data pre-processing, feature extraction, deep-learning modeling and optimization, and results evaluation. 
    This paper also identifies some research challenges and potential research directions to be further investigated in future work.
\end{abstract}

\ccsdesc[500]{Computing methodologies~Machine learning algorithms}
\ccsdesc[500]{Mathematics of computing~Probability and statistics}
\ccsdesc[500]{Applied computing~Forecasting}


\keywords{electricity, load, deep learning, short term}
\received{}
\received[revised]{}
\received[accepted]{}

\maketitle

\section{Introduction}
Electricity-Load Forecasting (ELF) aims to meet power systems' daily operational, management, and planning needs.
This can provide essential guidance and reference points for system operators and planners.
The absence of large-scale energy storage technologies means that power systems must ensure a constant power supply to meet current demands~\cite{pelka2020pattern}.
This means that ELF has become an essential component in the planning, scheduling, and operational management of power systems.
Short-Term Electricity-Load Forecasting (STELF) uses historical load data to predict future loads, over a period ranging from several hours to a few days.
STELF is a time-series forecasting task primarily used for the short-term scheduling of smart grids (including equipment maintenance, load distribution, and unit startup and shutdown), and for determining electricity prices~\cite{wu2021online}. 
Data from a British electricity company in 1984 showed that a 1\% reduction in forecasting error could save £10 million annually~\cite{bunn1985comparative}. 
Increasing the STELF accuracy could improve planning and scheduling, and reduce operational costs for power systems~\cite{kouhi2014new}. 
This practical impact of STELF has led to an increasing amount of attention from researchers.
Our review of the literature from the past decade highlights that most load forecasting articles have focused on STELF~\cite{ ghofrani2015hybrid, hu2017short, raza2017intelligent, hoori2019electric, choi2018short,guo2020short}.

Many factors can impact the electricity load, including climate, weather, economic conditions, seasonality, and electricity prices. 
Furthermore, advances in smart-grid technologies, and the widespread adoption of smart meters and other sensors have significantly increased both the complexity and the volume of electricity-load data.
The data exhibits strong non-linearity, randomness, volatility, and complexity.  
This is a challenge for STELF.
An accurate, robust, and fast STELF model is essential for the reliable daily operations of power systems~\cite{wang2019bi}, and the development of efficient forecasting models has become an important STELF research goal \cite{dou2018application}.

ELF has been extensively studied since the 1970s, with various methods having been proposed~\cite{fahiman2019robust}. 
STELF methods can be broadly categorized into three types: statistical methods, machine learning methods, and deep learning methods~\cite{dudek2016neural}. 
Statistical methods perform well when dealing with linear relationships, but are often inadequate for handling the nonlinear patterns commonly found in electricity-load data. 
Machine learning methods, such as support vector machines and decision trees, typically perform well with simple or moderately complex data patterns. 
Deep learning methods can capture and model complex nonlinear relationships through their multi-layered structures, which is particularly important for predicting dynamic changes in electricity loads. 
In summary, while traditional statistical and machine learning methods have their strengths, deep learning techniques are more suited to the dynamic and nonlinear characteristics of electricity-load data.

Deep learning approaches are among the most revolutionary breakthroughs in the fields of computer science and artificial intelligence in recent years~\cite{lecun2015deep}. 
The concept of deep learning builds on earlier work on Artificial Neural Networks (ANNs)~\cite{hinton2006fast}, which
were a type of shallow learning model ~\cite{schmidhuber2015deep}, consisting of an input layer, a hidden layer, and an output layer~\cite{almalaq2017review}. 
ANNs were used in early STELF studies \cite{hayati2007artificial,park1991electric}.
Deep Neural Networks (DNNs), a type of ANN with multiple hidden layers, can also be used for STELF~\cite{hosein2017load,lai2020load,alipour2020novel}. 
The multiple hidden layers in DNNs enable a complex computational framework that uses features as inputs to represent different levels of data abstraction.
Through a cascading network structure, each layer in the DNN is capable of extracting and recognizing different features of the data, forming a hierarchy from basic to advanced features, thereby significantly enhancing both the model's flexibility and its ability to handle complex issues. 
Recurrent Neural Networks (RNNs)~\cite{rumelhart1986learning} are DNNs that were designed specifically to process sequential data, making them highly suitable for time-series prediction tasks. 
Although RNNs are theoretically ideal for time-series data, they may face challenges, like vanishing or exploding gradients in practical applications~\cite{tang2019ensemble}.
Long Short-Term Memory (LSTM)~\cite{hochreiter1997long} and Gated Recurrent Units (GRUs)~\cite{wang2021bottom} have addressed gradient vanishing, making them more effective for practical applications. 
These advanced technologies, capable of handling large-scale, high-dimensional, and nonlinear data, provide more accurate and flexible solutions for STELF, making deep learning the preferred technique for STELF~\cite{ahajjam2022experimental,das2020occupant,li2023short}. 

Previous STELF reviews have compiled and examined the research achievements from various perspectives, examining all types of models, or concentrating on certain steps of the forecasting process~\cite{akhtar2023short,hou2022review,al2020comprehensive}.
Akhtar et al.~\cite{akhtar2023short}, for example, reviewed various STELF models (including time series and regression models), rather than focusing only on Artificial Intelligence (AI) models.
Although Hou et al.~\cite{hou2022review} reviewed load forecasting based on AI models, focusing on data processing and prediction models.
Al Mamun et al.~\cite{al2020comprehensive} reviewed load forecasting, but only focused on hybrid models based on machine learning algorithms.
To date, there has been no comprehensive and exhaustive review based on deep learning for STELF that covers the entire forecasting process.
This paper aims to fill this gap in the literature.

This article explores the application of deep learning in STELF, providing a comprehensive review of current relevant research.
The entire STELF process is examined through a comprehensive review of the literature from 2014 to 2023.
The paper is guided by eight research questions (RQs), each of which addresses a key aspect of the STELF process.
This survey paper addresses the following key points:
(1) a summary and analysis of the literature search results; 
(2) a classification and description of electricity load datasets; 
(3) an introduction to STELF data preprocessing methods; 
(4) an analysis of feature extraction; 
(5) a description, classification, and summary of STELF models based on deep learning; 
(6) a review of the optimization process; 
(7) a summary of evaluation metrics; and 
(8) a discussion of the challenges and trends for the future of STELF.

The rest of this paper is organized as follows: Section~\ref{SEC:Background} introduces some background information about the formal description of STELF and the basic deep learning models. 
Section~\ref{SEC:Methodology} explains the methodology of this review, including the eight RQs related to STELF, the literature retrieval methods, and the statistical results of the retrieval. 
Sections~\ref{SEC:Answer to RQ1} to \ref{SEC:Answer to RQ8} answer the eight RQs from Section~\ref{SEC:Methodology}, respectively.
Finally, Section~\ref{SEC:Conclusion} concludes the paper.

\section{Background
\label{SEC:Background}}

In this section, we introduce the task definition of STELF and the basic deep learning methods.
The primary objective is to quickly familiarize readers with STELF and provide them with an initial understanding of deep learning methods.

\subsection{STELF Task Definition}

STELF aims to predict the electricity load over a future period ranging from a few hours to several days.
The model's input consists of historical load data and some influencing factors, with the task being to learn a set of mapping functions from input to output.
If $y_t$ represents the load demand at time \( t \), the STELF goal is to predict the load demand within the next \( h \) hours, denoted as \( \hat{y}_{t+h} \). 
The prediction model can generally be expressed as:
\begin{equation}
    \hat{y}_{t+h} = f(y_{t}, y_{t-1}, \ldots, y_{t-n+1}, X_t),
\end{equation}
where 
\(\hat{y}_{t+h}\) is the predicted load at time \(t + h\); 
\(f\) is the prediction model (which can be statistical, a machine learning model, or a deep learning model); 
\(y_{t}, y_{t-1}, \ldots, y_{t-n+1}\) are the historical load data for the previous \(n\) time periods; and
\(X_t\) are the external variables at time \(t\) (such as weather data, calendar information, etc.).

\subsection{Basic Deep-Learning Models}
In this section, we introduce traditional deep learning models and explore their innovations and variations. 
We also provide an introduction to the basic definitions and structures of these models, establishing a basis for a more detailed exploration of the application of deep learning methods in STELF.

\subsubsection{Deep Neural Networks (DNNs)}
DNNs are a complex and highly non-linear method for representation learning, typically consisting of an input layer, multiple hidden layers, and an output layer~\cite{din2017short}. 
Each neuron in the hidden layers functions as a unit that performs mapping within a multi-dimensional data space. 
Together, these neurons extract complex abstract features and patterns from the input data. 
Both the width and the depth of DNN networks can be modified~\cite{hossen2017short}.  
Shallow neural networks, which have only a single hidden layer, offer only the number of neurons as an adjustable parameter.
The strength of DNNs lies not only in their deep structure but also in their non-linear activation functions.
Non-linear activation functions, such as ReLU, Sigmoid, or Tanh, enable the network to capture non-linear relationships and complex patterns in the input data. 
Considering the non-linear nature of ELF load curves (which are influenced by various external factors), the use of DNN as a predictive model is well justified~\cite{hossen2018residential}.

The Deep Belief Network (DBN) is a DNN variant that uses a layered unsupervised learning method for initial weight pre-training~\cite{hinton2006fast}. 
This layer-by-layer unsupervised training process ensures that each layer effectively captures the features of the preceding layer, allowing the most fundamental features to be extracted from the training set.
Typically, a DBN is composed of multiple stacked Restricted Boltzmann Machines (RBMs). 
An RBM is a type of ANN consisting of a visible layer and a hidden layer, where there are no connections between nodes within the same layer, but nodes between layers are fully connected~\cite{hafeez2020electric}. 
Stacked RBMs are used for model pre-training and unsupervised learning, with the top layer fine-tuned using a backpropagation neural network. 
Fundamentally, an RBM learns a feature representation of a probability distribution over the original input data while also extracting feature information~\cite{kong2019improved}.

\subsubsection{Recurrent Neural Networks (RNNs)}
RNNs are particularly effective at processing sequential data, such as time-series data~\cite{gurses2022introducing}. 
The RNN internal loops allow for the continuous transmission of information, which is the use of previous information to influence the current output, a capability also known as the memory function~\cite{kong2017short}.
However, RNNs often encounter vanishing or exploding gradient issues when processing long sequences, which hinders their ability to learn long-term dependencies~\cite{bashir2022short}. 
LSTM and GRU were designed to overcome the RNN gradient issues when handling long sequences.
They use different memory mechanisms to retain input information over extended periods~\cite{tayab2020short,li2022short}.
Faced with the distinct temporal sequences and cyclic patterns in ELF, LSTM, and GRU can utilize historical information for load forecasting and avoid gradient-related issues.

LSTM is a special kind of RNN capable of learning long-term dependencies, specifically designed to address the issue of vanishing gradients. 
The key to LSTM is its internal structure, the memory cell, which includes four main components: 
an input gate, a forget gate, an output gate, and a cell state that can maintain information over time~\cite{haque2022short}. 

GRU is an improved model based on LSTM, but with a simpler structure and shorter training times, which helps it to better capture long-term dependencies within sequential data.
GRU integrates the forget and input gates of LSTM into a single update gate, combining the cell state and hidden state. 
Compared to LSTM, GRU has fewer parameters, due to having one less gating unit.
This significantly improves the computational efficiency. 

Both LSTM and GRU improve the information flow control through their gating mechanisms. 
While LSTM provides precise control mechanisms, GRU improves the computational efficiency of these controls by simplifying them.
A choice between these two models typically depends on the specific demands of the task, the nature of the data, and the computational resources available.

\subsubsection{Convolutional Neural Networks (CNNs)}
In recent years, CNNs have become one of the most popular and widely utilized deep-learning models~\cite{lu2019hybrid}.
Although initially designed for processing image data, CNNs are also very effective at handling time-series data.
For such data, One-Dimensional (1-D) convolutional layers can capture local patterns and features through a sliding window mechanism. 
Additionally, CNNs can handle data with grid-like topologies, allowing sequence data to be converted into graph-structured data for processing.
A CNN model consists of four main parts~\cite{li2020deep,kim2019recurrent}: 
(i) convolutional layers (which create feature maps from the input data); 
(ii) pooling layers (which reduce the dimensionality of the convolutional features); 
(iii) flattening layers (which reshape the data into a column vector); and
(iv) fully connected layers (which link the features extracted by the convolutional and pooling layers to other layers).

\section{Methodology
\label{SEC:Methodology}}

In this section, we present the eight RQs related to STELF that guided our study.
This section also provides a detailed introduction to the literature retrieval methods, and the filtering methods used to screen search results. 
The research methodology of this paper was guided by previous work~\cite{huang2019survey,zhang2021review}, and represents a systematic, comprehensive, and rationality approach. 
\textbf{For ease of description, we use the term ``\textit{STELF}'' to represent the term ``\textit{deep-learning-based STELF}" in this paper, unless explicitly stated.}

\subsection{Research Questions}
This paper provides a comprehensive review of the application of deep learning in STELF. 
It examines the entire STELF process, structured around the following RQs:
\begin{itemize}
    \item  RQ1: What is the distribution and analysis of the literature search results?
    \item  RQ2: What are the electricity load datasets?
    \item  RQ3: How can a dataset be preprocessed for STELF?
    \item  RQ4: What are the methods for feature extraction?
    \item  RQ5: What are the deep-learning-based modeling methods for STELF?
    \item  RQ6: How can the training processes be optimized?
    \item  RQ7: How have the STELF research results been evaluated?
    \item  RQ8: What are the challenges and the future development trends of STELF?
\end{itemize}

\begin{figure}[ht]
  \centering
  \includegraphics[width=\textwidth]{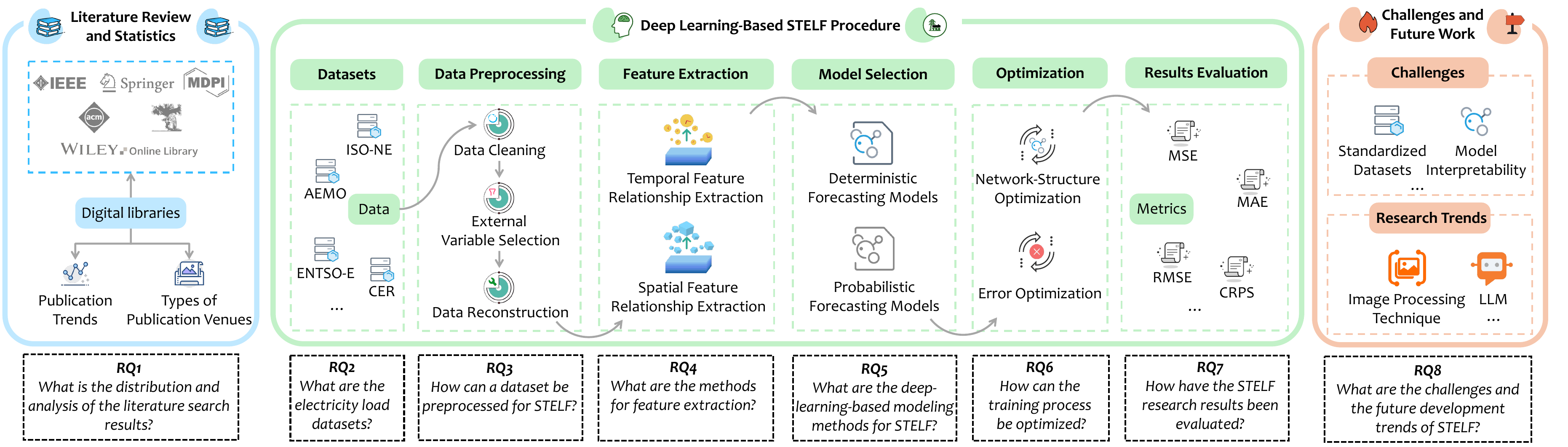 }
  \caption{The structure of this survey paper.}
  \Description{This figure provides an overview of the survey structure.} 
  \label{fig:workflow}
\end{figure}

RQ1 explores the distribution of literature related to the use of deep learning for STELF over the past decade, leading to a detailed analysis of this data.  
RQ2 leads to an overview of electricity load datasets. 
An answer to RQ3 includes the steps and methods used in data preprocessing. 
RQ4 leads to a discussion of the deep-learning feature-extraction techniques employed in the prediction process. 
Answering RQ5 classifies, describes, and analyzes the current state of deep-learning models in STELF.
RQ6 explores the methods for optimizing the model-training process.
The answer to RQ7 is an organized summary of the evaluation methods used for the forecasting results.
Finally, the answer to RQ8 lists the challenges and future development trends of STELF. 
In the following sections, we provide detailed responses to each RQ, as illustrated in Fig.~\ref{fig:workflow}.

\subsection{Literature Search} 
The literature search method employed in this paper follows the approach used by Huang et al.~\cite{huang2019survey}, involving the following mainstream databases to ensure a comprehensive data collection:
\begin{itemize}
    \item  ACM Digital Library;
    \item  Elsevier Science Direct;
    \item  IEEE Xplore Digital Library;
    \item  Springer Online Library;
    \item  Wiley Online Library;
    \item  MDPI.
\end{itemize}
The search time range was set to the period 2014 to 2023.
An initial attempt using certain keywords for the search~\cite{zhang2021review} in each database revealed that the ELF titles were not uniformly represented.
Some papers, for example, did not include words such as ``electricity'' or ``power'' in their titles, even though their actual contents were related to ELF.
Furthermore, not all STELF papers had the phrase ``short-term'' in their titles or keywords list.
In addition, because of the diverse terminology used in deep-learning methods, the inclusion of such terminology in the keyword search could result in the omission of relevant papers.
To broaden the scope, and avoid missing publications from a particular category, ``electricity'', ``power'', ``short-term", and ``deep learning'' were not included in the keywords. 
As a result, the final set of keywords used in the search was limited to four phrases: 
``load forecasting'', ``load forecast'', ``load prediction'', and ``load predicting''. 

\subsection{Literature Selection and Statistics}
A total of 2,823 papers were retrieved from the six databases, according to the search parameters. 
These papers were further filtered according to the following selection criteria:
\begin{enumerate}
    \item Not written in English.
    \item Not discussing ELF.
    \item Not using deep-learning methods.
    \item Not focused on ``short-term'' forecasting research.
    \item The paper was a review article.
\end{enumerate}
Based on these criteria, 628 papers were selected from the initial 2,823. 
Additionally, the references of these papers were examined according to the snowballing approach~\cite{huang2019survey}, yielding an additional 22 papers. 
In total, 650 papers were included in the preliminary review and statistical analysis.
The details of this search and filtering are shown in Table~\ref{tab:literature_database_summary}.

\begin{table}[ht]
    \centering
    \scriptsize
    \renewcommand{\arraystretch}{1.5}
    \caption{Literature search and selection results}
    \begin{tabularx}{\textwidth}{>{\centering\arraybackslash}X|c|c|c|c}
        \hline
        \thead{\textbf{Digital library } }& \thead{\textbf{No. of studies from} \\ \textbf{the search results} } & \thead{\textbf{No. of studies} \\ \textbf{after filtering}} & \thead{\textbf{No. of studies} \\ \textbf{by snowballing}} & \thead{\textbf{Total of filtering} \\ \textbf{and snowballing}} \\
        \hline
        ACM Digital Library & 80 & 14 & 0 & 14 \\
        \hline
        Elsevier Science Direct & 889 & 126 & 0 & 126 \\
        \hline
        IEEE Xplore Digital Library & 846 & 332 & 22 & 354 \\
        \hline
        Springer Online Library & 440 & 20 & 0 & 20 \\
        \hline
        Wiley Online Library & 112 & 25 & 0 & 25 \\
        \hline
        MDPI & 456 & 111 & 0 & 111 \\
        \hline
        \textbf{\textit{Total}} & 2823 & 628 & 22 & 650 \\
        \hline
    \end{tabularx}  
    \label{tab:literature_database_summary}
\end{table}

We manually processed each of the 650 papers. 
This process involved an initial checking of all papers, followed by the extraction and recording of key information (including the deep-learning models, datasets, data-preprocessing methods, prediction intervals, model block diagrams, and evaluation metrics).
Finally, the content was structured according to Fig.~\ref{fig:workflow}.
Although the statistics and analysis results are based on all 650 publications, not all 650 are cited in this paper.
Instead, we selectively cited papers with similar content based on the extracted information. 
Ultimately, this paper thoroughly reviews and cites approximately 200 articles.

\section{Answer to RQ1: 
The Distribution and Analysis of the Search Results
\label{SEC:Answer to RQ1}}

This section provides the answers to RQ1. 
Our approach involved an analysis of the publication year trends and their distribution across various literature sources, providing a framework for understanding the evolution and scope of the field.

\subsection{Publication Trends}
\begin{figure}[ht]
    \centering
    \begin{subfigure}[t]{0.48\textwidth}
        \centering
        \includegraphics[width=\textwidth]{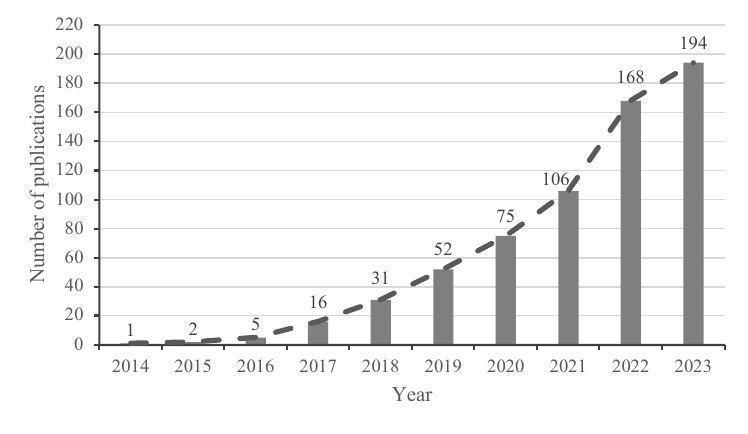}
        \caption{Number of publications per year.}
        \label{fig:number_of_publications_per_year}
    \end{subfigure}
    \hfill
    \begin{subfigure}[t]{0.48\textwidth}
        \centering
        \includegraphics[width=\textwidth]{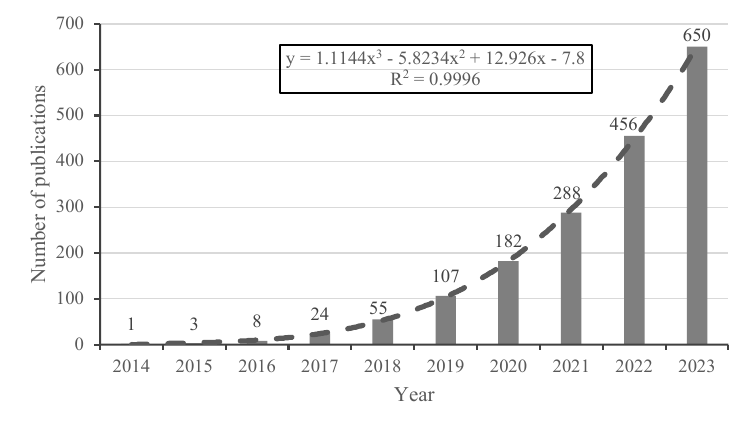}
        \caption{Cumulative number of publications per year.}
        \label{fig:cumulative_number_of_publications_per_year}
    \end{subfigure}
    \caption{STELF papers published between January 1, 2014, and December 31, 2023.}
    \Description{This figure depicts the distribution of publication years and the cumulative distribution over the years.}
    \label{fig:publication_trends}
\end{figure}

We gathered the publication year data of the 650 papers (shown in Fig.~\ref{fig:publication_trends}) to show the trends of STELF papers between 2014 and 2023.
Fig.~\ref{fig:number_of_publications_per_year} shows the number of publications per year, and Fig.~\ref{fig:cumulative_number_of_publications_per_year} shows the cumulative number of publications. 

Fig.~\ref{fig:number_of_publications_per_year} shows that there were fewer than 10 publications per year during the first three years (2014 to 2016). 
There has been a rapid growth since 2017, with the number of publications exceeding 100 per year by 2021. 
Furthermore, an examination of the cumulative numbers of publications (Fig.~\ref{fig:cumulative_number_of_publications_per_year}) reveals an exponential growth in the research output for STELF.
Fig.~\ref{fig:cumulative_number_of_publications_per_year} also shows a linear function with an exceptionally high coefficient of determination ($R^2 = 0.9996$).
The trend shown in Fig.~\ref{fig:number_of_publications_per_year} is directly related to the rapid development of deep-learning technologies in recent years.
It also highlights the significance of research in this field.

\subsection{Types of Publication Venues}

\begin{figure}[ht]
    \centering
    \begin{subfigure}[t]{0.48\textwidth}
        \centering
        \includegraphics[width=\textwidth]{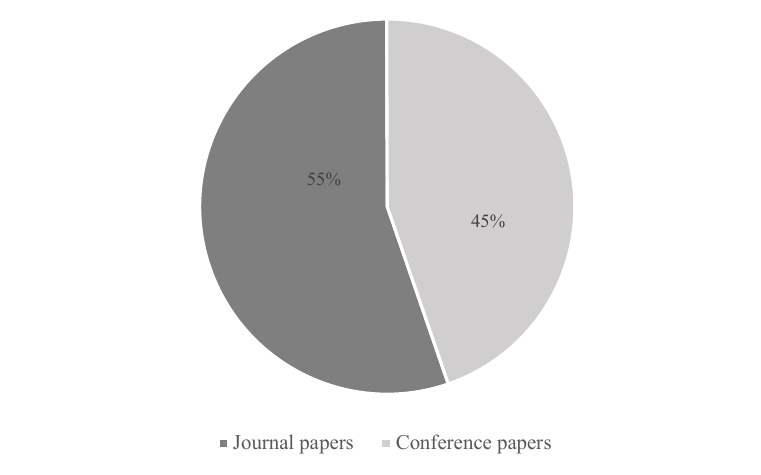}
        \caption{Proportion of journals and conferences.}
        \label{fig:Proportion of journals and conferences.}
    \end{subfigure}
    \hfill
    \begin{subfigure}[t]{0.48\textwidth}
        \centering
        \includegraphics[width=\textwidth]{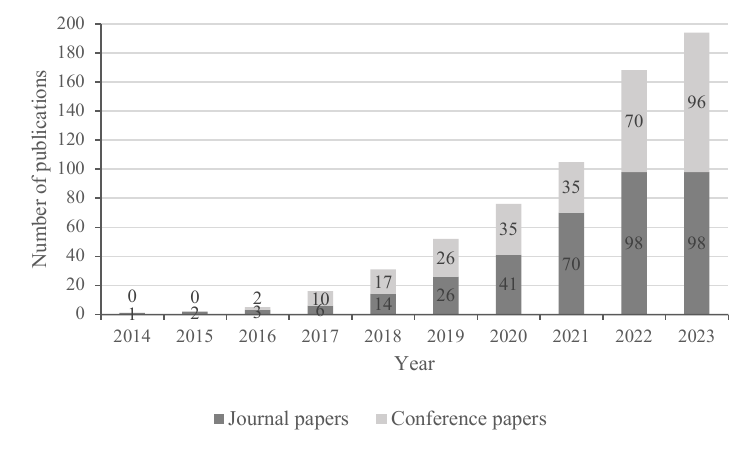}
        \caption{Venue distribution per year.}
        \label{fig:Venues distribution per year}
    \end{subfigure}
    \caption{Venue distribution of surveyed papers.}
    \Description{This figure provides a description of the classification and distribution for publication.}
    \label{fig:Venue distribution for papers.}
\end{figure}

The papers were sourced from multiple journals and conferences, with their proportion and distribution plotted in Fig.~\ref{fig:Venue distribution for papers.}.
Fig.~\ref{fig:Proportion of journals and conferences.} shows that the number of publications published in journals (55\%)  exceeds those in conferences (45\%).  
Fig.~\ref{fig:Venues distribution per year} shows that, apart from 2017, the number of journal publications consistently outpaces the number of conference publications.
Fig.~\ref{fig:Venues distribution per year} also shows a trend, for both journals and conferences, of increasing publication volume.

\section{Answer to RQ2: The Electricity Load Datasets
\label{SEC:Answer to RQ2}}

This section provides the answers to RQ2, which discusses the classification of datasets and examines some common public datasets.
When selecting electricity load datasets, researchers need to choose the appropriate type of dataset based on the specific requirements of the forecasting scenario, which may include loads for residential households, commercial buildings, industry, and entire cities' system-level load.

Electricity load datasets play a crucial role in STELF. 
ELF typically relies on historical data to predict future electricity demand over specific periods. 
With the advancement of deep learning methods, these models require substantial amounts of data to train for accurate predictions. 
Electricity load datasets provide comprehensive historical electricity usage information, including load changes during different periods, consumer usage patterns, and the impact of seasonal and weather factors on electricity demand~\cite{zhang2022similar}. 
This information is vital for a deep understanding and accurate prediction of electricity demand patterns.

\subsection{Classification of Datasets}
Electricity load datasets can be broadly categorized based on their accessibility as either public or not. 
Public datasets are typically available online, and are usually provided by government agencies, power market operators, or research institutions~\cite{giacomazzi2023short,li2020effective,xia2023combined,khan2022efficient,atef2020assessment}. 
Non-public datasets, in contrast, are usually not available due to considerations such as protecting competitive advantage, ensuring national security, complying with legal regulations, or protecting personal privacy.

\subsection{Common Public Datasets}
The review of the 650 papers revealed a wide variety of datasets. 
Some papers (such as~\cite{deepanraj2022intelligent,sun2018short,chen2021load}) did not mention the name or source of the dataset, while others (such as~\cite{wang2020short,arastehfar2022short,zhang2023general}) used multiple datasets, making it difficult to compile statistics.
Table \ref{table:datasets} lists some of the most frequently used public datasets, which are:
\begin{itemize}
    \item  
    Independent System Operator of New England (ISO-NE)\footnote{\url{https://www.iso-ne.com/}.}: 
    The dataset includes hourly electrical load, temperature, day type, and other information for the New England area of North America, from March 2003 to December 2014.
    
    \item  
    Australian Energy Market Operator (AEMO)\footnote{\url{https://www.aemo.com.au/energy-systems/electricity/national-electricity-market-nem/data-nem/aggregated-data}.}: 
    The dataset contains 30-minute time-series data on electrical loads for five regions in Australia (South Australia, Queensland, New South Wales, Western Australia, and Victoria).
    
    \item 
    Global Energy Forecasting Competition (GEFCom)\footnote{\url{https://www.kaggle.com/c/global-energy-forecasting-competition-2012-load-forecasting/data}.}: 
    Each competition provides different datasets that cover historical data and related influencing factors for various regions and periods. The design of the datasets reflects the real-world conditions of energy markets and system operations.
    
    \item  
    University of California, Irvine (UCI) Machine Learning Repository (MLR)\footnote{\url{https://archive.ics.uci.edu/ml/datasets}.}: 
    This dataset is a widely used public database covering simple datasets to complex multivariate time-series datasets. 
    The household load data uses a sampling frequency of one minute.
    
    \item  
    European Network of Transmission System Operators for Electricity (ENTSO-E)\footnote{\url{https://open-power-system-data.org/data-sources##1_European_load_data}.}: 
    This dataset is from an organization of electricity transmission system operators covering 35 European countries. 
    The dataset includes real-world hourly electricity load time series from across Europe.
    
    \item  PJM Interconnection (PJM)\footnote{\url{https://www.kaggle.com/datasets/robikscube/hourly-energy-consumption}.}:
    This dataset is from a regional transmission organization in the United States responsible for operating the Eastern Interconnection grid. 
    PJM transmits electricity to 14 regions in the United States, with data recorded at hourly MW intervals.
    
    \item  
    Commission for Energy Regulation (CER)\footnote{\url{https://www.cru.ie/}.}: 
    This dataset records the half-hourly load data of residential households and small to medium-sized enterprises in Ireland.
    
    \item 
    2016 China Electrical Mathematical Modeling Competition\footnote{\url{https://github.com/huberyCC/Load-datasets}.}: 
    This dataset originates from the China Electrical Engineering Mathematical Modeling Competition and includes electrical load data and weather data from 2009 to 2015.
\end{itemize}

All the datasets listed in Table~\ref{table:datasets} have been used at least 10 times. 
The frequent use of ISO-NE and AEMO especially, both of which exceed 30 times, highlights their significant role and the high level of activity these datasets sustain in ELF research.
The sampling frequency of the datasets varies from every minute to every hour, highlighting the diversity of real-time granularity, which supports a wide range of research and practical applications.
Many important power datasets are covered, and distributed across multiple regions (including North America, Australia, Europe, and China).
 
The majority of the public datasets can be accessed online through platforms such as Kaggle, official websites, and specialized data repositories.
This facilitates their access and use by researchers worldwide. 
Open access and faster updates are two major trends in the development of power datasets.

\begin{table}[ht]
\scriptsize
\caption{Commonly Used Electricity Datasets}
\renewcommand{\arraystretch}{1.5}
\centering
\begin{tabular}{>{\centering\arraybackslash}m{6.5cm}| c | c }
    \hline
    \textbf{Dataset}& \textbf{Usage Frequency} & \textbf{Sampling Frequency} \\ 
    \hline
    Independent System Operator of New England (ISO-NE) & 47 & 1 hour  \\ 
    \hline
    Australian Energy Market Operator (AEMO) & 35 & 30 minutes  \\ 
    \hline
    Global Energy Forecasting Competition (GEFCom) & 23 & 1 hour  \\ 
    \hline
    UCI Machine Learning Repository (UCI) & 23 & 1  minute   \\ 
    \hline
    European Network of Transmission System Operators for Electricity (ENTSO-E) & 15 & 1 hour   \\ 
    \hline
    PJM Interconnection (PJM) & 11 & 1 hour  \\ 
    \hline
    Commission for Energy Regulation (CER) & 10 & 30 minutes   \\ 
    \hline
    2016 China Electrical Mathematical Modeling Competition & 10 & 15 minutes  \\ 
    \hline
\end{tabular}
\label{table:datasets}
\end{table}

\section{Answer to RQ3: STELF Dataset Preprocessing
\label{SEC:Answer to RQ3}}

In this section, we address RQ3, introducing the main data-preprocessing methods (including data cleaning, selection of external variables, and data reconstruction).

Data preprocessing is a crucial step that directly impacts the accuracy and reliability of the prediction models. 
Raw electricity-load data frequently contains noise, anomalies, and inconsistent records. 
There are also often missing values. 
Unaddressed, these issues can significantly disrupt the learning process of models, leading to inaccurate or ineffective predictions~\cite{meng2022short}. 
Data preprocessing uses a series of methods (such as filling in missing values, and anomaly detection, correction, and normalization) to enhance the data quality~\cite{subbiah2022deep}. 
The data also often exhibits strong temporal characteristics and seasonal variability.
Appropriate preprocessing can help models capture these complex patterns, enhancing their understanding and responsiveness to temporal dynamics~\cite{lv2021vmd}.
This section examines the various data-preprocessing steps and methods. 
It should be noted that published papers generally only mention one or several data preprocessing steps, not all.
For example, Dong et al.~\cite{dong2021short} only discussed data normalization; 
Gao et al.~\cite{gao2023adaptive} focused only on handling missing data; and 
Huang et al.~\cite{dogra2023consumers} introduced data standardization and data reconstruction.
Specific preprocessing measures are usually applied based on the design requirements of the model.

\subsection{Data Cleaning}
The data-cleaning process involves handling missing values, outliers, erroneous records, duplicate data, and performing standardization.

Various data-imputation techniques can be used to fill in the gaps of missing data~\cite{rafati2020efficient,gan2017enhancing}. 
Some simple approaches include using the value from a previous time point or calculating the average of data from before and after the missing value \cite{li2019power,chen2023research,hua2023ensemble}.
Other methods include data clustering~\cite{li2022electric}, and using values from the same time on adjacent dates~\cite{zhang2022electricity}.

Outlier detection is often addressed using the three-sigma method~\cite{khan2022efficient,chandola2009anomaly}.
Sharma et al.~\cite{sharma2022novel} also used the interquartile range for this purpose, while Qin et al.~\cite{qin2022multi} used box plots.
Typically, when encountering outliers, erroneous records, or duplicated data, the main remedial strategies involve deletion or replacement (using established methods for handling missing values).

Data standardization relates to eliminating scale differences in the original data, allowing for comparison and calculation on the same scale.
The use of raw data for analysis may lead to biases towards features with larger numerical ranges~\cite{tan2022multi}.
Two methods for data standardization are Min-Max Scaling and Z-Score Normalization~\cite{wang2023shortb,huang2021decomposition}.

Min-Max Scaling adjusts the data to fit within a specified range, commonly between 0 and 1. 
The formula for Min-Max Scaling is:
\begin{equation}
    x_{\text{norm}} = \frac{x - \min(x)}{\max(x) - \min(x)},
\end{equation}
where \( x \) is the original data value; 
\(\min(x)\) and \(\max(x)\) are the minimum and maximum values of the original data, respectively; and 
\( x_{\text{norm}} \) is the scaled data.

Z-Score Normalization shifts the data's mean to 0 and standard deviation to 1, making it suitable for data that requires outlier mitigation and handling of skewed distributions. 
The formula for Z-Score Standardization is:
\begin{equation}
    z = \frac{(x - \mu)}{\sigma},
\end{equation}
where \( x \) is the original data value; 
\( \mu \) is the mean of the data; and 
\( \sigma \) is the standard deviation.

\subsection{External Variable Selection}
The accuracy of STELF is determined not only by the operational conditions within the power system, but also by carefully considering a series of important external variables~\cite{wang2023short}. 
The next task after completing the data cleaning is to identify the external variables that significantly impact the forecasting results~\cite{cai2020short}, such as the variability of climatic conditions, periodic fluctuations in temperature, holiday status, and dynamic changes in industrial activities. 
Appropriate consideration of these external factors can lead to a more comprehensive load forecasting model.
This usually involves correlation analysis and variable-importance evaluation.

Table~\ref{table:selection_methods} lists six commonly used methods for external variable selection, and some of the papers that use them.
The Pearson Correlation Coefficient, for example, calculates the Pearson correlation between two continuous variables, with values ranging from -1 to 1. 
By determining a threshold for the correlation coefficient, only variables whose correlation with the target variable exceeds this threshold are considered to be correlated.

Zheng et al.~\cite{zheng2018short} used the Least Absolute Shrinkage and Selection Operator (LASSO) method to perform variable selection. 
Subbiah et al.~\cite{subbiah2022deep} introduced the Robust ReliefF Mutual Information Recursive Feature Elimination Hybrid Feature Selection (RMR-HFS), which uses a combination of filter and wrapper methods for variable selection.
These selection methods cover a wide range of data analysis needs, from linear to nonlinear correlations, from time series analysis to dimensionality-reduction techniques.
Each method has its unique advantages, helping to better understand and uncover correlations within the data, and enabling the construction of more accurate and efficient predictive models.

\begin{table}[ht]
    \scriptsize
    \renewcommand{\arraystretch}{1.5}
    \centering
    \caption{Common Methods for Selecting External Variables}
    \begin{tabular}{c|>{\centering\arraybackslash}m{5cm}}
        \hline
        \textbf{Selection Methods} & \textbf{Example reference} \\
        \hline
        Pearson Correlation Coefficient & \cite{wang2020short,tang2019short,shaqour2022electrical,bian2022load,hossain2020short,dong2017short,xie2022multi} \\ 
        \hline
        Copula Function Theory & \cite{wang2023short,he2017short} \\
        \hline
        Maximum Information Coefficient & \cite{tang2022short,xiang2023power,jiao2021adaptive,lu2022short,huang2023gated} \\
        \hline
        Autocorrelation Function & \cite{li2021short,javed2022novel,farid2023conv1d} \\
        \hline
        Spearman's Rank Correlation Analysis & \cite{hong2023short,liu2022power,hu2022short,zamee2021online} \\
        \hline
        Principal Component Analysis & \cite{veeramsetty2022short,han2023research} \\
        \hline
    \end{tabular}
    \label{table:selection_methods}
\end{table}

\subsection{Data Reconstruction}
Data reconstruction plays a key role in dealing with the randomness, volatility, periodicity, and diversity \cite{kim2022short} of raw load data, and has become one of the key focus points in many STELF studies. 
A deep exploration of historical load data combined with advanced analytical methods (such as decomposition and clustering techniques) can reveal the key recurring patterns and trends in the data.
These things are crucial for predictive models, as they help the models better understand and capture the dependencies within time-series data.

Several common methods are used for data reconstruction.
The Variational Mode Decomposition (VMD) technique decomposes load data into a series of Intrinsic Mode Functions (IMFs), which are then used to reconstruct the data for training~\cite{ahajjam2022experimental}.
Zang et al.~\cite{zang2021residential} also used VMD technology to decompose the load data into modalities of different frequencies, employing LSTM with self-attention mechanism for forecasting. 
This multi-frequency analysis method allows for a more nuanced handling of the complexity within load data.
Similarly, Mathew et al.~\cite{mathew2021emd} used Empirical Mode Decomposition (EMD) to decompose the raw data into a series of IMFs, and then used the same model to train each mode.
Based on EMD, Ensemble Empirical Mode Decomposition (EEMD)~\cite{yue2022prediction} and Improved Complete Ensemble Empirical Mode Decomposition with Adaptive Noise (ICEEMDAN)~\cite{zhang2023highly} are also used for data reconstruction.

Advanced clustering methods have also been used for data reconstruction \cite{wu2022spatial,yang2019deep,wang2020ensemble}. 
Wu et al.~\cite{wu2022spatial} applied K-shape time-series clustering to categorize users with similar electricity usage habits and characteristics into multiple types.
Yang et al.~\cite{yang2019deep} used the K-means clustering algorithm to identify customer groups with similar electricity usage behaviors.
Wang et al.~\cite{wang2020ensemble} employed the Density-Based Spatial Clustering of Applications with Noise (DBSCAN) algorithm to deal with datasets that contained noise.
Chaturvedi et al.~\cite{chaturvedi2015short} used wavelet transform technology to decompose load data into four wavelet components and then trained a neural network for each component, achieving precise predictions of different frequency characteristics.

Table~\ref{table:reconstruction_methods} summarizes some common data-reconstruction methods. 
Whether using decomposition or clustering techniques, the goal is to reconstruct the overall data to capture the distribution characteristics and underlying patterns of the data.
Due to the complexity of load data, adopting a divide-and-conquer approach (where each part is trained using the same or different models) can enhance the efficiency and accuracy of the model.
Reconstruction techniques not only provide a solid data foundation for the models but also directly influence the model's design, the algorithm selection, and the precision of the forecasting outcomes.

\begin{table}[h!]
    \scriptsize
    \centering
    \caption{Summary of Common Data Reconstruction Methods}
    \begin{tabular}{>{\centering\arraybackslash}m{2.8cm}|c|>{\centering\arraybackslash}m{2cm}|>{\centering\arraybackslash}m{6.2cm}}
        \hline
        \textbf{Reconstruction method} & \textbf{Abbreviation} & \textbf{Example reference} & \textbf{Description} \\
        \hline
        Variational Mode Decomposition & VMD & \cite{wu2023novel,zang2021residential,zhuang2022reliable} & A decomposition method based on the variational principle, solving intrinsic mode functions through optimization problems. \\
        \hline
        Empirical Mode Decomposition & EMD & \cite{fan2020empirical,mounir2023short,ran2023short} & An empirical decomposition method that extracts intrinsic mode functions through an iterative process. \\
        \hline
        Wavelet Transform & WT & \cite{chaturvedi2015short,zhang2022similar} & 
        A method that uses a set of wavelet functions to analyze signals, providing both time (or spatial) and frequency information about the signals. \\
        \hline
        Density-Based Spatial Clustering of Applications with Noise & DBSCAN & \cite{yang2022combined,wang2020ensemble,kong2017short} & A density-based clustering algorithm that classifies points as cluster members, noise, or border points based on their density. \\
        \hline
        K-means Clustering Algorithm & K-means & \cite{hu2022load,yang2019deep} & A clustering algorithm that partitions a dataset into K distinct, non-overlapping groups based on the similarity of data points. \\
        \hline
        K-shape Clustering Algorithm & K-shape & \cite{wu2022spatial,fahiman2017improving} & A clustering algorithm for time-series data that partitions the data into different groups based on shape similarity. \\
        \hline
        Singular Spectrum Analysis & SSA & \cite{niu2016innovative,nie2020novel} & A method for decomposing time-series data by extracting trend, periodic, and noise components to analyze the time series. \\
        \hline
    \end{tabular}
    \label{table:reconstruction_methods}
\end{table}

\section{Answer to RQ4: Methods for Feature Extraction
\label{SEC:Answer to RQ4}}

This section provides the answers to RQ4, examining how deep learning can effectively extract deep non-linear features from the data, enhancing the model's predictive capabilities. 
The purpose of feature extraction is to mine complex relationships from the raw load data that aid the model in understanding load changes. 
In STELF, feature extraction is a critical step, as it directly impacts the performance of the predictive model.

Deep learning can extract features in an unsupervised learning manner, automatically learning useful feature representations from the data~\cite{wang2019review}. 
At different levels of the ANN, features at various levels of abstraction can be learned, providing the model with rich information. 
In STELF, in addition to considering the feature relationships within the time series itself, spatial feature relationships must also be considered \cite{cheung2021leveraging}, such as for large-scale ELF involving regional or urban power grids with distinct spatial relationships. 
The spatial characteristics are also crucial for fully understanding load change patterns and enhancing the accuracy of forecasts.

\subsection{Temporal Feature Relationship Extraction}
Because electricity-load data are time-series data, RNNs can be used to capture the temporal dependencies.
LSTMs and GRUs are often used in temporal feature relationship extraction.
Xu et al.~\cite{xu2018long} used an LSTM to extract deep features of electricity loads and employed an Extreme Learning Machine (ELM) to model shallow patterns.
Abdel et al.~\cite{abdel2022stlf} proposed STLF-Net, using a GRU to get the long-term temporal representations of data.

The success of feature extraction models based on the Multi-Channel One-Dimensional Convolutional Neural Network (MCNN) \cite{dong2023parallel} suggests that convolution operations can also be used to extract feature relationships in time-series data.
This allows for the direct capture of inter-feature relationships on sequence data through One-Dimensional Convolution (Conv1D) operations. 
Although One-Dimensional CNNs (1-D CNNs) are functionally similar to RNNs (such as LSTM and GRU), they also have unique network structural designs to accommodate the varying characteristics and requirements of time-series data. 
These structural designs make it possible for 1-D CNNs to be optimized for specific types of sequence data, making them better at extracting useful feature relationships.
Dai et al.~\cite{dai2023optimized} used CNNs with Conv1D and pooling layers, where the Conv1D layer extracts pivotal features from the input data, and a pooling layer reduces the dimensionality and spatial complexity of the features.

Temporal Convolutional Networks (TCNs) are another type of ANN designed for sequence modeling tasks, especially those involving time-series data.
Because the TCN is based on CNN \cite{bai2018empirical}, it has also been widely used to extract feature vectors and long-term temporal dependencies \cite{bian2022load,wang2020short,zhang2023general}.
Zhang et al.~\cite{zhang2023general} proposed a hybrid network architecture combining TCN and LSTM to address the issue of model complexity.
Their model leveraged the TCN's ability to capture the receptive field in time series and effectively model temporal dependencies, while also incorporating the LSTM's ability to handle long-term dependency problems.

\subsection{Spatial Feature Relationship Extraction}
Extraction of spatial feature relationships involves the construction of an adjacency matrix to represent the connections between nodes in the power network.
Multi-dimensional convolution is used to obtain the spatial features. 
Hua et al.~\cite{hua2023ensemble} proposed a predictive model that combines CNN and GRU, extracting spatial features through the CNN and temporal features through the GRU.
Wan et al.~\cite{wan2023short} used CNNs to extract spatial information from the load data, with the resulting features then input into the RNN for training.

Another approach involves the construction of a spatio-temporal graph model that can simultaneously capture the spatial and temporal dependencies of electricity-load data.
Yu et al.~\cite{yu2021correlated} fed graph-structured data into a Spatio-Temporal Synchronous Graph Convolutional Network (STSGCN) model to perform load forecasting by extracting the inherent spatio-temporal features from historical load data.
The spatio-temporal graph model constructs a similarity-weighted spatio-temporal graph by combining the feature sets of multiple nodes~\cite{huang2023gated}.
The Spatial Convolutional Layer (SCL) extracts the features of neighboring nodes for each node in the graph, thereby enhancing the full-domain node features.

Features extracted using deep learning cannot be directly applied to STELF.
A supervised learning-regression process is required to transform these nonlinear features into prediction results.
A variety of methods (such as linear/nonlinear regression, or neural networks) can be used to perform this mapping \cite{wang2019review}.

\section{Answer to RQ5: Deep-Learning-Based Modeling Methods for STELF
\label{SEC:Answer to RQ5}}

This section provides the answers to RQ5, offering an extensive review of the literature on deep-learning-based predictive models.
From the perspective of forecasting outcomes, predictive models can be categorized as either deterministic or probabilistic  \cite{benidis2022deep}.

\subsection{Deterministic Forecasting Models}
In STELF, deterministic forecasting provides an exact numerical prediction, offering precise predictions of the load level at a specific point in time or over a certain period in the future. 
This type of forecasting focuses on delivering a concrete value rather than a range or distribution. 
Due to the extensive literature and methods involved, we examine this type of forecasting from the perspectives of single and hybrid models.

\subsubsection{Single Models}
Simplicity is the main advantage of the single model, which is easy to understand and construct. 
Deep learning algorithms make use of deep networks, consisting of a series of complex hidden layers~\cite{eren2024comprehensive}.
Early STELF achieved predictive results by stacking ANNs~\cite{singh2018integration,din2017short}.
Chen et al.~\cite{chen2019day} and Hossen et al.~\cite{hossen2017short} explored ELF using DNNs. 
Chen et al. proposed a method based on two-terminal sparse coding and deep neural network fusion, while Hossen et al. examined the impact of single-layer versus double-layer DNN architectures.
A DBN, similar in structure to DNN, combines multiple RBMs for ELF and uses a layer-by-layer unsupervised learning method to pre-train the initial weights~\cite{dedinec2016deep}.

DNNs or DBNs formed by stacking multiple layers typically lack memory capability, which means that they may not be able to use previous information effectively when processing time-series data.
RNNs use recurrent connections, allowing the network to retain previous information while processing sequence data, potentially adjusting the output based on earlier elements in the sequence. 
This makes RNNs (including LSTM and GRU, as introduced in Section~\ref{SEC:Background}) very popular for time-series forecasting.
LSTMs and GRUs are advanced RNNs that have been used in STELF~\cite{kong2017short,morais2023short}.
Zhu et al.~\cite{zhu2022lstm} proposed a dual-attention encoder-decoder structure using attention mechanisms, using an LSTM as a specific encoder and decoder for nonlinear dynamic time modeling. 
This attention mechanism dynamically focused on the importance of different parts of the input sequence, significantly enhancing the performance of RNNs used in conjunction.
Aseeri et al.~\cite{aseeri2023effective} also used a GRU structure to focus on key variables, improving performance, particularly with longer sequences.

A bidirectional RNN is an innovative ANN architecture that integrates two recurrent layers, each with a distinct role. 
One layer captures the forward flow of the sequence, while the other focuses on the backward flow.
This bidirectional processing mechanism makes it possible to comprehensively understand the sequence data, enabling feature and information extraction from both temporal directions simultaneously \cite{mughees2021deep}.
This bidirectional structure, such as in the Bidirectional Long Short-Term Memory (BiLSTM)~\cite{wang2019bi} and Bidirectional Gated Recurrent Unit (BiGRU)~\cite{shaqour2022electrical}, has a strong predictive capability, and is popular in STELF research.

Using a CNN as a standalone STELF model involves treating time-series data directly as one-dimensional images or creating images from the sequence values of multivariate time series~\cite{sadaei2019short}. 
The convolutional layers in CNNs are used to extract local features; while pooling layers reduce the dimensionality of time-series data
---
this helps with both extracting more abstract features and reducing computational complexity~\cite{jalali2021novel}.
Finally, one or more fully connected layers are used to generate the output sequence.
As a variant of CNNs, TCNs extend the receptive field and capture long-range sequence dependencies by increasing the spacing of convolutional kernels \cite{bai2018empirical}.
TCNs extract the complex interactions between time-series and non-time-series data, resulting in precise feature quantities~\cite{bian2022research}.
Yin et al.~\cite{yin2021multi} used TCNs to extract key features from multiple spatial scale samples, yielding initial predictive outcomes.
A similar model based on CNN, WaveNet, is a generative model developed by DeepMind, primarily used for generating audio waveforms~\cite{van2016wavenet}.
WaveNet has also been used for STELF~\cite{voss2018residential,lin2021spatial}.

The Transformer model, proposed by Vaswani et al.~\cite{vaswani2017attention} in 2017, has seen great success in various natural language processing tasks, and has also been extended to other fields such as image processing and time-series forecasting.
The Transformer's powerful feature-extraction and sequence-modeling capabilities also make it a good choice for STELF~\cite{zhang2023transformgraph,zhang2022short,nawar2023transfer}.

Two models based on the Transformer are the Informer and the Temporal Fusion Transformer (TFT)~\cite{gao2023short,santos2023deep}. 
The Informer model focuses on the most critical time steps, using a probabilistic sparsity approach, ignoring less important information.
Gao et al.~\cite{gao2023short}, for example, established a hybrid ELF model based on the Informer.
Yu et al.~\cite{yu2022self} proposed a self-attention-based STELF method considering demand-side management, using an informer to independently predict and reconstruct the decomposed intrinsic mode function components.
The TFT model focuses on how to integrate different types of time-series data to improve prediction accuracy.
Giacomazzi et al.~\cite{giacomazzi2023short} explored the potential of TFT for hourly STELF across different time ranges (such as the previous day and the previous week).

\subsubsection{Hybrid Models}
In STELF, hybrid deep-learning models are becoming a key technology for solving complex forecasting problems.
Hybrid models integrate a variety of deep-learning models, with advantages including their diversity and flexibility.
Complementing each other's strengths, they enhance the accuracy and robustness of predictions~\cite{lin2022hybrid}.
The hybrid model primarily employs two strategies for combination: stage-wise training and joint training. 
Stage-wise training involves focusing on specific learning tasks at each stage, while joint training involves training all components of the hybrid model simultaneously.

(1) \textit{Stage-wise Training:}
Stage-wise training strategy addresses some challenges for training complex models by breaking down the training process into a series of orderly stages. 
In each stage, a part of the model is independently trained and optimized to learn specific patterns.

A hybrid model integrating CNN with RNN has become a widely adopted solution, due to its exceptional performance~\cite{feng2024stgnet,shi2023short,yi2023deep,sekhar2023robust,aouad2022cnn,guo2021short}. 
Zhang et al.~\cite{zhang2023cnn} proposed a hybrid model based on CNN and LSTM, using CNN layers for feature extraction from the input dataset and an LSTM model for sequence prediction, supporting multi-step forecasting of time-series data.
Their approach leverages the efficient CNN capability to extract local features, using their output feature vectors as inputs for the RNN, which allows the RNN to further capture the dynamic changes and long-term dependencies in the time series.
Jin et al.~\cite{jin2022short} used a CNN-GRU hybrid model based on parameter-transfer learning.
By transferring the parameters of a trained model from one with a large dataset to another trained with a smaller dataset, the model's performance and predictive accuracy were enhanced.
Here, transfer learning was used to address the issue of insufficient training data~\cite{ozer2021combined}.    
    
There are also some novel hybrid models based on the stage-wise structure.
The combination of Graph Neural Network (GNN) and TCN passes the features extracted by the GNN to the TCN for further training~\cite{lin2021spatial}.
The combination of LSTM and TFT uses the LSTM as an encoder-decoder to preprocess the data, which is then fed as training data into the TFT~\cite{santos2023deep}.
A more complex fusion proposed by Bu et al.~\cite{bu2023hybrid} involves a hybrid model combining Conditional Generative Adversarial Networks (CGAN) with CNN.
The CGAN uses the CNN's ability to accurately capture internal features to generate realistic fake samples, while the semi-supervised regression layer optimizes the discriminator to enhance sample authenticity recognition.
   
 (2) \textit{Joint Training:}
Joint training integrates various components of a model into a unified training framework, enabling synchronous training of each component. 
Data reconstruction techniques can result in different modalities, requiring models to be trained for each modality \cite{wang2023short,zhang2023improved,luo2022ensemble}. 
This approach leverages the characteristics of the different modalities, training concurrently, enhancing the overall performance.
    
Data can be divided into multiple components, with different models being used to train separately and simultaneously on their respective data.
He et al.~\cite{he2021per} used Per-unit Curve Rotation Decoupling (PCRD) to decompose the load into three parts: the rotating unit-load curve; the zero AM load; and the daily average load.
A CNN extracted the shape features of the rotating unit-load curve, while a TCN simultaneously extracted the temporal features of the zero AM load and daily average load.
This divide-and-conquer mechanism generates multiple preliminary prediction results, which ultimately need to be synthesized into a final prediction using weighted averaging, voting mechanisms, or attention mechanisms.
Hua et al.~\cite{hua2023ensemble} used an attention mechanism to dynamically connect the preliminary results, with a CNN capturing spatial features and a GRU capturing temporal features.
    
Unlike existing ELF methods that place separate convolutional layers on top of the entire RNN, Liu et al.~\cite{liu2022image} embedded a 3-D convolutional filter within the LSTM unit, enabling the capture of translation-invariant local patterns both within and across spatial neighborhoods in the channels.
This strategy of embedding one deep-learning model into another, through hierarchical model integration, enabled deep abstraction and effective utilization of data features.
Eskandari et al.~\cite{eskandari2021convolutional} also used this strategy, combining a GRU and an LSTM to form a bidirectionally-propagating neural network.
This used multidimensional features extracted by a 2-D CNN as input and provided these features to the bidirectional units for hourly ELF.

\subsection{Probabilistic Forecasting Models}
Probabilistic ELF predicts future demand using the uncertainty and randomness of the load.
Unlike traditional deterministic methods, probabilistic models provide a quantified expression of predictive uncertainty, which has significant advantages for power grid planning and operation~\cite{yang2018power}.
Probabilistic models express the uncertainty of prediction results by generating a distribution of outcomes, rather than a single value~\cite{jalali2022advanced}.

There are two main approaches for probabilistic ELF: 
parametric and non-parametric methods\cite{hou2022review}.
Parametric methods are based on assumptions about the distribution of the load data, typically assuming that the data follows a known probability distribution (such as the normal distribution or the Poisson distribution).
Due to their use of fewer assumptions, non-parametric methods have much broader applicability~\cite{van2018review}.
Non-parametric methods do not require that the data follow a specific distribution form, starting instead directly from the actual data, uncovering the probabilistic distribution characteristics through the data itself~\cite{huang2020improved}.
To the best of our knowledge, the application of deep learning in probabilistic ELF extends from deterministic point predictions to probabilistic distribution forecasting~\cite{feng2019reinforced,cheng2021probabilistic,chen2018short}.
This process requires the generation of point predictions, and then uses non-parametric techniques (such as quantile regression, bootstrapping, confidence interval estimation, gradient boosting, and kernel density estimation~\cite{wang2019review}) to construct a probabilistic ELF model.

Lin et al.~\cite{lin2022short} introduced an LSTM with a two-stage attention mechanism for probabilistic short-term regional load forecasting, enhancing uncertainty estimation and accuracy when trained with quantile loss.
The combination of deep learning and non-parametric methods is an innovative solution to probabilistic ELF. 
This harnesses the deep-learning ability to capture data complexity and generate accurate point predictions, 
and enables forecasts with probability distributions through non-parametric techniques.
Wang et al.~\cite{wang2019probabilistic} extended the traditional LSTM-based point prediction to a quantile-based probabilistic forecast.
Liu et al.~\cite{liu2022short} integrated the GRU deep-feature-extraction capability with the CNN's efficient parallel processing, while also employing kernel density estimation to accurately fit the probability density.
Zhang et al.~\cite{zhang2023novel} proposed the Quantile Regression Convolutional Bidirectional Long Short-Term Memory (QRCNNBiLSTM), which integrates quantile regression with feature extraction and bidirectional data processing.
Through this integration, QRCNNBiLSTM can make precise joint predictions for the upper and lower bounds of the forecast interval.

The integration of deep learning and non-parametric methods in probabilistic ELF can more accurately capture the uncertainty in load variations, providing more comprehensive and reliable decision support for power systems.

\section{Answer to RQ6: Optimizing the Training Process
\label{SEC:Answer to RQ6}}
This section provides the answers to RQ6, examining ways to optimize deep-learning training processes~\cite{zhang2021review}.

Optimization mainly involves two aspects: 
network-structure optimization and error optimization.
Optimization of the network structure relates to the architectural design of the model, and involves adjusting the number of layers, configuring the neurons, and modifying the connection methods.
This all aims at constructing a robust model capable of capturing the complex features of the data.
Error optimization involves thorough analysis and fine-tuning of the prediction errors, aiming to minimize the discrepancy between the model's predictions and the actual observed values, thereby enhancing the accuracy and reliability.

\subsection{Network-Structure Optimization}
Network architecture design is a very important task, requiring careful selection of the number of neurons in each layer and the number of hidden layers in the network, etc~\cite{jiang2022deep}.
The selection is not completed in a single step, but rather needs to be determined based on the nature of the problem, the characteristics of the data, and the expected performance of the model.
Selecting the optimal network structure and model parameters is a complex process, with a number of trial-and-error methods and heuristic optimization algorithms having been proposed~\cite{wang2019review}.

As a fundamental problem-solving method, the trial-and-error approach involves gradually approaching a solution through continuous attempts and adjustments.
This is not limited to only simple experiments and adjustments, but has been combined with heuristic optimization techniques to form efficient and systematic optimization strategies.
The heuristic algorithms draw inspiration from optimization mechanisms found in nature and social phenomena, including Particle Swarm Optimization (PSO)~\cite{hong2023short}, Genetic Algorithms (GAs)~\cite{dong2021electrical}, the Whale Optimization Algorithm (WOA)~\cite{haiyan2020short}, Grey Wolf Optimizer (GWO)~\cite{sekhar2023robust}, and the Grasshopper Optimization Algorithm (GOA)~\cite{hu2022development}.  

Methods based on Bayesian optimization~\cite{bayram2023lstm,xu2022probabilistic} predict parameter performance using probabilistic models, guiding the search process to more intelligently select the next set of candidate parameters.
This is particularly well-suited for high-dimensional parameter spaces, significantly reducing the number of evaluations required.
In summary, more accurate optimization algorithms can help to more effectively explore the potentially vast parameter space and identify parameter combinations that significantly enhance performance.

\subsection{Error Optimization}
Once the loss function is defined, the network weights need to be updated by calculating the loss-function gradient using optimization algorithms, such as gradient descent and its variants (Adam, RMSprop, etc.~\cite{wu2023pulse,ganjouri2023spatial})~\cite{hong2016probabilistic}.
These algorithms adjust the model's parameters to gradually reduce the loss-function value, improving the model's accuracy.
The following is a list of some commonly-used optimizers:
\begin{itemize}
    \item 
    SGD~\cite{sakib2021data}: 
    This uses only one sample to compute the gradient and update the parameters in each iteration. 
    This allows for quick updates, but can result in high variance, due to only using a single sample.
    
    \item 
    RMSprop~\cite{langevin2023efficient,yazici2022deep,he2017load}: 
    This adjusts the learning rate by maintaining a decaying average of the squared gradients for each parameter, achieving adaptive learning rates. 
    This is particularly effective for handling different learning rates for different parameters, accelerating convergence, and avoiding local minima or saddle points.
  
    \item 
    AdaBelief~\cite{yu2022self}: 
    This employs a concept of belief, which depends on the ratio of the squared gradient to its historical mean. 
    When this ratio is greater than 1, AdaBelief is more inclined to trust the current gradient information; 
    otherwise, it relies more on previous information.
    
    \item 
    Adam~\cite{wang2023electrical,su2023residential,sun2020short}: 
    This is an adaptive gradient descent method that independently adjusts the learning rate for each parameter by combining the exponentially weighted averages of the first and second moments. 
    This approach achieves a fast and robust optimization process.
\end{itemize}

\section{Answer to RQ7: Evaluation of Forecast Results
\label{SEC:Answer to RQ7}}
This section provides the answers to RQ7, examining ways to evaluate the accuracy of the prediction results.
This not only involves verifying the accuracy of the model outputs (to ensure that the predicted results are close to the actual load values), but also relates to the model's reliability and effectiveness.

A comprehensive evaluation of the prediction model performance requires a set of evaluation metrics. 
These metrics should be able to quantify the size and distribution of prediction errors from different perspectives.
They should enable a fair and objective comparison of the prediction performance of different models, under a unified standard.
Metrics for the evaluation of deterministic load forecasting include:
Mean Squared Error (MSE); 
Root Mean Squared Error (RMSE); 
Mean Absolute Error (MAE);
Mean Absolute Percentage Error (MAPE); and 
$R^2$. 
Similarly, metrics for the evaluation of probabilistic load forecasting include:
Continuous Ranked Probability Score (CRPS);
Prediction Interval Coverage Probability (PICP); and 
Pinball Loss (PL).
Typically, multiple evaluation metrics are used together to enable a multi-faceted assessment.
Table~\ref{tab:evaluation_metrics} lists the formulas and descriptions for some commonly-used evaluation metrics. 

\renewcommand{\arraystretch}{2}
\begin{table}[htbp]
    \centering
    \caption{Common Evaluation Metrics}
    \begin{adjustbox}{width=\textwidth}
    \begin{tabular}{c|c|>{\centering\arraybackslash}m{6cm}|>{\centering\arraybackslash}m{4cm}}
        \hline
        \textbf{Evaluation Metrics} & \textbf{Formula} & \textbf{Description} & \textbf{Purpose} \\
        \hline
        MSE & $\frac{1}{n} \sum_{i=1}^{n} (y_i - \hat{y}_i)^2$ & Measures the difference between predicted values and actual values, focusing on large errors& \multirow{5}{*}{Deterministic Forecasting} \\
         \cline{1-3}
        RMSE & $\sqrt{\frac{1}{n} \sum_{i=1}^{n} (y_i - \hat{y}_i)^2}$ & The square root of the MSE, preserving its properties while also keeping consistent with the original data, making interpretation easier&  \\
         \cline{1-3}
        MAE & $\frac{1}{n} \sum_{i=1}^{n} |y_i - \hat{y}_i|$ & Measures the average absolute difference between predicted and actual values, focusing on small errors&  \\
        \cline{1-3}
        MAPE & $\frac{1}{n} \sum_{i=1}^{n} \left| \frac{y_i - \hat{y}_i}{y_i} \right| \times 100\%$ & Measures the error as a percentage relative to the actual values, suitable for proportional data&  \\
         \cline{1-3}
        R² & $1 - \frac{\sum_{i=1}^{n} (y_i - \hat{y}_i)^2}{\sum_{i=1}^{n} (y_i - \overline{y})^2}$ & Assesses the model's explanatory power, with values closer to 1 indicating a superior fit to the data&  \\
         \hline
        CRPS & $\int_{-\infty}^{\infty} \left( F(x) - \mathbf{1}_{\{x \geq y\}} \right)^2 dx$ & Measures the difference between the probabilistic forecast distribution and the actual observations& \multirow{3}{*}{Probabilistic Forecasting} \\
         \cline{1-3}
        PICP & $\frac{1}{N} \sum_{i=1}^{N} \mathbf{1}_{\{a_i \leq y_i \leq b_i\}}$ & Assesses how often the prediction intervals contain the actual observations& \\
         \cline{1-3}
        PL & $\frac{1}{n} \sum_{i=1}^{n} \left( \tau (y_i - \hat{y}_i) \mathbf{1}_{\{y_i \geq \hat{y}_i\}} + (1 - \tau) (\hat{y}_i - y_i) \mathbf{1}_{\{y_i < \hat{y}_i\}} \right)$ & Assesses the effectiveness of quantile predictions& \\
        \hline
    \end{tabular}
    \end{adjustbox}
    \label{tab:evaluation_metrics}
\end{table}

Recently, some novel and improved evaluation metrics have also been introduced.
Faustine et al.~\cite{faustine2022fpseq2q} and Ganjouri et al.~\cite{ganjouri2023spatial}, for example, used a Normalized Root Mean Square Error (NRMSE) as evaluation metrics; 
Wang et al.~\cite{wang2023shortc} used an Average Interval Score (AIS), Coverage Probability (CP), and Specificity Probability (SP) to evaluate the performance of interval probabilistic forecasting; and
Zhang et al.~\cite{zhang2023regional} introduced the Coverage Rate (CR) to evaluate the predictive coverage of the model (which represents the proportion of times the confidence intervals generated by the model cover the true values out of the total number of instances) and the Interval Average Convergence (IAC) to assess the model's convergence.

\section{Answer to RQ8: Challenges and Future Development Trends of STELF
\label{SEC:Answer to RQ8}}
This section provides the answers to RQ8, examining the challenges and potential future opportunities for the application of deep learning in STELF. 

\subsection{Challenges}
Although there has been significant effort dedicated to exploring the application of deep learning in STELF, especially in recent years, some challenges remain.
Some of these challenges include:
the need for standardized datasets; 
the generalizability of models; 
insufficient research in probabilistic load forecasting; 
the interpretability of deep learning results; and 
real-time prediction capabilities. 
Addressing and overcoming these challenges will be critical for the widespread application of deep learning in the field of STELF. 
In particular:
\begin{itemize}
    \item 
    STELF researchers use a variety of datasets, including both private and public data. 
    This can present a challenge for verifying model performance. 
    The lack of standardized benchmark datasets makes it complex and difficult to compare performances.
    Developing some widely recognized standardized datasets is thus of urgent importance for advancing STELF research.
    
    \item 
    Although many studies have addressed the common goal of ELF, they often focus on different application scenarios and loads, such as building loads~\cite{chiu2023hybrid}, household loads~\cite{fekri2021deep}, and city-level loads~\cite{yang2023iterative}. 
    The specificity of these scenarios may lead to models being overly optimized for a particular environment, raising challenges for the model's generalizability.
    Therefore, It is necessary to study models with strong generalization ability for use in different scenarios.

    \item 
    Although the academic community has extensively studied deterministic STELF, there remains a lack of research into probabilistic load forecasting. 
    Compared with deterministic forecasting, there is significantly less focus on deep-learning-based probabilistic ELF models. 
    Therefore, more research needs to be applied to probabilistic forecasting models, which will enhance their ability to cope with uncertainties.
    
    \item 
    The interpretability of deep-learning-based forecasting models is an unresolved challenge. 
    The complex mechanisms and decision pathways used by these models are often unknown, and lack intuitive transparency.
    This can impact the acceptance and effectiveness of the models in practical applications. 
    Therefore, improving the interpretability of deep learning is crucial for its widespread adoption, long-term operation, and decision support in power systems.
    
    \item 
    The demand for STELF is increasing, especially for real-time capabilities. 
    However, most previous research is based on offline learning modes, which rely on large amounts of historical data for training, and may not easily incorporate the latest data for real-time learning~\cite{eren2024comprehensive}. 
    To address this challenge, online learning mechanisms will be essential.

\end{itemize}

\subsection{Research Trends}
Recent STELF research trends include moving towards better integration, precision, and intelligence. 
Ongoing work to develop more sophisticated and enhanced forecasting models aims at improving prediction accuracy and reliability. 
This section examines the future STELF research directions of image-processing techniques, Large Language Models (LLMs), and optimization.
\begin{itemize}

\item 
As power networks evolve, dynamic graph models can adapt to changes in nodes and edges, maintaining the flexibility of the forecasting model. 
Graph techniques can integrate multiple data sources (such as geographical locations, historical loads, and weather conditions).
Our review found that most STELF graph techniques use convolution operations to extract temporal or spatial feature relationships. 
However, Liu et al.~\cite{liu2022image} converted the data into an image, and part of the future values to be predicted was transformed into blank patches.
Thus, estimating future values became a similar problem to generating pixels for the missing regions of an image.
Similar to image inpainting techniques, image generation and segmentation technologies have not yet been widely applied in STELF.
Although these techniques may seem unrelated to STELF, their core ideas and methodologies can provide new perspectives and possibilities for STELF research.
By applying these key image-processing concepts and algorithms to STELF, new research directions can be created, and models' ability to identify and predict complex load patterns can be enhanced.

\item 
The recent rapid development of LLM technology has led to the exploration of its application to time-series forecasting tasks, opening up new areas in time-series prediction~\cite{yu2023temporal,chang2023llm4ts}.
However, the use of LLMs for STELF has not yet been developed. 
Nevertheless, it is anticipated that application of LLMs to STELF shall become an important research direction.
This may provide us with new ways to address the challenge of insufficient generalizability of forecasting models, while also opening up the possibility to achieve zero-shot learning in ELF~\cite{gruver2024large}.
This type of model may not only demonstrate stronger adaptability and predictive power on diverse datasets, but also make reasonable predictions on unseen data.

\item
Optimization is a critical stage in the development of deep-learning models, and has been receiving attention in STELF ~\cite{zhang2021review}.
Future research will continue to explore more efficient optimization algorithms, such as improved SSA~\cite{neeraj2021long}, Adam~\cite{hong2022week}, and other optimization methods tailored to specific problems.
Network architecture optimization is evolving towards more lightweight networks, with the relevant error-optimization process becoming more refined (including in-depth research and customization of loss functions).
Model compression techniques and acceleration algorithms will continue to evolve, leading the optimization process to place greater emphasis on computational efficiency.

\end{itemize}

\section{Conclusion
\label{SEC:Conclusion}}
This review paper has examined advances in the application of deep learning in STELF over the past decade. 
Over this period, the application of deep learning to STELF has grown in popularity, and will continue to do so.
We employed a comprehensive review-research methodology to identify the relevant literature, ensuring the breadth and depth of the research findings.
We used specific keywords to search six major databases and conducted manual screening to ensure the completeness of the data.
During the literature review process, we specifically extracted and recorded key information from each paper based on the designed eight RQs.
We carefully organized the extracted information and created specific charts and tables to help readers understand.

In this paper, we have analyzed and summarized the search results, and provided a detailed analysis of publication trends.
We organized the structure of the paper according to the practical workflow of STELF, (including the introduction of the dataset, data preprocessing methods, feature extraction methods, the introduction of deep learning models, optimization methods, and evaluation metrics) and conducted an in-depth analysis of each stage.
The content of each prediction step is categorized according to a specific method to ensure it is presented in an organized manner.
We have provided concise explanations of commonly-used techniques in conjunction with cited literature.
This structured presentation makes the content clear and logical, while also helping researchers quickly understand the research trends and core issues in this domain.
We have also summarized the challenges and future research trends in the STELF field.
Overall, this review paper, with its comprehensive, systematic approach, and guidance, provides significant academic value and practical relevance.

\bibliographystyle{ACM-Reference-Format}
\bibliography{STELF}


\begin{thebibliography}{209}


\ifx \showCODEN    \undefined \def \showCODEN     #1{\unskip}     \fi
\ifx \showDOI      \undefined \def \showDOI       #1{#1}\fi
\ifx \showISBNx    \undefined \def \showISBNx     #1{\unskip}     \fi
\ifx \showISBNxiii \undefined \def \showISBNxiii  #1{\unskip}     \fi
\ifx \showISSN     \undefined \def \showISSN      #1{\unskip}     \fi
\ifx \showLCCN     \undefined \def \showLCCN      #1{\unskip}     \fi
\ifx \shownote     \undefined \def \shownote      #1{#1}          \fi
\ifx \showarticletitle \undefined \def \showarticletitle #1{#1}   \fi
\ifx \showURL      \undefined \def \showURL       {\relax}        \fi
\providecommand\bibfield[2]{#2}
\providecommand\bibinfo[2]{#2}
\providecommand\natexlab[1]{#1}
\providecommand\showeprint[2][]{arXiv:#2}

\bibitem[Abdel-Basset et~al\mbox{.}(2022)]%
        {abdel2022stlf}
\bibfield{author}{\bibinfo{person}{Mohamed Abdel-Basset}, \bibinfo{person}{Hossam Hawash}, \bibinfo{person}{Karam Sallam}, \bibinfo{person}{Sameh~S Askar}, {and} \bibinfo{person}{Mohamed Abouhawwash}.} \bibinfo{year}{2022}\natexlab{}.
\newblock \showarticletitle{{STLF-Net}: Two-stream deep network for short-term load forecasting in residential buildings}.
\newblock \bibinfo{journal}{\emph{Journal of King Saud University-Computer and Information Sciences}} \bibinfo{volume}{34}, \bibinfo{number}{7} (\bibinfo{year}{2022}), \bibinfo{pages}{4296--4311}.
\newblock


\bibitem[Ahajjam et~al\mbox{.}(2022)]%
        {ahajjam2022experimental}
\bibfield{author}{\bibinfo{person}{Mohamed~Aymane Ahajjam}, \bibinfo{person}{Daniel~Bonilla Licea}, \bibinfo{person}{Mounir Ghogho}, {and} \bibinfo{person}{Abdellatif Kobbane}.} \bibinfo{year}{2022}\natexlab{}.
\newblock \showarticletitle{Experimental investigation of variational mode decomposition and deep learning for short-term multi-horizon residential electric load forecasting}.
\newblock \bibinfo{journal}{\emph{Applied Energy}}  \bibinfo{volume}{326} (\bibinfo{year}{2022}), \bibinfo{pages}{119963}.
\newblock


\bibitem[Akhtar et~al\mbox{.}(2023)]%
        {akhtar2023short}
\bibfield{author}{\bibinfo{person}{Saima Akhtar}, \bibinfo{person}{Sulman Shahzad}, \bibinfo{person}{Asad Zaheer}, \bibinfo{person}{Hafiz~Sami Ullah}, \bibinfo{person}{Heybet Kilic}, \bibinfo{person}{Radomir Gono}, \bibinfo{person}{Micha{\l} Jasi{\'n}ski}, {and} \bibinfo{person}{Zbigniew Leonowicz}.} \bibinfo{year}{2023}\natexlab{}.
\newblock \showarticletitle{Short-term load forecasting models: {A} review of challenges, progress, and the road ahead}.
\newblock \bibinfo{journal}{\emph{Energies}} \bibinfo{volume}{16}, \bibinfo{number}{10} (\bibinfo{year}{2023}), \bibinfo{pages}{4060}.
\newblock


\bibitem[Al~Mamun et~al\mbox{.}(2020)]%
        {al2020comprehensive}
\bibfield{author}{\bibinfo{person}{Abdullah Al~Mamun}, \bibinfo{person}{Md Sohel}, \bibinfo{person}{Naeem Mohammad}, \bibinfo{person}{Md~Samiul~Haque Sunny}, \bibinfo{person}{Debopriya~Roy Dipta}, {and} \bibinfo{person}{Eklas Hossain}.} \bibinfo{year}{2020}\natexlab{}.
\newblock \showarticletitle{A comprehensive review of the load forecasting techniques using single and hybrid predictive models}.
\newblock \bibinfo{journal}{\emph{IEEE Access}}  \bibinfo{volume}{8} (\bibinfo{year}{2020}), \bibinfo{pages}{134911--134939}.
\newblock


\bibitem[Alipour et~al\mbox{.}(2020)]%
        {alipour2020novel}
\bibfield{author}{\bibinfo{person}{Mohammadali Alipour}, \bibinfo{person}{Jamshid Aghaei}, \bibinfo{person}{Mohammadali Norouzi}, \bibinfo{person}{Taher Niknam}, \bibinfo{person}{Sattar Hashemi}, {and} \bibinfo{person}{Matti Lehtonen}.} \bibinfo{year}{2020}\natexlab{}.
\newblock \showarticletitle{A novel electrical net-load forecasting model based on deep neural networks and wavelet transform integration}.
\newblock \bibinfo{journal}{\emph{Energy}}  \bibinfo{volume}{205} (\bibinfo{year}{2020}), \bibinfo{pages}{118106}.
\newblock


\bibitem[Almalaq and Edwards(2017)]%
        {almalaq2017review}
\bibfield{author}{\bibinfo{person}{Abdulaziz Almalaq} {and} \bibinfo{person}{George Edwards}.} \bibinfo{year}{2017}\natexlab{}.
\newblock \showarticletitle{A review of deep learning methods applied on load forecasting}. In \bibinfo{booktitle}{\emph{Proceedings of the 16th IEEE International Conference on Machine Learning and Applications (ICMLA'17)}}. \bibinfo{pages}{511--516}.
\newblock


\bibitem[Aouad et~al\mbox{.}(2022)]%
        {aouad2022cnn}
\bibfield{author}{\bibinfo{person}{Mosbah Aouad}, \bibinfo{person}{Hazem Hajj}, \bibinfo{person}{Khaled Shaban}, \bibinfo{person}{Rabih~A Jabr}, {and} \bibinfo{person}{Wassim El-Hajj}.} \bibinfo{year}{2022}\natexlab{}.
\newblock \showarticletitle{A {CNN-Sequence-to-Sequence} network with attention for residential short-term load forecasting}.
\newblock \bibinfo{journal}{\emph{Electric Power Systems Research}}  \bibinfo{volume}{211} (\bibinfo{year}{2022}), \bibinfo{pages}{108152}.
\newblock


\bibitem[Arastehfar et~al\mbox{.}(2022)]%
        {arastehfar2022short}
\bibfield{author}{\bibinfo{person}{Sana Arastehfar}, \bibinfo{person}{Mohammadjavad Matinkia}, {and} \bibinfo{person}{Mohammad~Reza Jabbarpour}.} \bibinfo{year}{2022}\natexlab{}.
\newblock \showarticletitle{Short-term residential load forecasting using graph convolutional recurrent neural networks}.
\newblock \bibinfo{journal}{\emph{Engineering Applications of Artificial Intelligence}}  \bibinfo{volume}{116} (\bibinfo{year}{2022}), \bibinfo{pages}{105358}.
\newblock


\bibitem[Aseeri(2023)]%
        {aseeri2023effective}
\bibfield{author}{\bibinfo{person}{Ahmad~O Aseeri}.} \bibinfo{year}{2023}\natexlab{}.
\newblock \showarticletitle{Effective RNN-based forecasting methodology design for improving short-term power load forecasts: Application to large-scale power-grid time series}.
\newblock \bibinfo{journal}{\emph{Journal of Computational Science}}  \bibinfo{volume}{68} (\bibinfo{year}{2023}), \bibinfo{pages}{101984}.
\newblock


\bibitem[Atef and Eltawil(2020)]%
        {atef2020assessment}
\bibfield{author}{\bibinfo{person}{Sara Atef} {and} \bibinfo{person}{Amr~B Eltawil}.} \bibinfo{year}{2020}\natexlab{}.
\newblock \showarticletitle{Assessment of stacked unidirectional and bidirectional long short-term memory networks for electricity load forecasting}.
\newblock \bibinfo{journal}{\emph{Electric Power Systems Research}}  \bibinfo{volume}{187} (\bibinfo{year}{2020}), \bibinfo{pages}{106489}.
\newblock


\bibitem[Bai et~al\mbox{.}(2018)]%
        {bai2018empirical}
\bibfield{author}{\bibinfo{person}{Shaojie Bai}, \bibinfo{person}{J~Zico Kolter}, {and} \bibinfo{person}{Vladlen Koltun}.} \bibinfo{year}{2018}\natexlab{}.
\newblock \showarticletitle{An empirical evaluation of generic convolutional and recurrent networks for sequence modeling}.
\newblock \bibinfo{journal}{\emph{arXiv preprint arXiv:1803.01271}} (\bibinfo{year}{2018}).
\newblock


\bibitem[Bashir et~al\mbox{.}(2022)]%
        {bashir2022short}
\bibfield{author}{\bibinfo{person}{Tasarruf Bashir}, \bibinfo{person}{Chen Haoyong}, \bibinfo{person}{Muhammad~Faizan Tahir}, {and} \bibinfo{person}{Zhu Liqiang}.} \bibinfo{year}{2022}\natexlab{}.
\newblock \showarticletitle{Short term electricity load forecasting using hybrid prophet-LSTM model optimized by BPNN}.
\newblock \bibinfo{journal}{\emph{Energy reports}}  \bibinfo{volume}{8} (\bibinfo{year}{2022}), \bibinfo{pages}{1678--1686}.
\newblock


\bibitem[Bayram et~al\mbox{.}(2023)]%
        {bayram2023lstm}
\bibfield{author}{\bibinfo{person}{Firas Bayram}, \bibinfo{person}{Phil Aupke}, \bibinfo{person}{Bestoun~S Ahmed}, \bibinfo{person}{Andreas Kassler}, \bibinfo{person}{Andreas Theocharis}, {and} \bibinfo{person}{Jonas Forsman}.} \bibinfo{year}{2023}\natexlab{}.
\newblock \showarticletitle{{DA-LSTM}: A dynamic drift-adaptive learning framework for interval load forecasting with {LSTM} networks}.
\newblock \bibinfo{journal}{\emph{Engineering Applications of Artificial Intelligence}}  \bibinfo{volume}{123} (\bibinfo{year}{2023}), \bibinfo{pages}{106480}.
\newblock


\bibitem[Benidis et~al\mbox{.}(2022)]%
        {benidis2022deep}
\bibfield{author}{\bibinfo{person}{Konstantinos Benidis}, \bibinfo{person}{Syama~Sundar Rangapuram}, \bibinfo{person}{Valentin Flunkert}, \bibinfo{person}{Yuyang Wang}, \bibinfo{person}{Danielle Maddix}, \bibinfo{person}{Caner Turkmen}, \bibinfo{person}{Jan Gasthaus}, \bibinfo{person}{Michael Bohlke-Schneider}, \bibinfo{person}{David Salinas}, \bibinfo{person}{Lorenzo Stella}, {et~al\mbox{.}}} \bibinfo{year}{2022}\natexlab{}.
\newblock \showarticletitle{Deep learning for time series forecasting: {T}utorial and literature survey}.
\newblock \bibinfo{journal}{\emph{Comput. Surveys}} \bibinfo{volume}{55}, \bibinfo{number}{6} (\bibinfo{year}{2022}), \bibinfo{pages}{1--36}.
\newblock


\bibitem[Bian et~al\mbox{.}(2022a)]%
        {bian2022load}
\bibfield{author}{\bibinfo{person}{Haihong Bian}, \bibinfo{person}{Qian Wang}, \bibinfo{person}{Guozheng Xu}, {and} \bibinfo{person}{Xiu Zhao}.} \bibinfo{year}{2022}\natexlab{a}.
\newblock \showarticletitle{Load forecasting of hybrid deep learning model considering accumulated temperature effect}.
\newblock \bibinfo{journal}{\emph{Energy Reports}}  \bibinfo{volume}{8} (\bibinfo{year}{2022}), \bibinfo{pages}{205--215}.
\newblock


\bibitem[Bian et~al\mbox{.}(2022b)]%
        {bian2022research}
\bibfield{author}{\bibinfo{person}{Haihong Bian}, \bibinfo{person}{Qian Wang}, \bibinfo{person}{Guozheng Xu}, {and} \bibinfo{person}{Xiu Zhao}.} \bibinfo{year}{2022}\natexlab{b}.
\newblock \showarticletitle{Research on short-term load forecasting based on accumulated temperature effect and improved temporal convolutional network}.
\newblock \bibinfo{journal}{\emph{Energy Reports}}  \bibinfo{volume}{8} (\bibinfo{year}{2022}), \bibinfo{pages}{1482--1491}.
\newblock


\bibitem[Bu et~al\mbox{.}(2023)]%
        {bu2023hybrid}
\bibfield{author}{\bibinfo{person}{Xiangya Bu}, \bibinfo{person}{Qiuwei Wu}, \bibinfo{person}{Bin Zhou}, {and} \bibinfo{person}{Canbing Li}.} \bibinfo{year}{2023}\natexlab{}.
\newblock \showarticletitle{Hybrid short-term load forecasting using {CGAN} with {CNN} and semi-supervised regression}.
\newblock \bibinfo{journal}{\emph{Applied Energy}}  \bibinfo{volume}{338} (\bibinfo{year}{2023}), \bibinfo{pages}{120920}.
\newblock


\bibitem[Bunn and Farmer(1985)]%
        {bunn1985comparative}
\bibfield{author}{\bibinfo{person}{D Bunn} {and} \bibinfo{person}{E~Dillon Farmer}.} \bibinfo{year}{1985}\natexlab{}.
\newblock \showarticletitle{Comparative models for electrical load forecasting}.
\newblock  (\bibinfo{year}{1985}).
\newblock


\bibitem[Cai et~al\mbox{.}(2020)]%
        {cai2020short}
\bibfield{author}{\bibinfo{person}{Qiuna Cai}, \bibinfo{person}{Binjie Yan}, \bibinfo{person}{Binghong Su}, \bibinfo{person}{Sijie Liu}, \bibinfo{person}{Mingxu Xiang}, \bibinfo{person}{Yakun Wen}, \bibinfo{person}{Yanyu Cheng}, {and} \bibinfo{person}{Nan Feng}.} \bibinfo{year}{2020}\natexlab{}.
\newblock \showarticletitle{Short-term load forecasting method based on deep neural network with sample weights}.
\newblock \bibinfo{journal}{\emph{International Transactions on Electrical Energy Systems}} \bibinfo{volume}{30}, \bibinfo{number}{5} (\bibinfo{year}{2020}), \bibinfo{pages}{e12340}.
\newblock


\bibitem[Chandola et~al\mbox{.}(2009)]%
        {chandola2009anomaly}
\bibfield{author}{\bibinfo{person}{Varun Chandola}, \bibinfo{person}{Arindam Banerjee}, {and} \bibinfo{person}{Vipin Kumar}.} \bibinfo{year}{2009}\natexlab{}.
\newblock \showarticletitle{Anomaly detection: A survey}.
\newblock \bibinfo{journal}{\emph{ACM computing surveys (CSUR)}} \bibinfo{volume}{41}, \bibinfo{number}{3} (\bibinfo{year}{2009}), \bibinfo{pages}{1--58}.
\newblock


\bibitem[Chang et~al\mbox{.}(2023)]%
        {chang2023llm4ts}
\bibfield{author}{\bibinfo{person}{Ching Chang}, \bibinfo{person}{Wen-Chih Peng}, {and} \bibinfo{person}{Tien-Fu Chen}.} \bibinfo{year}{2023}\natexlab{}.
\newblock \showarticletitle{Llm4ts: {T}wo-stage fine-tuning for time-series forecasting with pre-trained {LLMs}}.
\newblock \bibinfo{journal}{\emph{arXiv preprint arXiv:2308.08469}} (\bibinfo{year}{2023}).
\newblock


\bibitem[Chaturvedi et~al\mbox{.}(2015)]%
        {chaturvedi2015short}
\bibfield{author}{\bibinfo{person}{DK Chaturvedi}, \bibinfo{person}{AP Sinha}, {and} \bibinfo{person}{OP Malik}.} \bibinfo{year}{2015}\natexlab{}.
\newblock \showarticletitle{Short term load forecast using fuzzy logic and wavelet transform integrated generalized neural network}.
\newblock \bibinfo{journal}{\emph{International Journal of Electrical Power \& Energy Systems}}  \bibinfo{volume}{67} (\bibinfo{year}{2015}), \bibinfo{pages}{230--237}.
\newblock


\bibitem[Chen et~al\mbox{.}(2019)]%
        {chen2019day}
\bibfield{author}{\bibinfo{person}{Haiwen Chen}, \bibinfo{person}{Shouxiang Wang}, \bibinfo{person}{Shaomin Wang}, {and} \bibinfo{person}{Ye Li}.} \bibinfo{year}{2019}\natexlab{}.
\newblock \showarticletitle{Day-ahead aggregated load forecasting based on two-terminal sparse coding and deep neural network fusion}.
\newblock \bibinfo{journal}{\emph{Electric Power Systems Research}}  \bibinfo{volume}{177} (\bibinfo{year}{2019}), \bibinfo{pages}{105987}.
\newblock


\bibitem[Chen et~al\mbox{.}(2023)]%
        {chen2023research}
\bibfield{author}{\bibinfo{person}{Houhe Chen}, \bibinfo{person}{Mingyang Zhu}, \bibinfo{person}{Xiao Hu}, \bibinfo{person}{Jiarui Wang}, \bibinfo{person}{Yong Sun}, {and} \bibinfo{person}{Jinduo Yang}.} \bibinfo{year}{2023}\natexlab{}.
\newblock \showarticletitle{Research on short-term load forecasting of new-type power system based on {GCN-LSTM} considering multiple influencing factors}.
\newblock \bibinfo{journal}{\emph{Energy Reports}}  \bibinfo{volume}{9} (\bibinfo{year}{2023}), \bibinfo{pages}{1022--1031}.
\newblock


\bibitem[Chen et~al\mbox{.}(2018)]%
        {chen2018short}
\bibfield{author}{\bibinfo{person}{Kunjin Chen}, \bibinfo{person}{Kunlong Chen}, \bibinfo{person}{Qin Wang}, \bibinfo{person}{Ziyu He}, \bibinfo{person}{Jun Hu}, {and} \bibinfo{person}{Jinliang He}.} \bibinfo{year}{2018}\natexlab{}.
\newblock \showarticletitle{Short-term load forecasting with deep residual networks}.
\newblock \bibinfo{journal}{\emph{IEEE Transactions on Smart Grid}} \bibinfo{volume}{10}, \bibinfo{number}{4} (\bibinfo{year}{2018}), \bibinfo{pages}{3943--3952}.
\newblock


\bibitem[Chen et~al\mbox{.}(2021)]%
        {chen2021load}
\bibfield{author}{\bibinfo{person}{Zexi Chen}, \bibinfo{person}{Delong Zhang}, \bibinfo{person}{Haoran Jiang}, \bibinfo{person}{Longze Wang}, \bibinfo{person}{Yongcong Chen}, \bibinfo{person}{Yang Xiao}, \bibinfo{person}{Jinxin Liu}, \bibinfo{person}{Yan Zhang}, {and} \bibinfo{person}{Meicheng Li}.} \bibinfo{year}{2021}\natexlab{}.
\newblock \showarticletitle{Load forecasting based on {LSTM} neural network and applicable to loads of “replacement of coal with electricity”}.
\newblock \bibinfo{journal}{\emph{Journal of Electrical Engineering \& Technology}} \bibinfo{volume}{16}, \bibinfo{number}{5} (\bibinfo{year}{2021}), \bibinfo{pages}{2333--2342}.
\newblock


\bibitem[Cheng et~al\mbox{.}(2021)]%
        {cheng2021probabilistic}
\bibfield{author}{\bibinfo{person}{Lilin Cheng}, \bibinfo{person}{Haixiang Zang}, \bibinfo{person}{Yan Xu}, \bibinfo{person}{Zhinong Wei}, {and} \bibinfo{person}{Guoqiang Sun}.} \bibinfo{year}{2021}\natexlab{}.
\newblock \showarticletitle{Probabilistic residential load forecasting based on micrometeorological data and customer consumption pattern}.
\newblock \bibinfo{journal}{\emph{IEEE Transactions on Power systems}} \bibinfo{volume}{36}, \bibinfo{number}{4} (\bibinfo{year}{2021}), \bibinfo{pages}{3762--3775}.
\newblock


\bibitem[Cheung et~al\mbox{.}(2021)]%
        {cheung2021leveraging}
\bibfield{author}{\bibinfo{person}{Chung~Ming Cheung}, \bibinfo{person}{Sanmukh Kuppannagari}, \bibinfo{person}{Rajgopal Kannan}, {and} \bibinfo{person}{Viktor~K Prasanna}.} \bibinfo{year}{2021}\natexlab{}.
\newblock \showarticletitle{Leveraging spatial information in smart grids using {STGCN} for short-term load forecasting}. In \bibinfo{booktitle}{\emph{Proceedings of the 13th International Conference on Contemporary Computing (IC3'21)}}. \bibinfo{pages}{159--167}.
\newblock


\bibitem[Chiu et~al\mbox{.}(2023)]%
        {chiu2023hybrid}
\bibfield{author}{\bibinfo{person}{Ming-Chuan Chiu}, \bibinfo{person}{Hsin-Wei Hsu}, \bibinfo{person}{Ke-Sin Chen}, {and} \bibinfo{person}{Chih-Yuan Wen}.} \bibinfo{year}{2023}\natexlab{}.
\newblock \showarticletitle{A hybrid {CNN-GRU} based probabilistic model for load forecasting from individual household to commercial building}.
\newblock \bibinfo{journal}{\emph{Energy Reports}}  \bibinfo{volume}{9} (\bibinfo{year}{2023}), \bibinfo{pages}{94--105}.
\newblock


\bibitem[Choi et~al\mbox{.}(2018)]%
        {choi2018short}
\bibfield{author}{\bibinfo{person}{Hyungeun Choi}, \bibinfo{person}{Seunghyoung Ryu}, {and} \bibinfo{person}{Hongseok Kim}.} \bibinfo{year}{2018}\natexlab{}.
\newblock \showarticletitle{Short-term load forecasting based on {ResNet} and {LSTM}}. In \bibinfo{booktitle}{\emph{Proceedings of the 2018 IEEE International Conference on Communications, Control, and Computing Technologies for Smart Grids (SmartGridComm'18)}}. \bibinfo{pages}{1--6}.
\newblock


\bibitem[Dai et~al\mbox{.}(2023)]%
        {dai2023optimized}
\bibfield{author}{\bibinfo{person}{Yeming Dai}, \bibinfo{person}{Xinyu Yang}, {and} \bibinfo{person}{Mingming Leng}.} \bibinfo{year}{2023}\natexlab{}.
\newblock \showarticletitle{Optimized Seq2Seq model based on multiple methods for short-term power load forecasting}.
\newblock \bibinfo{journal}{\emph{Applied Soft Computing}}  \bibinfo{volume}{142} (\bibinfo{year}{2023}), \bibinfo{pages}{110335}.
\newblock


\bibitem[Das et~al\mbox{.}(2020)]%
        {das2020occupant}
\bibfield{author}{\bibinfo{person}{Anooshmita Das}, \bibinfo{person}{Masab~Khalid Annaqeeb}, \bibinfo{person}{Elie Azar}, \bibinfo{person}{Vojislav Novakovic}, {and} \bibinfo{person}{Mikkel~Baun Kj{\ae}rgaard}.} \bibinfo{year}{2020}\natexlab{}.
\newblock \showarticletitle{Occupant-centric miscellaneous electric loads prediction in buildings using state-of-the-art deep learning methods}.
\newblock \bibinfo{journal}{\emph{Applied Energy}}  \bibinfo{volume}{269} (\bibinfo{year}{2020}), \bibinfo{pages}{115135}.
\newblock


\bibitem[Dedinec et~al\mbox{.}(2016)]%
        {dedinec2016deep}
\bibfield{author}{\bibinfo{person}{Aleksandra Dedinec}, \bibinfo{person}{Sonja Filiposka}, \bibinfo{person}{Aleksandar Dedinec}, {and} \bibinfo{person}{Ljupco Kocarev}.} \bibinfo{year}{2016}\natexlab{}.
\newblock \showarticletitle{Deep belief network based electricity load forecasting: {A}n analysis of Macedonian case}.
\newblock \bibinfo{journal}{\emph{Energy}}  \bibinfo{volume}{115} (\bibinfo{year}{2016}), \bibinfo{pages}{1688--1700}.
\newblock


\bibitem[Deepanraj et~al\mbox{.}(2022)]%
        {deepanraj2022intelligent}
\bibfield{author}{\bibinfo{person}{B Deepanraj}, \bibinfo{person}{N Senthilkumar}, \bibinfo{person}{T Jarin}, \bibinfo{person}{Ali~Etem Gurel}, \bibinfo{person}{L~Syam Sundar}, {and} \bibinfo{person}{A~Vivek Anand}.} \bibinfo{year}{2022}\natexlab{}.
\newblock \showarticletitle{Intelligent wild geese algorithm with deep learning driven short term load forecasting for sustainable energy management in microgrids}.
\newblock \bibinfo{journal}{\emph{Sustainable Computing: Informatics and Systems}}  \bibinfo{volume}{36} (\bibinfo{year}{2022}), \bibinfo{pages}{100813}.
\newblock


\bibitem[Din and Marnerides(2017)]%
        {din2017short}
\bibfield{author}{\bibinfo{person}{Ghulam Mohi~Ud Din} {and} \bibinfo{person}{Angelos~K Marnerides}.} \bibinfo{year}{2017}\natexlab{}.
\newblock \showarticletitle{Short term power load forecasting using deep neural networks}. In \bibinfo{booktitle}{\emph{Proceedings of the 2017 International Conference on Computing, Networking and Communications (ICNC'17)}}. \bibinfo{pages}{594--598}.
\newblock


\bibitem[Dogra et~al\mbox{.}(2023)]%
        {dogra2023consumers}
\bibfield{author}{\bibinfo{person}{Atharvan Dogra}, \bibinfo{person}{Ashima Anand}, {and} \bibinfo{person}{Jatin Bedi}.} \bibinfo{year}{2023}\natexlab{}.
\newblock \showarticletitle{Consumers profiling based federated learning approach for energy load forecasting}.
\newblock \bibinfo{journal}{\emph{Sustainable Cities and Society}}  \bibinfo{volume}{98} (\bibinfo{year}{2023}), \bibinfo{pages}{104815}.
\newblock


\bibitem[Dong et~al\mbox{.}(2023)]%
        {dong2023parallel}
\bibfield{author}{\bibinfo{person}{Jizhe Dong}, \bibinfo{person}{Long Luo}, \bibinfo{person}{Yu Lu}, {and} \bibinfo{person}{Qi Zhang}.} \bibinfo{year}{2023}\natexlab{}.
\newblock \showarticletitle{A parallel short-term power load forecasting method considering high-level elastic loads}.
\newblock \bibinfo{journal}{\emph{IEEE Transactions on Instrumentation and Measurement}}  \bibinfo{volume}{72} (\bibinfo{year}{2023}), \bibinfo{pages}{1--10}.
\newblock


\bibitem[Dong et~al\mbox{.}(2017)]%
        {dong2017short}
\bibfield{author}{\bibinfo{person}{Xishuang Dong}, \bibinfo{person}{Lijun Qian}, {and} \bibinfo{person}{Lei Huang}.} \bibinfo{year}{2017}\natexlab{}.
\newblock \showarticletitle{Short-term load forecasting in smart grid: {A} combined {CNN} and K-means clustering approach}. In \bibinfo{booktitle}{\emph{Proceedings of the 2017 IEEE International Conference on Big Data and Smart Computing (BigComp'17)}}. \bibinfo{pages}{119--125}.
\newblock


\bibitem[Dong et~al\mbox{.}(2021a)]%
        {dong2021short}
\bibfield{author}{\bibinfo{person}{Yi Dong}, \bibinfo{person}{Zhen Dong}, \bibinfo{person}{Tianqiao Zhao}, \bibinfo{person}{Zhongguo Li}, {and} \bibinfo{person}{Zhengtao Ding}.} \bibinfo{year}{2021}\natexlab{a}.
\newblock \showarticletitle{Short term load forecasting with markovian switching distributed deep belief networks}.
\newblock \bibinfo{journal}{\emph{International Journal of Electrical Power \& Energy Systems}}  \bibinfo{volume}{130} (\bibinfo{year}{2021}), \bibinfo{pages}{106942}.
\newblock


\bibitem[Dong et~al\mbox{.}(2021b)]%
        {dong2021electrical}
\bibfield{author}{\bibinfo{person}{Yunxuan Dong}, \bibinfo{person}{Xuejiao Ma}, {and} \bibinfo{person}{Tonglin Fu}.} \bibinfo{year}{2021}\natexlab{b}.
\newblock \showarticletitle{Electrical load forecasting: A deep learning approach based on K-nearest neighbors}.
\newblock \bibinfo{journal}{\emph{Applied Soft Computing}}  \bibinfo{volume}{99} (\bibinfo{year}{2021}), \bibinfo{pages}{106900}.
\newblock


\bibitem[Dou et~al\mbox{.}(2018)]%
        {dou2018application}
\bibfield{author}{\bibinfo{person}{Yuchen Dou}, \bibinfo{person}{Xinman Zhang}, \bibinfo{person}{Zhihui Wu}, {and} \bibinfo{person}{Hang Zhang}.} \bibinfo{year}{2018}\natexlab{}.
\newblock \showarticletitle{Application of deep learning method in short-term load forecasting of characteristic enterprises}. In \bibinfo{booktitle}{\emph{Proceedings of the 2018 Artificial Intelligence and Cloud Computing Conference (AICCC'18)}}. \bibinfo{pages}{35--40}.
\newblock


\bibitem[Dudek(2016)]%
        {dudek2016neural}
\bibfield{author}{\bibinfo{person}{Grzegorz Dudek}.} \bibinfo{year}{2016}\natexlab{}.
\newblock \showarticletitle{Neural networks for pattern-based short-term load forecasting: {A} comparative study}.
\newblock \bibinfo{journal}{\emph{Neurocomputing}}  \bibinfo{volume}{205} (\bibinfo{year}{2016}), \bibinfo{pages}{64--74}.
\newblock


\bibitem[Eren and K{\"u}{\c{c}}{\"u}kdemiral(2024)]%
        {eren2024comprehensive}
\bibfield{author}{\bibinfo{person}{Yavuz Eren} {and} \bibinfo{person}{{\.I}brahim K{\"u}{\c{c}}{\"u}kdemiral}.} \bibinfo{year}{2024}\natexlab{}.
\newblock \showarticletitle{A comprehensive review on deep learning approaches for short-term load forecasting}.
\newblock \bibinfo{journal}{\emph{Renewable and Sustainable Energy Reviews}}  \bibinfo{volume}{189} (\bibinfo{year}{2024}), \bibinfo{pages}{114031}.
\newblock


\bibitem[Eskandari et~al\mbox{.}(2021)]%
        {eskandari2021convolutional}
\bibfield{author}{\bibinfo{person}{Hosein Eskandari}, \bibinfo{person}{Maryam Imani}, {and} \bibinfo{person}{Mohsen~Parsa Moghaddam}.} \bibinfo{year}{2021}\natexlab{}.
\newblock \showarticletitle{Convolutional and recurrent neural network based model for short-term load forecasting}.
\newblock \bibinfo{journal}{\emph{Electric Power Systems Research}}  \bibinfo{volume}{195} (\bibinfo{year}{2021}), \bibinfo{pages}{107173}.
\newblock


\bibitem[Fahiman et~al\mbox{.}(2019)]%
        {fahiman2019robust}
\bibfield{author}{\bibinfo{person}{Fateme Fahiman}, \bibinfo{person}{Sarah~M Erfani}, {and} \bibinfo{person}{Christopher Leckie}.} \bibinfo{year}{2019}\natexlab{}.
\newblock \showarticletitle{Robust and accurate short-term load forecasting: {A} cluster oriented ensemble learning approach}. In \bibinfo{booktitle}{\emph{Proceedings of the 2019 International Joint Conference on Neural Networks (IJCNN'19)}}. \bibinfo{pages}{1--8}.
\newblock


\bibitem[Fahiman et~al\mbox{.}(2017)]%
        {fahiman2017improving}
\bibfield{author}{\bibinfo{person}{Fateme Fahiman}, \bibinfo{person}{Sarah~M Erfani}, \bibinfo{person}{Sutharshan Rajasegarar}, \bibinfo{person}{Marimuthu Palaniswami}, {and} \bibinfo{person}{Christopher Leckie}.} \bibinfo{year}{2017}\natexlab{}.
\newblock \showarticletitle{Improving load forecasting based on deep learning and K-shape clustering}. In \bibinfo{booktitle}{\emph{Proceedings of the 2017 International Joint Conference on Neural Networks (IJCNN'17)}}. \bibinfo{pages}{4134--4141}.
\newblock


\bibitem[Fan et~al\mbox{.}(2020)]%
        {fan2020empirical}
\bibfield{author}{\bibinfo{person}{Chaodong Fan}, \bibinfo{person}{Changkun Ding}, \bibinfo{person}{Jinhua Zheng}, \bibinfo{person}{Leyi Xiao}, {and} \bibinfo{person}{Zhaoyang Ai}.} \bibinfo{year}{2020}\natexlab{}.
\newblock \showarticletitle{Empirical mode decomposition based multi-objective deep belief network for short-term power load forecasting}.
\newblock \bibinfo{journal}{\emph{Neurocomputing}}  \bibinfo{volume}{388} (\bibinfo{year}{2020}), \bibinfo{pages}{110--123}.
\newblock


\bibitem[Farid et~al\mbox{.}(2023)]%
        {farid2023conv1d}
\bibfield{author}{\bibinfo{person}{Karim~S Farid}, \bibinfo{person}{AA Ali}, \bibinfo{person}{Sameh~A Salem}, {and} \bibinfo{person}{Amr~E Mohamed}.} \bibinfo{year}{2023}\natexlab{}.
\newblock \showarticletitle{{CONV1D-GRU}: {A} hybrid model for short-term electrical load forecasting}. In \bibinfo{booktitle}{\emph{Proceedings of the 2023 International Telecommunications Conference (ITC-Egypt'23)}}. \bibinfo{pages}{281--286}.
\newblock


\bibitem[Faustine and Pereira(2022)]%
        {faustine2022fpseq2q}
\bibfield{author}{\bibinfo{person}{Anthony Faustine} {and} \bibinfo{person}{Lucas Pereira}.} \bibinfo{year}{2022}\natexlab{}.
\newblock \showarticletitle{FPSeq2Q: Fully parameterized sequence to quantile regression for net-load forecasting with uncertainty estimates}.
\newblock \bibinfo{journal}{\emph{IEEE Transactions on Smart Grid}} \bibinfo{volume}{13}, \bibinfo{number}{3} (\bibinfo{year}{2022}), \bibinfo{pages}{2440--2451}.
\newblock


\bibitem[Fekri et~al\mbox{.}(2021)]%
        {fekri2021deep}
\bibfield{author}{\bibinfo{person}{Mohammad~Navid Fekri}, \bibinfo{person}{Harsh Patel}, \bibinfo{person}{Katarina Grolinger}, {and} \bibinfo{person}{Vinay Sharma}.} \bibinfo{year}{2021}\natexlab{}.
\newblock \showarticletitle{Deep learning for load forecasting with smart meter data: Online adaptive recurrent neural network}.
\newblock \bibinfo{journal}{\emph{Applied Energy}}  \bibinfo{volume}{282} (\bibinfo{year}{2021}), \bibinfo{pages}{116177}.
\newblock


\bibitem[Feng et~al\mbox{.}(2019)]%
        {feng2019reinforced}
\bibfield{author}{\bibinfo{person}{Cong Feng}, \bibinfo{person}{Mucun Sun}, {and} \bibinfo{person}{Jie Zhang}.} \bibinfo{year}{2019}\natexlab{}.
\newblock \showarticletitle{Reinforced deterministic and probabilistic load forecasting via {Q}-learning dynamic model selection}.
\newblock \bibinfo{journal}{\emph{IEEE Transactions on Smart Grid}} \bibinfo{volume}{11}, \bibinfo{number}{2} (\bibinfo{year}{2019}), \bibinfo{pages}{1377--1386}.
\newblock


\bibitem[Feng et~al\mbox{.}(2024)]%
        {feng2024stgnet}
\bibfield{author}{\bibinfo{person}{Ding Feng}, \bibinfo{person}{Dengao Li}, \bibinfo{person}{Yu Zhou}, \bibinfo{person}{Jumin Zhao}, {and} \bibinfo{person}{Kenan Zhang}.} \bibinfo{year}{2024}\natexlab{}.
\newblock \showarticletitle{{STGNet}: Short-term residential load forecasting with spatial--temporal gated fusion network}.
\newblock \bibinfo{journal}{\emph{Energy Science \& Engineering}} \bibinfo{volume}{12}, \bibinfo{number}{3} (\bibinfo{year}{2024}), \bibinfo{pages}{541--560}.
\newblock


\bibitem[Gan et~al\mbox{.}(2017)]%
        {gan2017enhancing}
\bibfield{author}{\bibinfo{person}{Dahua Gan}, \bibinfo{person}{Yi Wang}, \bibinfo{person}{Ning Zhang}, {and} \bibinfo{person}{Wenjun Zhu}.} \bibinfo{year}{2017}\natexlab{}.
\newblock \showarticletitle{Enhancing short-term probabilistic residential load forecasting with quantile long--short-term memory}.
\newblock \bibinfo{journal}{\emph{The Journal of Engineering}} \bibinfo{volume}{2017}, \bibinfo{number}{14} (\bibinfo{year}{2017}), \bibinfo{pages}{2622--2627}.
\newblock


\bibitem[Ganjouri et~al\mbox{.}(2023)]%
        {ganjouri2023spatial}
\bibfield{author}{\bibinfo{person}{Mahtab Ganjouri}, \bibinfo{person}{Mazda Moattari}, \bibinfo{person}{Ahmad Forouzantabar}, {and} \bibinfo{person}{Mohammad Azadi}.} \bibinfo{year}{2023}\natexlab{}.
\newblock \showarticletitle{Spatial-temporal learning structure for short-term load forecasting}.
\newblock \bibinfo{journal}{\emph{IET Generation, Transmission \& Distribution}} \bibinfo{volume}{17}, \bibinfo{number}{2} (\bibinfo{year}{2023}), \bibinfo{pages}{427--437}.
\newblock


\bibitem[Gao et~al\mbox{.}(2023a)]%
        {gao2023adaptive}
\bibfield{author}{\bibinfo{person}{Jiaxin Gao}, \bibinfo{person}{Yuntian Chen}, \bibinfo{person}{Wenbo Hu}, {and} \bibinfo{person}{Dongxiao Zhang}.} \bibinfo{year}{2023}\natexlab{a}.
\newblock \showarticletitle{An adaptive deep-learning load forecasting framework by integrating transformer and domain knowledge}.
\newblock \bibinfo{journal}{\emph{Advances in Applied Energy}}  \bibinfo{volume}{10} (\bibinfo{year}{2023}), \bibinfo{pages}{100142}.
\newblock


\bibitem[Gao et~al\mbox{.}(2023b)]%
        {gao2023short}
\bibfield{author}{\bibinfo{person}{Qiang Gao}, \bibinfo{person}{Kaiyi Liu}, \bibinfo{person}{Kaibin Wu}, \bibinfo{person}{Menghan You}, {and} \bibinfo{person}{Hang Liu}.} \bibinfo{year}{2023}\natexlab{b}.
\newblock \showarticletitle{Short-term load forecasting for typical buildings based on {VMD-Informer-DMD} model}. In \bibinfo{booktitle}{\emph{Proceedings of the IEEE 2nd Industrial Electronics Society Annual On-Line Conference (ONCON'23)}}. \bibinfo{pages}{1--6}.
\newblock


\bibitem[Ghofrani et~al\mbox{.}(2015)]%
        {ghofrani2015hybrid}
\bibfield{author}{\bibinfo{person}{Mahmoud Ghofrani}, \bibinfo{person}{M Ghayekhloo}, \bibinfo{person}{A Arabali}, {and} \bibinfo{person}{A Ghayekhloo}.} \bibinfo{year}{2015}\natexlab{}.
\newblock \showarticletitle{A hybrid short-term load forecasting with a new input selection framework}.
\newblock \bibinfo{journal}{\emph{Energy}}  \bibinfo{volume}{81} (\bibinfo{year}{2015}), \bibinfo{pages}{777--786}.
\newblock


\bibitem[Giacomazzi et~al\mbox{.}(2023)]%
        {giacomazzi2023short}
\bibfield{author}{\bibinfo{person}{Elena Giacomazzi}, \bibinfo{person}{Felix Haag}, {and} \bibinfo{person}{Konstantin Hopf}.} \bibinfo{year}{2023}\natexlab{}.
\newblock \showarticletitle{Short-term electricity load forecasting using the temporal fusion transformer: {E}ffect of grid hierarchies and data sources}. In \bibinfo{booktitle}{\emph{Proceedings of the 14th ACM International Conference on Future Energy Systems (e-Energy'23)}}. \bibinfo{pages}{353--360}.
\newblock


\bibitem[Gruver et~al\mbox{.}(2024)]%
        {gruver2024large}
\bibfield{author}{\bibinfo{person}{Nate Gruver}, \bibinfo{person}{Marc Finzi}, \bibinfo{person}{Shikai Qiu}, {and} \bibinfo{person}{Andrew~G Wilson}.} \bibinfo{year}{2024}\natexlab{}.
\newblock \showarticletitle{Large language models are zero-shot time series forecasters}.
\newblock \bibinfo{journal}{\emph{Advances in Neural Information Processing Systems}}  \bibinfo{volume}{36} (\bibinfo{year}{2024}).
\newblock


\bibitem[Guo et~al\mbox{.}(2021)]%
        {guo2021short}
\bibfield{author}{\bibinfo{person}{Xifeng Guo}, \bibinfo{person}{Ye Gao}, \bibinfo{person}{Yupeng Li}, \bibinfo{person}{Di Zheng}, {and} \bibinfo{person}{Dan Shan}.} \bibinfo{year}{2021}\natexlab{}.
\newblock \showarticletitle{Short-term household load forecasting based on long-and short-term time-series network}.
\newblock \bibinfo{journal}{\emph{Energy Reports}}  \bibinfo{volume}{7} (\bibinfo{year}{2021}), \bibinfo{pages}{58--64}.
\newblock


\bibitem[Guo et~al\mbox{.}(2020)]%
        {guo2020short}
\bibfield{author}{\bibinfo{person}{Xifeng Guo}, \bibinfo{person}{Qiannan Zhao}, \bibinfo{person}{Di Zheng}, \bibinfo{person}{Yi Ning}, {and} \bibinfo{person}{Ye Gao}.} \bibinfo{year}{2020}\natexlab{}.
\newblock \showarticletitle{A short-term load forecasting model of multi-scale CNN-LSTM hybrid neural network considering the real-time electricity price}.
\newblock \bibinfo{journal}{\emph{Energy Reports}}  \bibinfo{volume}{6} (\bibinfo{year}{2020}), \bibinfo{pages}{1046--1053}.
\newblock


\bibitem[G{\"u}rses-Tran et~al\mbox{.}(2022)]%
        {gurses2022introducing}
\bibfield{author}{\bibinfo{person}{Gonca G{\"u}rses-Tran}, \bibinfo{person}{Tobias~Alexander K{\"o}rner}, {and} \bibinfo{person}{Antonello Monti}.} \bibinfo{year}{2022}\natexlab{}.
\newblock \showarticletitle{Introducing explainability in sequence-to-sequence learning for short-term load forecasting}.
\newblock \bibinfo{journal}{\emph{Electric Power Systems Research}}  \bibinfo{volume}{212} (\bibinfo{year}{2022}), \bibinfo{pages}{108366}.
\newblock


\bibitem[Hafeez et~al\mbox{.}(2020)]%
        {hafeez2020electric}
\bibfield{author}{\bibinfo{person}{Ghulam Hafeez}, \bibinfo{person}{Khurram~Saleem Alimgeer}, {and} \bibinfo{person}{Imran Khan}.} \bibinfo{year}{2020}\natexlab{}.
\newblock \showarticletitle{Electric load forecasting based on deep learning and optimized by heuristic algorithm in smart grid}.
\newblock \bibinfo{journal}{\emph{Applied Energy}}  \bibinfo{volume}{269} (\bibinfo{year}{2020}), \bibinfo{pages}{114915}.
\newblock


\bibitem[Haiyan et~al\mbox{.}(2020)]%
        {haiyan2020short}
\bibfield{author}{\bibinfo{person}{Wang Haiyan}, \bibinfo{person}{Lv Xinhang}, {and} \bibinfo{person}{Luo Xiaonan}.} \bibinfo{year}{2020}\natexlab{}.
\newblock \showarticletitle{Short-term load forecasting of power grid based on improved {WOA} optimized {LSTM}}. In \bibinfo{booktitle}{\emph{Proceedings of the 5th International Conference on Power and Renewable Energy (ICPRE'20)}}. \bibinfo{pages}{54--60}.
\newblock


\bibitem[Han et~al\mbox{.}(2023)]%
        {han2023research}
\bibfield{author}{\bibinfo{person}{Shuwei Han}, \bibinfo{person}{Huitong Ru}, \bibinfo{person}{Guangling Wang}, \bibinfo{person}{Xuezhi Fu}, \bibinfo{person}{Guoxu Zhou}, {and} \bibinfo{person}{Chengqiao Yang}.} \bibinfo{year}{2023}\natexlab{}.
\newblock \showarticletitle{Research on power load forecasting of {PCA-CNN-LSTM} based on sliding window}. In \bibinfo{booktitle}{\emph{Proceedings of the 3rd International Conference on New Energy and Power Engineering (ICNEPE'23)}}. \bibinfo{pages}{466--471}.
\newblock


\bibitem[Haque and Rahman(2022)]%
        {haque2022short}
\bibfield{author}{\bibinfo{person}{Ashraful Haque} {and} \bibinfo{person}{Saifur Rahman}.} \bibinfo{year}{2022}\natexlab{}.
\newblock \showarticletitle{Short-term electrical load forecasting through heuristic configuration of regularized deep neural network}.
\newblock \bibinfo{journal}{\emph{Applied Soft Computing}}  \bibinfo{volume}{122} (\bibinfo{year}{2022}), \bibinfo{pages}{108877}.
\newblock


\bibitem[Hayati and Shirvany(2007)]%
        {hayati2007artificial}
\bibfield{author}{\bibinfo{person}{Mohsen Hayati} {and} \bibinfo{person}{Yazdan Shirvany}.} \bibinfo{year}{2007}\natexlab{}.
\newblock \showarticletitle{Artificial neural network approach for short term load forecasting for Illam region}.
\newblock \bibinfo{journal}{\emph{World Academy of Science, Engineering and Technology}}  \bibinfo{volume}{28} (\bibinfo{year}{2007}), \bibinfo{pages}{280--284}.
\newblock


\bibitem[He et~al\mbox{.}(2021)]%
        {he2021per}
\bibfield{author}{\bibinfo{person}{Shengtao He}, \bibinfo{person}{Canbing Li}, \bibinfo{person}{Xubin Liu}, \bibinfo{person}{Xinyu Chen}, \bibinfo{person}{Mohammad Shahidehpour}, \bibinfo{person}{Tao Chen}, \bibinfo{person}{Bin Zhou}, {and} \bibinfo{person}{Qiuwei Wu}.} \bibinfo{year}{2021}\natexlab{}.
\newblock \showarticletitle{A per-unit curve rotated decoupling method for {CNN-TCN} based day-ahead load forecasting}.
\newblock \bibinfo{journal}{\emph{IET Generation, Transmission \& Distribution}} \bibinfo{volume}{15}, \bibinfo{number}{19} (\bibinfo{year}{2021}), \bibinfo{pages}{2773--2786}.
\newblock


\bibitem[He(2017)]%
        {he2017load}
\bibfield{author}{\bibinfo{person}{Wan He}.} \bibinfo{year}{2017}\natexlab{}.
\newblock \showarticletitle{Load forecasting via deep neural networks}.
\newblock \bibinfo{journal}{\emph{Procedia Computer Science}}  \bibinfo{volume}{122} (\bibinfo{year}{2017}), \bibinfo{pages}{308--314}.
\newblock


\bibitem[He et~al\mbox{.}(2017)]%
        {he2017short}
\bibfield{author}{\bibinfo{person}{Yusen He}, \bibinfo{person}{Jiahao Deng}, {and} \bibinfo{person}{Huajin Li}.} \bibinfo{year}{2017}\natexlab{}.
\newblock \showarticletitle{Short-term power load forecasting with deep belief network and copula models}. In \bibinfo{booktitle}{\emph{Proceedings of the 9th International Conference on Intelligent Human-Machine Systems and Cybernetics (IHMSC'17)}}. \bibinfo{pages}{191--194}.
\newblock


\bibitem[Hinton et~al\mbox{.}(2006)]%
        {hinton2006fast}
\bibfield{author}{\bibinfo{person}{Geoffrey~E Hinton}, \bibinfo{person}{Simon Osindero}, {and} \bibinfo{person}{Yee-Whye Teh}.} \bibinfo{year}{2006}\natexlab{}.
\newblock \showarticletitle{A fast learning algorithm for deep belief nets}.
\newblock \bibinfo{journal}{\emph{Neural computation}} \bibinfo{volume}{18}, \bibinfo{number}{7} (\bibinfo{year}{2006}), \bibinfo{pages}{1527--1554}.
\newblock


\bibitem[Hochreiter and Schmidhuber(1997)]%
        {hochreiter1997long}
\bibfield{author}{\bibinfo{person}{Sepp Hochreiter} {and} \bibinfo{person}{J{\"u}rgen Schmidhuber}.} \bibinfo{year}{1997}\natexlab{}.
\newblock \showarticletitle{Long short-term memory}.
\newblock \bibinfo{journal}{\emph{Neural computation}} \bibinfo{volume}{9}, \bibinfo{number}{8} (\bibinfo{year}{1997}), \bibinfo{pages}{1735--1780}.
\newblock


\bibitem[Hong and Fan(2016)]%
        {hong2016probabilistic}
\bibfield{author}{\bibinfo{person}{Tao Hong} {and} \bibinfo{person}{Shu Fan}.} \bibinfo{year}{2016}\natexlab{}.
\newblock \showarticletitle{Probabilistic electric load forecasting: {A} tutorial review}.
\newblock \bibinfo{journal}{\emph{International Journal of Forecasting}} \bibinfo{volume}{32}, \bibinfo{number}{3} (\bibinfo{year}{2016}), \bibinfo{pages}{914--938}.
\newblock


\bibitem[Hong and Chan(2023)]%
        {hong2023short}
\bibfield{author}{\bibinfo{person}{Ying-Yi Hong} {and} \bibinfo{person}{Yu-Hsuan Chan}.} \bibinfo{year}{2023}\natexlab{}.
\newblock \showarticletitle{Short-term electric load forecasting using particle swarm optimization-based convolutional neural network}.
\newblock \bibinfo{journal}{\emph{Engineering Applications of Artificial Intelligence}}  \bibinfo{volume}{126} (\bibinfo{year}{2023}), \bibinfo{pages}{106773}.
\newblock


\bibitem[Hong et~al\mbox{.}(2022)]%
        {hong2022week}
\bibfield{author}{\bibinfo{person}{Ying-Yi Hong}, \bibinfo{person}{Yu-Hsuan Chan}, \bibinfo{person}{Yung-Han Cheng}, \bibinfo{person}{Yih-Der Lee}, \bibinfo{person}{Jheng-Lun Jiang}, {and} \bibinfo{person}{Shen-Szu Wang}.} \bibinfo{year}{2022}\natexlab{}.
\newblock \showarticletitle{Week-ahead daily peak load forecasting using genetic algorithm-based hybrid convolutional neural network}.
\newblock \bibinfo{journal}{\emph{IET Generation, Transmission \& Distribution}} \bibinfo{volume}{16}, \bibinfo{number}{12} (\bibinfo{year}{2022}), \bibinfo{pages}{2416--2424}.
\newblock


\bibitem[Hoori et~al\mbox{.}(2019)]%
        {hoori2019electric}
\bibfield{author}{\bibinfo{person}{Ammar~O Hoori}, \bibinfo{person}{Ahmad Al~Kazzaz}, \bibinfo{person}{Rameez Khimani}, \bibinfo{person}{Yuichi Motai}, {and} \bibinfo{person}{Alex~J Aved}.} \bibinfo{year}{2019}\natexlab{}.
\newblock \showarticletitle{Electric load forecasting model using a multicolumn deep neural networks}.
\newblock \bibinfo{journal}{\emph{IEEE Transactions on Industrial Electronics}} \bibinfo{volume}{67}, \bibinfo{number}{8} (\bibinfo{year}{2019}), \bibinfo{pages}{6473--6482}.
\newblock


\bibitem[Hosein and Hosein(2017)]%
        {hosein2017load}
\bibfield{author}{\bibinfo{person}{Stefan Hosein} {and} \bibinfo{person}{Patrick Hosein}.} \bibinfo{year}{2017}\natexlab{}.
\newblock \showarticletitle{Load forecasting using deep neural networks}. In \bibinfo{booktitle}{\emph{Proceedings of the 2017 IEEE Power \& Energy Society Innovative Smart Grid Technologies Conference (ISGT'17)}}. \bibinfo{pages}{1--5}.
\newblock


\bibitem[Hossain and Mahmood(2020)]%
        {hossain2020short}
\bibfield{author}{\bibinfo{person}{Mohammad~Safayet Hossain} {and} \bibinfo{person}{Hisham Mahmood}.} \bibinfo{year}{2020}\natexlab{}.
\newblock \showarticletitle{Short-term load forecasting using an {LSTM} neural network}. In \bibinfo{booktitle}{\emph{Proceedings of the 2020 IEEE Power and Energy Conference at Illinois (PECI'20)}}. \bibinfo{pages}{1--6}.
\newblock


\bibitem[Hossen et~al\mbox{.}(2018)]%
        {hossen2018residential}
\bibfield{author}{\bibinfo{person}{Tareq Hossen}, \bibinfo{person}{Arun~Sukumaran Nair}, \bibinfo{person}{Radhakrishnan~Angamuthu Chinnathambi}, {and} \bibinfo{person}{Prakash Ranganathan}.} \bibinfo{year}{2018}\natexlab{}.
\newblock \showarticletitle{Residential load forecasting using deep neural networks ({DNN})}. In \bibinfo{booktitle}{\emph{Proceedings of the 2018 North American Power Symposium (NAPS'18)}}. \bibinfo{pages}{1--5}.
\newblock


\bibitem[Hossen et~al\mbox{.}(2017)]%
        {hossen2017short}
\bibfield{author}{\bibinfo{person}{Tareq Hossen}, \bibinfo{person}{Siby~Jose Plathottam}, \bibinfo{person}{Radha~Krishnan Angamuthu}, \bibinfo{person}{Prakash Ranganathan}, {and} \bibinfo{person}{Hossein Salehfar}.} \bibinfo{year}{2017}\natexlab{}.
\newblock \showarticletitle{Short-term load forecasting using deep neural networks ({DNN})}. In \bibinfo{booktitle}{\emph{Proceedings of the 2017 North American Power Symposium (NAPS'17)}}. \bibinfo{pages}{1--6}.
\newblock


\bibitem[Hou et~al\mbox{.}(2022)]%
        {hou2022review}
\bibfield{author}{\bibinfo{person}{Hui Hou}, \bibinfo{person}{Chao Liu}, \bibinfo{person}{Qing Wang}, \bibinfo{person}{Xixiu Wu}, \bibinfo{person}{Jinrui Tang}, \bibinfo{person}{Ying Shi}, {and} \bibinfo{person}{Changjun Xie}.} \bibinfo{year}{2022}\natexlab{}.
\newblock \showarticletitle{Review of load forecasting based on artificial intelligence methodologies, models, and challenges}.
\newblock \bibinfo{journal}{\emph{Electric Power Systems Research}}  \bibinfo{volume}{210} (\bibinfo{year}{2022}), \bibinfo{pages}{108067}.
\newblock


\bibitem[Hu et~al\mbox{.}(2022b)]%
        {hu2022development}
\bibfield{author}{\bibinfo{person}{Haowen Hu}, \bibinfo{person}{Xin Xia}, \bibinfo{person}{Yuanlin Luo}, \bibinfo{person}{Chu Zhang}, \bibinfo{person}{Muhammad~Shahzad Nazir}, {and} \bibinfo{person}{Tian Peng}.} \bibinfo{year}{2022}\natexlab{b}.
\newblock \showarticletitle{Development and application of an evolutionary deep learning framework of {LSTM} based on improved grasshopper optimization algorithm for short-term load forecasting}.
\newblock \bibinfo{journal}{\emph{Journal of Building Engineering}}  \bibinfo{volume}{57} (\bibinfo{year}{2022}), \bibinfo{pages}{104975}.
\newblock


\bibitem[Hu et~al\mbox{.}(2017)]%
        {hu2017short}
\bibfield{author}{\bibinfo{person}{Rui Hu}, \bibinfo{person}{Shiping Wen}, \bibinfo{person}{Zhigang Zeng}, {and} \bibinfo{person}{Tingwen Huang}.} \bibinfo{year}{2017}\natexlab{}.
\newblock \showarticletitle{A short-term power load forecasting model based on the generalized regression neural network with decreasing step fruit fly optimization algorithm}.
\newblock \bibinfo{journal}{\emph{Neurocomputing}}  \bibinfo{volume}{221} (\bibinfo{year}{2017}), \bibinfo{pages}{24--31}.
\newblock


\bibitem[Hu et~al\mbox{.}(2022c)]%
        {hu2022short}
\bibfield{author}{\bibinfo{person}{Weimin Hu}, \bibinfo{person}{Chao Yan}, \bibinfo{person}{Liping Fan}, \bibinfo{person}{Jie Yu}, \bibinfo{person}{Mei Yu}, \bibinfo{person}{Sheng Hua}, {and} \bibinfo{person}{Chonghao Yue}.} \bibinfo{year}{2022}\natexlab{c}.
\newblock \showarticletitle{Short-term power load forecasting based on {VMD-SSA-LSTM}}. In \bibinfo{booktitle}{\emph{Proceedings of the 2022 International Conference on High Performance Big Data and Intelligent Systems (HDIS'22)}}. \bibinfo{pages}{287--293}.
\newblock


\bibitem[Hu et~al\mbox{.}(2022a)]%
        {hu2022load}
\bibfield{author}{\bibinfo{person}{Xin Hu}, \bibinfo{person}{Keyi Li}, \bibinfo{person}{Jingfu Li}, \bibinfo{person}{Taotao Zhong}, \bibinfo{person}{Weinong Wu}, \bibinfo{person}{Xia Zhang}, {and} \bibinfo{person}{Wenjiang Feng}.} \bibinfo{year}{2022}\natexlab{a}.
\newblock \showarticletitle{Load forecasting model consisting of data mining based orthogonal greedy algorithm and long short-term memory network}.
\newblock \bibinfo{journal}{\emph{Energy Reports}}  \bibinfo{volume}{8} (\bibinfo{year}{2022}), \bibinfo{pages}{235--242}.
\newblock


\bibitem[Hua et~al\mbox{.}(2023)]%
        {hua2023ensemble}
\bibfield{author}{\bibinfo{person}{Heng Hua}, \bibinfo{person}{Mingping Liu}, \bibinfo{person}{Yuqin Li}, \bibinfo{person}{Suhui Deng}, {and} \bibinfo{person}{Qingnian Wang}.} \bibinfo{year}{2023}\natexlab{}.
\newblock \showarticletitle{An ensemble framework for short-term load forecasting based on parallel {CNN} and {GRU} with improved ResNet}.
\newblock \bibinfo{journal}{\emph{Electric Power Systems Research}}  \bibinfo{volume}{216} (\bibinfo{year}{2023}), \bibinfo{pages}{109057}.
\newblock


\bibitem[Huang et~al\mbox{.}(2021)]%
        {huang2021decomposition}
\bibfield{author}{\bibinfo{person}{Jiehui Huang}, \bibinfo{person}{Zhiwang Zhou}, \bibinfo{person}{Chunquan Li}, \bibinfo{person}{Zhiyuan Liao}, {and} \bibinfo{person}{Peter~X Liu}.} \bibinfo{year}{2021}\natexlab{}.
\newblock \showarticletitle{A decomposition-based multi-time dimension long short-term memory model for short-term electric load forecasting}.
\newblock \bibinfo{journal}{\emph{IET Generation, Transmission \& Distribution}} \bibinfo{volume}{15}, \bibinfo{number}{24} (\bibinfo{year}{2021}), \bibinfo{pages}{3459--3473}.
\newblock


\bibitem[Huang et~al\mbox{.}(2023)]%
        {huang2023gated}
\bibfield{author}{\bibinfo{person}{Nantian Huang}, \bibinfo{person}{Shengyuan Wang}, \bibinfo{person}{Rijun Wang}, \bibinfo{person}{Guowei Cai}, \bibinfo{person}{Yang Liu}, {and} \bibinfo{person}{Qianbin Dai}.} \bibinfo{year}{2023}\natexlab{}.
\newblock \showarticletitle{Gated spatial-temporal graph neural network based short-term load forecasting for wide-area multiple buses}.
\newblock \bibinfo{journal}{\emph{International Journal of Electrical Power \& Energy Systems}}  \bibinfo{volume}{145} (\bibinfo{year}{2023}), \bibinfo{pages}{108651}.
\newblock


\bibitem[Huang et~al\mbox{.}(2020)]%
        {huang2020improved}
\bibfield{author}{\bibinfo{person}{Qian Huang}, \bibinfo{person}{Jinghua Li}, {and} \bibinfo{person}{Mengshu Zhu}.} \bibinfo{year}{2020}\natexlab{}.
\newblock \showarticletitle{An improved convolutional neural network with load range discretization for probabilistic load forecasting}.
\newblock \bibinfo{journal}{\emph{Energy}}  \bibinfo{volume}{203} (\bibinfo{year}{2020}), \bibinfo{pages}{117902}.
\newblock


\bibitem[Huang et~al\mbox{.}(21)]%
        {huang2019survey}
\bibfield{author}{\bibinfo{person}{Rubing Huang}, \bibinfo{person}{Weifeng Sun}, \bibinfo{person}{Yinyin Xu}, \bibinfo{person}{Haibo Chen}, \bibinfo{person}{Dave Towey}, {and} \bibinfo{person}{Xin Xia}.} \bibinfo{year}{21}\natexlab{}.
\newblock \showarticletitle{A survey on adaptive random testing}.
\newblock \bibinfo{journal}{\emph{IEEE Transactions on Software Engineering}} \bibinfo{volume}{47}, \bibinfo{number}{10} (\bibinfo{year}{21}), \bibinfo{pages}{2052--2083}.
\newblock


\bibitem[Jalali et~al\mbox{.}(2021)]%
        {jalali2021novel}
\bibfield{author}{\bibinfo{person}{Seyed Mohammad~Jafar Jalali}, \bibinfo{person}{Sajad Ahmadian}, \bibinfo{person}{Abbas Khosravi}, \bibinfo{person}{Miadreza Shafie-khah}, \bibinfo{person}{Saeid Nahavandi}, {and} \bibinfo{person}{Jo{\~a}o~PS Catal{\~a}o}.} \bibinfo{year}{2021}\natexlab{}.
\newblock \showarticletitle{A novel evolutionary-based deep convolutional neural network model for intelligent load forecasting}.
\newblock \bibinfo{journal}{\emph{IEEE Transactions on Industrial Informatics}} \bibinfo{volume}{17}, \bibinfo{number}{12} (\bibinfo{year}{2021}), \bibinfo{pages}{8243--8253}.
\newblock


\bibitem[Jalali et~al\mbox{.}(2022)]%
        {jalali2022advanced}
\bibfield{author}{\bibinfo{person}{Seyed Mohammad~Jafar Jalali}, \bibinfo{person}{Parul Arora}, \bibinfo{person}{BK Panigrahi}, \bibinfo{person}{Abbas Khosravi}, \bibinfo{person}{Saeid Nahavandi}, \bibinfo{person}{Gerardo~J Os{\'o}rio}, {and} \bibinfo{person}{Jo{\~a}o~PS Catal{\~a}o}.} \bibinfo{year}{2022}\natexlab{}.
\newblock \showarticletitle{An advanced deep neuroevolution model for probabilistic load forecasting}.
\newblock \bibinfo{journal}{\emph{Electric Power Systems Research}}  \bibinfo{volume}{211} (\bibinfo{year}{2022}), \bibinfo{pages}{108351}.
\newblock


\bibitem[Javed et~al\mbox{.}(2022)]%
        {javed2022novel}
\bibfield{author}{\bibinfo{person}{Umar Javed}, \bibinfo{person}{Khalid Ijaz}, \bibinfo{person}{Muhammad Jawad}, \bibinfo{person}{Ikramullah Khosa}, \bibinfo{person}{Ejaz~Ahmad Ansari}, \bibinfo{person}{Khurram~Shabih Zaidi}, \bibinfo{person}{Muhammad~Nadeem Rafiq}, {and} \bibinfo{person}{Noman Shabbir}.} \bibinfo{year}{2022}\natexlab{}.
\newblock \showarticletitle{A novel short receptive field based dilated causal convolutional network integrated with Bidirectional LSTM for short-term load forecasting}.
\newblock \bibinfo{journal}{\emph{Expert Systems with Applications}}  \bibinfo{volume}{205} (\bibinfo{year}{2022}), \bibinfo{pages}{117689}.
\newblock


\bibitem[Jiang and Zheng(2022)]%
        {jiang2022deep}
\bibfield{author}{\bibinfo{person}{He Jiang} {and} \bibinfo{person}{Weihua Zheng}.} \bibinfo{year}{2022}\natexlab{}.
\newblock \showarticletitle{Deep learning with regularized robust long-and short-term memory network for probabilistic short-term load forecasting}.
\newblock \bibinfo{journal}{\emph{Journal of Forecasting}} \bibinfo{volume}{41}, \bibinfo{number}{6} (\bibinfo{year}{2022}), \bibinfo{pages}{1201--1216}.
\newblock


\bibitem[Jiao et~al\mbox{.}(2021)]%
        {jiao2021adaptive}
\bibfield{author}{\bibinfo{person}{Runhai Jiao}, \bibinfo{person}{Shuangkun Wang}, \bibinfo{person}{Tianle Zhang}, \bibinfo{person}{Hui Lu}, \bibinfo{person}{Hui He}, {and} \bibinfo{person}{Brij~B Gupta}.} \bibinfo{year}{2021}\natexlab{}.
\newblock \showarticletitle{Adaptive feature selection and construction for day-ahead load forecasting use deep learning method}.
\newblock \bibinfo{journal}{\emph{IEEE Transactions on Network and Service Management}} \bibinfo{volume}{18}, \bibinfo{number}{4} (\bibinfo{year}{2021}), \bibinfo{pages}{4019--4029}.
\newblock


\bibitem[Jin et~al\mbox{.}(2022)]%
        {jin2022short}
\bibfield{author}{\bibinfo{person}{Yuwei Jin}, \bibinfo{person}{Moses~Amoasi Acquah}, \bibinfo{person}{Mingyu Seo}, {and} \bibinfo{person}{Sekyung Han}.} \bibinfo{year}{2022}\natexlab{}.
\newblock \showarticletitle{Short-term electric load prediction using transfer learning with interval estimate adjustment}.
\newblock \bibinfo{journal}{\emph{Energy and Buildings}}  \bibinfo{volume}{258} (\bibinfo{year}{2022}), \bibinfo{pages}{111846}.
\newblock


\bibitem[Khan et~al\mbox{.}(2022)]%
        {khan2022efficient}
\bibfield{author}{\bibinfo{person}{Zulfiqar~Ahmad Khan}, \bibinfo{person}{Amin Ullah}, \bibinfo{person}{Ijaz~Ul Haq}, \bibinfo{person}{Mohamed Hamdy}, \bibinfo{person}{Gerardo~Maria Mauro}, \bibinfo{person}{Khan Muhammad}, \bibinfo{person}{Mohammad Hijji}, {and} \bibinfo{person}{Sung~Wook Baik}.} \bibinfo{year}{2022}\natexlab{}.
\newblock \showarticletitle{Efficient short-term electricity load forecasting for effective energy management}.
\newblock \bibinfo{journal}{\emph{Sustainable Energy Technologies and Assessments}}  \bibinfo{volume}{53} (\bibinfo{year}{2022}), \bibinfo{pages}{102337}.
\newblock


\bibitem[Kim et~al\mbox{.}(2019)]%
        {kim2019recurrent}
\bibfield{author}{\bibinfo{person}{Junhong Kim}, \bibinfo{person}{Jihoon Moon}, \bibinfo{person}{Eenjun Hwang}, {and} \bibinfo{person}{Pilsung Kang}.} \bibinfo{year}{2019}\natexlab{}.
\newblock \showarticletitle{Recurrent inception convolution neural network for multi short-term load forecasting}.
\newblock \bibinfo{journal}{\emph{Energy and buildings}}  \bibinfo{volume}{194} (\bibinfo{year}{2019}), \bibinfo{pages}{328--341}.
\newblock


\bibitem[Kim et~al\mbox{.}(2022)]%
        {kim2022short}
\bibfield{author}{\bibinfo{person}{Nakyoung Kim}, \bibinfo{person}{Hyunseo Park}, \bibinfo{person}{Joohyung Lee}, {and} \bibinfo{person}{Jun~Kyun Choi}.} \bibinfo{year}{2022}\natexlab{}.
\newblock \showarticletitle{Short-term electrical load forecasting with multidimensional feature extraction}.
\newblock \bibinfo{journal}{\emph{IEEE Transactions on Smart Grid}} \bibinfo{volume}{13}, \bibinfo{number}{4} (\bibinfo{year}{2022}), \bibinfo{pages}{2999--3013}.
\newblock


\bibitem[Kong et~al\mbox{.}(2017)]%
        {kong2017short}
\bibfield{author}{\bibinfo{person}{Weicong Kong}, \bibinfo{person}{Zhao~Yang Dong}, \bibinfo{person}{Youwei Jia}, \bibinfo{person}{David~J Hill}, \bibinfo{person}{Yan Xu}, {and} \bibinfo{person}{Yuan Zhang}.} \bibinfo{year}{2017}\natexlab{}.
\newblock \showarticletitle{Short-term residential load forecasting based on {LSTM} recurrent neural network}.
\newblock \bibinfo{journal}{\emph{IEEE Transactions on Smart Grid}} \bibinfo{volume}{10}, \bibinfo{number}{1} (\bibinfo{year}{2017}), \bibinfo{pages}{841--851}.
\newblock


\bibitem[Kong et~al\mbox{.}(2019)]%
        {kong2019improved}
\bibfield{author}{\bibinfo{person}{Xiangyu Kong}, \bibinfo{person}{Chuang Li}, \bibinfo{person}{Feng Zheng}, {and} \bibinfo{person}{Chengshan Wang}.} \bibinfo{year}{2019}\natexlab{}.
\newblock \showarticletitle{Improved deep belief network for short-term load forecasting considering demand-side management}.
\newblock \bibinfo{journal}{\emph{IEEE Transactions on Power Systems}} \bibinfo{volume}{35}, \bibinfo{number}{2} (\bibinfo{year}{2019}), \bibinfo{pages}{1531--1538}.
\newblock


\bibitem[Kouhi et~al\mbox{.}(2014)]%
        {kouhi2014new}
\bibfield{author}{\bibinfo{person}{Sajjad Kouhi}, \bibinfo{person}{Farshid Keynia}, {and} \bibinfo{person}{Sajad~Najafi Ravadanegh}.} \bibinfo{year}{2014}\natexlab{}.
\newblock \showarticletitle{A new short-term load forecast method based on neuro-evolutionary algorithm and chaotic feature selection}.
\newblock \bibinfo{journal}{\emph{International Journal of Electrical Power \& Energy Systems}}  \bibinfo{volume}{62} (\bibinfo{year}{2014}), \bibinfo{pages}{862--867}.
\newblock


\bibitem[Lai et~al\mbox{.}(2020)]%
        {lai2020load}
\bibfield{author}{\bibinfo{person}{Chun~Sing Lai}, \bibinfo{person}{Zhenyao Mo}, \bibinfo{person}{Ting Wang}, \bibinfo{person}{Haoliang Yuan}, \bibinfo{person}{Wing~WY Ng}, {and} \bibinfo{person}{Loi~Lei Lai}.} \bibinfo{year}{2020}\natexlab{}.
\newblock \showarticletitle{Load forecasting based on deep neural network and historical data augmentation}.
\newblock \bibinfo{journal}{\emph{IET Generation, Transmission \& Distribution}} \bibinfo{volume}{14}, \bibinfo{number}{24} (\bibinfo{year}{2020}), \bibinfo{pages}{5927--5934}.
\newblock


\bibitem[Langevin et~al\mbox{.}(2023)]%
        {langevin2023efficient}
\bibfield{author}{\bibinfo{person}{Antoine Langevin}, \bibinfo{person}{Mohamed Cheriet}, {and} \bibinfo{person}{Ghyslain Gagnon}.} \bibinfo{year}{2023}\natexlab{}.
\newblock \showarticletitle{Efficient deep generative model for short-term household load forecasting using non-intrusive load monitoring}.
\newblock \bibinfo{journal}{\emph{Sustainable Energy, Grids and Networks}}  \bibinfo{volume}{34} (\bibinfo{year}{2023}), \bibinfo{pages}{101006}.
\newblock


\bibitem[LeCun et~al\mbox{.}(2015)]%
        {lecun2015deep}
\bibfield{author}{\bibinfo{person}{Yann LeCun}, \bibinfo{person}{Yoshua Bengio}, {and} \bibinfo{person}{Geoffrey Hinton}.} \bibinfo{year}{2015}\natexlab{}.
\newblock \showarticletitle{Deep learning}.
\newblock \bibinfo{journal}{\emph{Nature}} \bibinfo{volume}{521}, \bibinfo{number}{7553} (\bibinfo{year}{2015}), \bibinfo{pages}{436--444}.
\newblock


\bibitem[Li et~al\mbox{.}(2023)]%
        {li2023short}
\bibfield{author}{\bibinfo{person}{Bin Li}, \bibinfo{person}{Yulu Mo}, \bibinfo{person}{Feng Gao}, {and} \bibinfo{person}{Xiaoqing Bai}.} \bibinfo{year}{2023}\natexlab{}.
\newblock \showarticletitle{Short-term probabilistic load forecasting method based on uncertainty estimation and deep learning model considering meteorological factors}.
\newblock \bibinfo{journal}{\emph{Electric Power Systems Research}}  \bibinfo{volume}{225} (\bibinfo{year}{2023}), \bibinfo{pages}{109804}.
\newblock


\bibitem[Li et~al\mbox{.}(2019)]%
        {li2019power}
\bibfield{author}{\bibinfo{person}{Chen Li}, \bibinfo{person}{Zhenyu Chen}, \bibinfo{person}{Jinbo Liu}, \bibinfo{person}{Dapeng Li}, \bibinfo{person}{Xingyu Gao}, \bibinfo{person}{Fangchun Di}, \bibinfo{person}{Lixin Li}, {and} \bibinfo{person}{Xiaohui Ji}.} \bibinfo{year}{2019}\natexlab{}.
\newblock \showarticletitle{Power load forecasting based on the combined model of {LSTM} and {XGBoost}}. In \bibinfo{booktitle}{\emph{Proceedings of the 2019 International Conference on Pattern Recognition and Artificial Intelligence (PRAI'19)}}. \bibinfo{pages}{46--51}.
\newblock


\bibitem[Li et~al\mbox{.}(2022a)]%
        {li2022short}
\bibfield{author}{\bibinfo{person}{Dan Li}, \bibinfo{person}{Guangfan Sun}, \bibinfo{person}{Shuwei Miao}, \bibinfo{person}{Yingzhong Gu}, \bibinfo{person}{Yuanhang Zhang}, {and} \bibinfo{person}{Shuai He}.} \bibinfo{year}{2022}\natexlab{a}.
\newblock \showarticletitle{A short-term electric load forecast method based on improved sequence-to-sequence GRU with adaptive temporal dependence}.
\newblock \bibinfo{journal}{\emph{International Journal of Electrical Power \& Energy Systems}}  \bibinfo{volume}{137} (\bibinfo{year}{2022}), \bibinfo{pages}{107627}.
\newblock


\bibitem[Li et~al\mbox{.}(2021)]%
        {li2021short}
\bibfield{author}{\bibinfo{person}{Lechen Li}, \bibinfo{person}{Christoph~J Meinrenken}, \bibinfo{person}{Vijay Modi}, {and} \bibinfo{person}{Patricia~J Culligan}.} \bibinfo{year}{2021}\natexlab{}.
\newblock \showarticletitle{Short-term apartment-level load forecasting using a modified neural network with selected auto-regressive features}.
\newblock \bibinfo{journal}{\emph{Applied Energy}}  \bibinfo{volume}{287} (\bibinfo{year}{2021}), \bibinfo{pages}{116509}.
\newblock


\bibitem[Li et~al\mbox{.}(2020b)]%
        {li2020effective}
\bibfield{author}{\bibinfo{person}{Ning Li}, \bibinfo{person}{Lu Wang}, \bibinfo{person}{Xinquan Li}, {and} \bibinfo{person}{Qing Zhu}.} \bibinfo{year}{2020}\natexlab{b}.
\newblock \showarticletitle{An effective deep learning neural network model for short-term load forecasting}.
\newblock \bibinfo{journal}{\emph{Concurrency and Computation: Practice and Experience}} \bibinfo{volume}{32}, \bibinfo{number}{7} (\bibinfo{year}{2020}), \bibinfo{pages}{e5595}.
\newblock


\bibitem[Li et~al\mbox{.}(2022b)]%
        {li2022electric}
\bibfield{author}{\bibinfo{person}{Xiaole Li}, \bibinfo{person}{Yiqin Wang}, \bibinfo{person}{Guibo Ma}, \bibinfo{person}{Xin Chen}, \bibinfo{person}{Qianxiang Shen}, {and} \bibinfo{person}{Bo Yang}.} \bibinfo{year}{2022}\natexlab{b}.
\newblock \showarticletitle{Electric load forecasting based on Long-Short-Term-Memory network via simplex optimizer during {COVID-19}}.
\newblock \bibinfo{journal}{\emph{Energy Reports}}  \bibinfo{volume}{8} (\bibinfo{year}{2022}), \bibinfo{pages}{1--12}.
\newblock


\bibitem[Li et~al\mbox{.}(2020a)]%
        {li2020deep}
\bibfield{author}{\bibinfo{person}{Zhuoling Li}, \bibinfo{person}{Yuanzheng Li}, \bibinfo{person}{Yun Liu}, \bibinfo{person}{Ping Wang}, \bibinfo{person}{Renzhi Lu}, {and} \bibinfo{person}{Hoay~Beng Gooi}.} \bibinfo{year}{2020}\natexlab{a}.
\newblock \showarticletitle{Deep learning based densely connected network for load forecasting}.
\newblock \bibinfo{journal}{\emph{IEEE Transactions on Power Systems}} \bibinfo{volume}{36}, \bibinfo{number}{4} (\bibinfo{year}{2020}), \bibinfo{pages}{2829--2840}.
\newblock


\bibitem[Lin et~al\mbox{.}(2022a)]%
        {lin2022short}
\bibfield{author}{\bibinfo{person}{Jun Lin}, \bibinfo{person}{Jin Ma}, \bibinfo{person}{Jianguo Zhu}, {and} \bibinfo{person}{Yu Cui}.} \bibinfo{year}{2022}\natexlab{a}.
\newblock \showarticletitle{Short-term load forecasting based on {LSTM} networks considering attention mechanism}.
\newblock \bibinfo{journal}{\emph{International Journal of Electrical Power \& Energy Systems}}  \bibinfo{volume}{137} (\bibinfo{year}{2022}), \bibinfo{pages}{107818}.
\newblock


\bibitem[Lin et~al\mbox{.}(2021)]%
        {lin2021spatial}
\bibfield{author}{\bibinfo{person}{Weixuan Lin}, \bibinfo{person}{Di Wu}, {and} \bibinfo{person}{Benoit Boulet}.} \bibinfo{year}{2021}\natexlab{}.
\newblock \showarticletitle{Spatial-temporal residential short-term load forecasting via graph neural networks}.
\newblock \bibinfo{journal}{\emph{IEEE Transactions on Smart Grid}} \bibinfo{volume}{12}, \bibinfo{number}{6} (\bibinfo{year}{2021}), \bibinfo{pages}{5373--5384}.
\newblock


\bibitem[Lin et~al\mbox{.}(2022b)]%
        {lin2022hybrid}
\bibfield{author}{\bibinfo{person}{Xin Lin}, \bibinfo{person}{Ramon Zamora}, \bibinfo{person}{Craig~A Baguley}, {and} \bibinfo{person}{Anurag~K Srivastava}.} \bibinfo{year}{2022}\natexlab{b}.
\newblock \showarticletitle{A hybrid short-term load forecasting approach for individual residential customer}.
\newblock \bibinfo{journal}{\emph{IEEE Transactions on Power Delivery}} \bibinfo{volume}{38}, \bibinfo{number}{1} (\bibinfo{year}{2022}), \bibinfo{pages}{26--37}.
\newblock


\bibitem[Liu et~al\mbox{.}(2022c)]%
        {liu2022power}
\bibfield{author}{\bibinfo{person}{Jiefeng Liu}, \bibinfo{person}{Zhenhao Zhang}, \bibinfo{person}{Xianhao Fan}, \bibinfo{person}{Yiyi Zhang}, \bibinfo{person}{Jiaqi Wang}, \bibinfo{person}{Ke Zhou}, \bibinfo{person}{Shuo Liang}, \bibinfo{person}{Xiaoyong Yu}, {and} \bibinfo{person}{Wei Zhang}.} \bibinfo{year}{2022}\natexlab{c}.
\newblock \showarticletitle{Power system load forecasting using mobility optimization and multi-task learning in {COVID-19}}.
\newblock \bibinfo{journal}{\emph{Applied Energy}}  \bibinfo{volume}{310} (\bibinfo{year}{2022}), \bibinfo{pages}{118303}.
\newblock


\bibitem[Liu et~al\mbox{.}(2022a)]%
        {liu2022short}
\bibfield{author}{\bibinfo{person}{Ronghui Liu}, \bibinfo{person}{Teng Chen}, \bibinfo{person}{Gaiping Sun}, \bibinfo{person}{SM Muyeen}, \bibinfo{person}{Shunfu Lin}, {and} \bibinfo{person}{Yang Mi}.} \bibinfo{year}{2022}\natexlab{a}.
\newblock \showarticletitle{Short-term probabilistic building load forecasting based on feature integrated artificial intelligent approach}.
\newblock \bibinfo{journal}{\emph{Electric Power Systems Research}}  \bibinfo{volume}{206} (\bibinfo{year}{2022}), \bibinfo{pages}{107802}.
\newblock


\bibitem[Liu et~al\mbox{.}(2022b)]%
        {liu2022image}
\bibfield{author}{\bibinfo{person}{Yanzhu Liu}, \bibinfo{person}{Shreya Dutta}, \bibinfo{person}{Adams Wai~Kin Kong}, {and} \bibinfo{person}{Chai~Kiat Yeo}.} \bibinfo{year}{2022}\natexlab{b}.
\newblock \showarticletitle{An image inpainting approach to short-term load forecasting}.
\newblock \bibinfo{journal}{\emph{IEEE Transactions on Power Systems}} \bibinfo{volume}{38}, \bibinfo{number}{1} (\bibinfo{year}{2022}), \bibinfo{pages}{177--187}.
\newblock


\bibitem[Lu et~al\mbox{.}(2019)]%
        {lu2019hybrid}
\bibfield{author}{\bibinfo{person}{Jixiang Lu}, \bibinfo{person}{Qipei Zhang}, \bibinfo{person}{Zhihong Yang}, {and} \bibinfo{person}{Mengfu Tu}.} \bibinfo{year}{2019}\natexlab{}.
\newblock \showarticletitle{A hybrid model based on convolutional neural network and long short-term memory for short-term load forecasting}. In \bibinfo{booktitle}{\emph{Proceedings of the 2019 IEEE Power \& Energy Society General Meeting (PESGM'19)}}. \bibinfo{pages}{1--5}.
\newblock


\bibitem[Lu et~al\mbox{.}(2022)]%
        {lu2022short}
\bibfield{author}{\bibinfo{person}{Yuting Lu}, \bibinfo{person}{Gaocai Wang}, {and} \bibinfo{person}{Shuqiang Huang}.} \bibinfo{year}{2022}\natexlab{}.
\newblock \showarticletitle{A short-term load forecasting model based on mixup and transfer learning}.
\newblock \bibinfo{journal}{\emph{Electric Power Systems Research}}  \bibinfo{volume}{207} (\bibinfo{year}{2022}), \bibinfo{pages}{107837}.
\newblock


\bibitem[Luo et~al\mbox{.}(2022)]%
        {luo2022ensemble}
\bibfield{author}{\bibinfo{person}{Hua Luo}, \bibinfo{person}{Haipeng Zhang}, {and} \bibinfo{person}{Jianzhou Wang}.} \bibinfo{year}{2022}\natexlab{}.
\newblock \showarticletitle{Ensemble power load forecasting based on competitive-inhibition selection strategy and deep learning}.
\newblock \bibinfo{journal}{\emph{Sustainable Energy Technologies and Assessments}}  \bibinfo{volume}{51} (\bibinfo{year}{2022}), \bibinfo{pages}{101940}.
\newblock


\bibitem[Lv et~al\mbox{.}(2021)]%
        {lv2021vmd}
\bibfield{author}{\bibinfo{person}{Lingling Lv}, \bibinfo{person}{Zongyu Wu}, \bibinfo{person}{Jinhua Zhang}, \bibinfo{person}{Lei Zhang}, \bibinfo{person}{Zhiyuan Tan}, {and} \bibinfo{person}{Zhihong Tian}.} \bibinfo{year}{2021}\natexlab{}.
\newblock \showarticletitle{A {VMD} and {LSTM} based hybrid model of load forecasting for power grid security}.
\newblock \bibinfo{journal}{\emph{IEEE Transactions on Industrial Informatics}} \bibinfo{volume}{18}, \bibinfo{number}{9} (\bibinfo{year}{2021}), \bibinfo{pages}{6474--6482}.
\newblock


\bibitem[Mathew et~al\mbox{.}(2021)]%
        {mathew2021emd}
\bibfield{author}{\bibinfo{person}{Jimson Mathew}, \bibinfo{person}{Ranjan~Kumar Behera}, {et~al\mbox{.}}} \bibinfo{year}{2021}\natexlab{}.
\newblock \showarticletitle{{EMD-Att-LSTM}: {A} data-driven strategy combined with deep learning for short-term load forecasting}.
\newblock \bibinfo{journal}{\emph{Journal of Modern Power Systems and Clean Energy}} \bibinfo{volume}{10}, \bibinfo{number}{5} (\bibinfo{year}{2021}), \bibinfo{pages}{1229--1240}.
\newblock


\bibitem[Meng et~al\mbox{.}(2022)]%
        {meng2022short}
\bibfield{author}{\bibinfo{person}{Zhaorui Meng}, \bibinfo{person}{Yanqi Xie}, {and} \bibinfo{person}{Jinhua Sun}.} \bibinfo{year}{2022}\natexlab{}.
\newblock \showarticletitle{Short-term load forecasting using neural attention model based on {EMD}}.
\newblock \bibinfo{journal}{\emph{Electrical Engineering}} (\bibinfo{year}{2022}), \bibinfo{pages}{1--10}.
\newblock


\bibitem[Morais et~al\mbox{.}(2023)]%
        {morais2023short}
\bibfield{author}{\bibinfo{person}{Lucas Barros~Scianni Morais}, \bibinfo{person}{Giancarlo Aquila}, \bibinfo{person}{Victor Augusto~Dur{\~a}es de Faria}, \bibinfo{person}{Luana Medeiros~Marangon Lima}, \bibinfo{person}{Jos{\'e} Wanderley~Marangon Lima}, {and} \bibinfo{person}{Anderson~Rodrigo de Queiroz}.} \bibinfo{year}{2023}\natexlab{}.
\newblock \showarticletitle{Short-term load forecasting using neural networks and global climate models: An application to a large-scale electrical power system}.
\newblock \bibinfo{journal}{\emph{Applied Energy}}  \bibinfo{volume}{348} (\bibinfo{year}{2023}), \bibinfo{pages}{121439}.
\newblock


\bibitem[Mounir et~al\mbox{.}(2023)]%
        {mounir2023short}
\bibfield{author}{\bibinfo{person}{Nada Mounir}, \bibinfo{person}{Hamid Ouadi}, {and} \bibinfo{person}{Ismael Jrhilifa}.} \bibinfo{year}{2023}\natexlab{}.
\newblock \showarticletitle{Short-term electric load forecasting using an {EMD-BI-LSTM} approach for smart grid energy management system}.
\newblock \bibinfo{journal}{\emph{Energy and Buildings}}  \bibinfo{volume}{288} (\bibinfo{year}{2023}), \bibinfo{pages}{113022}.
\newblock


\bibitem[Mughees et~al\mbox{.}(2021)]%
        {mughees2021deep}
\bibfield{author}{\bibinfo{person}{Neelam Mughees}, \bibinfo{person}{Syed~Ali Mohsin}, \bibinfo{person}{Abdullah Mughees}, {and} \bibinfo{person}{Anam Mughees}.} \bibinfo{year}{2021}\natexlab{}.
\newblock \showarticletitle{Deep sequence to sequence {Bi-LSTM} neural networks for day-ahead peak load forecasting}.
\newblock \bibinfo{journal}{\emph{Expert Systems with Applications}}  \bibinfo{volume}{175} (\bibinfo{year}{2021}), \bibinfo{pages}{114844}.
\newblock


\bibitem[Nawar et~al\mbox{.}(2023)]%
        {nawar2023transfer}
\bibfield{author}{\bibinfo{person}{Menna Nawar}, \bibinfo{person}{Moustafa Shomer}, \bibinfo{person}{Samy Faddel}, {and} \bibinfo{person}{Huangjie Gong}.} \bibinfo{year}{2023}\natexlab{}.
\newblock \showarticletitle{Transfer learning in deep learning models for building load forecasting: {C}ase of limited data}. In \bibinfo{booktitle}{\emph{Proceedings of the IEEE SoutheastCon 2023}}. \bibinfo{pages}{532--538}.
\newblock


\bibitem[Neeraj et~al\mbox{.}(2021)]%
        {neeraj2021long}
\bibfield{author}{\bibinfo{person}{Neeraj Neeraj}, \bibinfo{person}{Jimson Mathew}, \bibinfo{person}{Mayank Agarwal}, {and} \bibinfo{person}{Ranjan~Kumar Behera}.} \bibinfo{year}{2021}\natexlab{}.
\newblock \showarticletitle{Long short-term memory-singular spectrum analysis-based model for electric load forecasting}.
\newblock \bibinfo{journal}{\emph{Electrical Engineering}} \bibinfo{volume}{103}, \bibinfo{number}{2} (\bibinfo{year}{2021}), \bibinfo{pages}{1067--1082}.
\newblock


\bibitem[Nie et~al\mbox{.}(2020)]%
        {nie2020novel}
\bibfield{author}{\bibinfo{person}{Ying Nie}, \bibinfo{person}{Ping Jiang}, {and} \bibinfo{person}{Haipeng Zhang}.} \bibinfo{year}{2020}\natexlab{}.
\newblock \showarticletitle{A novel hybrid model based on combined preprocessing method and advanced optimization algorithm for power load forecasting}.
\newblock \bibinfo{journal}{\emph{Applied Soft Computing}}  \bibinfo{volume}{97} (\bibinfo{year}{2020}), \bibinfo{pages}{106809}.
\newblock


\bibitem[Niu et~al\mbox{.}(2016)]%
        {niu2016innovative}
\bibfield{author}{\bibinfo{person}{Mingfei Niu}, \bibinfo{person}{Shaolong Sun}, \bibinfo{person}{Jing Wu}, \bibinfo{person}{Lean Yu}, {and} \bibinfo{person}{Jianzhou Wang}.} \bibinfo{year}{2016}\natexlab{}.
\newblock \showarticletitle{An innovative integrated model using the singular spectrum analysis and nonlinear multi-layer perceptron network optimized by hybrid intelligent algorithm for short-term load forecasting}.
\newblock \bibinfo{journal}{\emph{Applied Mathematical Modelling}} \bibinfo{volume}{40}, \bibinfo{number}{5-6} (\bibinfo{year}{2016}), \bibinfo{pages}{4079--4093}.
\newblock


\bibitem[Ozer et~al\mbox{.}(2021)]%
        {ozer2021combined}
\bibfield{author}{\bibinfo{person}{Ilyas Ozer}, \bibinfo{person}{Serhat~Berat Efe}, {and} \bibinfo{person}{Harun Ozbay}.} \bibinfo{year}{2021}\natexlab{}.
\newblock \showarticletitle{A combined deep learning application for short term load forecasting}.
\newblock \bibinfo{journal}{\emph{Alexandria Engineering Journal}} \bibinfo{volume}{60}, \bibinfo{number}{4} (\bibinfo{year}{2021}), \bibinfo{pages}{3807--3818}.
\newblock


\bibitem[Park et~al\mbox{.}(1991)]%
        {park1991electric}
\bibfield{author}{\bibinfo{person}{Dong~C Park}, \bibinfo{person}{MA El-Sharkawi}, \bibinfo{person}{RJ Marks}, \bibinfo{person}{LE Atlas}, {and} \bibinfo{person}{MJ Damborg}.} \bibinfo{year}{1991}\natexlab{}.
\newblock \showarticletitle{Electric load forecasting using an artificial neural network}.
\newblock \bibinfo{journal}{\emph{IEEE Transactions on Power Systems}} \bibinfo{volume}{6}, \bibinfo{number}{2} (\bibinfo{year}{1991}), \bibinfo{pages}{442--449}.
\newblock


\bibitem[Pe{\l}ka and Dudek(2020)]%
        {pelka2020pattern}
\bibfield{author}{\bibinfo{person}{Pawe{\l} Pe{\l}ka} {and} \bibinfo{person}{Grzegorz Dudek}.} \bibinfo{year}{2020}\natexlab{}.
\newblock \showarticletitle{Pattern-based long short-term memory for mid-term electrical load forecasting}. In \bibinfo{booktitle}{\emph{Proceedings of the 2020 International Joint Conference on Neural Networks (IJCNN'20)}}. \bibinfo{pages}{1--8}.
\newblock


\bibitem[Qin et~al\mbox{.}(2022)]%
        {qin2022multi}
\bibfield{author}{\bibinfo{person}{Jiaqi Qin}, \bibinfo{person}{Yi Zhang}, \bibinfo{person}{Shixiong Fan}, \bibinfo{person}{Xiaonan Hu}, \bibinfo{person}{Yongqiang Huang}, \bibinfo{person}{Zexin Lu}, {and} \bibinfo{person}{Yan Liu}.} \bibinfo{year}{2022}\natexlab{}.
\newblock \showarticletitle{Multi-task short-term reactive and active load forecasting method based on attention-LSTM model}.
\newblock \bibinfo{journal}{\emph{International Journal of Electrical Power \& Energy Systems}}  \bibinfo{volume}{135} (\bibinfo{year}{2022}), \bibinfo{pages}{107517}.
\newblock


\bibitem[Rafati et~al\mbox{.}(2020)]%
        {rafati2020efficient}
\bibfield{author}{\bibinfo{person}{Amir Rafati}, \bibinfo{person}{Mahmood Joorabian}, {and} \bibinfo{person}{Elaheh Mashhour}.} \bibinfo{year}{2020}\natexlab{}.
\newblock \showarticletitle{An efficient hour-ahead electrical load forecasting method based on innovative features}.
\newblock \bibinfo{journal}{\emph{Energy}}  \bibinfo{volume}{201} (\bibinfo{year}{2020}), \bibinfo{pages}{117511}.
\newblock


\bibitem[Ran et~al\mbox{.}(2023)]%
        {ran2023short}
\bibfield{author}{\bibinfo{person}{Peng Ran}, \bibinfo{person}{Kun Dong}, \bibinfo{person}{Xu Liu}, {and} \bibinfo{person}{Jing Wang}.} \bibinfo{year}{2023}\natexlab{}.
\newblock \showarticletitle{Short-term load forecasting based on CEEMDAN and Transformer}.
\newblock \bibinfo{journal}{\emph{Electric Power Systems Research}}  \bibinfo{volume}{214} (\bibinfo{year}{2023}), \bibinfo{pages}{108885}.
\newblock


\bibitem[Raza et~al\mbox{.}(2017)]%
        {raza2017intelligent}
\bibfield{author}{\bibinfo{person}{Muhammad~Qamar Raza}, \bibinfo{person}{Mithulananthan Nadarajah}, \bibinfo{person}{Duong~Quoc Hung}, {and} \bibinfo{person}{Zuhairi Baharudin}.} \bibinfo{year}{2017}\natexlab{}.
\newblock \showarticletitle{An intelligent hybrid short-term load forecasting model for smart power grids}.
\newblock \bibinfo{journal}{\emph{Sustainable Cities and Society}}  \bibinfo{volume}{31} (\bibinfo{year}{2017}), \bibinfo{pages}{264--275}.
\newblock


\bibitem[Rumelhart et~al\mbox{.}(1986)]%
        {rumelhart1986learning}
\bibfield{author}{\bibinfo{person}{David~E Rumelhart}, \bibinfo{person}{Geoffrey~E Hinton}, {and} \bibinfo{person}{Ronald~J Williams}.} \bibinfo{year}{1986}\natexlab{}.
\newblock \showarticletitle{Learning representations by back-propagating errors}.
\newblock \bibinfo{journal}{\emph{Nature}} \bibinfo{volume}{323}, \bibinfo{number}{6088} (\bibinfo{year}{1986}), \bibinfo{pages}{533--536}.
\newblock


\bibitem[Sadaei et~al\mbox{.}(2019)]%
        {sadaei2019short}
\bibfield{author}{\bibinfo{person}{Hossein~Javedani Sadaei}, \bibinfo{person}{Petr{\^o}nio C{\^a}ndido de~Lima e Silva}, \bibinfo{person}{Frederico~Gadelha Guimaraes}, {and} \bibinfo{person}{Muhammad~Hisyam Lee}.} \bibinfo{year}{2019}\natexlab{}.
\newblock \showarticletitle{Short-term load forecasting by using a combined method of convolutional neural networks and fuzzy time series}.
\newblock \bibinfo{journal}{\emph{Energy}}  \bibinfo{volume}{175} (\bibinfo{year}{2019}), \bibinfo{pages}{365--377}.
\newblock


\bibitem[Sakib et~al\mbox{.}(2021)]%
        {sakib2021data}
\bibfield{author}{\bibinfo{person}{Shadman Sakib}, \bibinfo{person}{Khan~Md Hasib}, \bibinfo{person}{Ihtyaz~Kader Tasawar}, \bibinfo{person}{Abyaz~Kader Tanzeem}, \bibinfo{person}{Md~Fahim Arefin}, \bibinfo{person}{Saharul Islam}, {and} \bibinfo{person}{Mohammad~Shafiul Alam}.} \bibinfo{year}{2021}\natexlab{}.
\newblock \showarticletitle{A data-driven hybrid optimization based deep network model for short-term residential load forecasting}. In \bibinfo{booktitle}{\emph{Proceedings of the IEEE 12th Annual Information Technology, Electronics and Mobile Communication Conference (IEMCON'21)}}. \bibinfo{pages}{0187--0193}.
\newblock


\bibitem[Santos et~al\mbox{.}(2023)]%
        {santos2023deep}
\bibfield{author}{\bibinfo{person}{Miguel~L{\'o}pez Santos}, \bibinfo{person}{Sa{\'u}l~D{\'\i}az Garc{\'\i}a}, \bibinfo{person}{Xela Garc{\'\i}a-Santiago}, \bibinfo{person}{Ana Ogando-Mart{\'\i}nez}, \bibinfo{person}{Fernando~Echevarr{\'\i}a Camarero}, \bibinfo{person}{Gonzalo~Bl{\'a}zquez Gil}, {and} \bibinfo{person}{Pablo~Carrasco Ortega}.} \bibinfo{year}{2023}\natexlab{}.
\newblock \showarticletitle{Deep learning and transfer learning techniques applied to short-term load forecasting of data-poor buildings in local energy communities}.
\newblock \bibinfo{journal}{\emph{Energy and Buildings}}  \bibinfo{volume}{292} (\bibinfo{year}{2023}), \bibinfo{pages}{113164}.
\newblock


\bibitem[Schmidhuber(2015)]%
        {schmidhuber2015deep}
\bibfield{author}{\bibinfo{person}{J{\"u}rgen Schmidhuber}.} \bibinfo{year}{2015}\natexlab{}.
\newblock \showarticletitle{Deep learning in neural networks: {A}n overview}.
\newblock \bibinfo{journal}{\emph{Neural networks}}  \bibinfo{volume}{61} (\bibinfo{year}{2015}), \bibinfo{pages}{85--117}.
\newblock


\bibitem[Sekhar and Dahiya(2023)]%
        {sekhar2023robust}
\bibfield{author}{\bibinfo{person}{Charan Sekhar} {and} \bibinfo{person}{Ratna Dahiya}.} \bibinfo{year}{2023}\natexlab{}.
\newblock \showarticletitle{Robust framework based on hybrid deep learning approach for short term load forecasting of building electricity demand}.
\newblock \bibinfo{journal}{\emph{Energy}}  \bibinfo{volume}{268} (\bibinfo{year}{2023}), \bibinfo{pages}{126660}.
\newblock


\bibitem[Shaqour et~al\mbox{.}(2022)]%
        {shaqour2022electrical}
\bibfield{author}{\bibinfo{person}{Ayas Shaqour}, \bibinfo{person}{Tetsushi Ono}, \bibinfo{person}{Aya Hagishima}, {and} \bibinfo{person}{Hooman Farzaneh}.} \bibinfo{year}{2022}\natexlab{}.
\newblock \showarticletitle{Electrical demand aggregation effects on the performance of deep learning-based short-term load forecasting of a residential building}.
\newblock \bibinfo{journal}{\emph{Energy and AI}}  \bibinfo{volume}{8} (\bibinfo{year}{2022}), \bibinfo{pages}{100141}.
\newblock


\bibitem[Sharma and Jain(2022)]%
        {sharma2022novel}
\bibfield{author}{\bibinfo{person}{Abhishek Sharma} {and} \bibinfo{person}{Sachin~Kumar Jain}.} \bibinfo{year}{2022}\natexlab{}.
\newblock \showarticletitle{A novel seasonal segmentation approach for day-ahead load forecasting}.
\newblock \bibinfo{journal}{\emph{Energy}}  \bibinfo{volume}{257} (\bibinfo{year}{2022}), \bibinfo{pages}{124752}.
\newblock


\bibitem[Shi et~al\mbox{.}(2023)]%
        {shi2023short}
\bibfield{author}{\bibinfo{person}{Huifeng Shi}, \bibinfo{person}{Kai Miao}, {and} \bibinfo{person}{Xiaochen Ren}.} \bibinfo{year}{2023}\natexlab{}.
\newblock \showarticletitle{Short-term load forecasting based on {CNN-BiLSTM} with Bayesian optimization and attention mechanism}.
\newblock \bibinfo{journal}{\emph{Concurrency and Computation: Practice and Experience}} \bibinfo{volume}{35}, \bibinfo{number}{17} (\bibinfo{year}{2023}), \bibinfo{pages}{e6676}.
\newblock


\bibitem[Singh and Dwivedi(2018)]%
        {singh2018integration}
\bibfield{author}{\bibinfo{person}{Priyanka Singh} {and} \bibinfo{person}{Pragya Dwivedi}.} \bibinfo{year}{2018}\natexlab{}.
\newblock \showarticletitle{Integration of new evolutionary approach with artificial neural network for solving short term load forecast problem}.
\newblock \bibinfo{journal}{\emph{Applied energy}}  \bibinfo{volume}{217} (\bibinfo{year}{2018}), \bibinfo{pages}{537--549}.
\newblock


\bibitem[Su et~al\mbox{.}(2023)]%
        {su2023residential}
\bibfield{author}{\bibinfo{person}{Yongxin Su}, \bibinfo{person}{Qiyao He}, \bibinfo{person}{Jie Chen}, {and} \bibinfo{person}{Mao Tan}.} \bibinfo{year}{2023}\natexlab{}.
\newblock \showarticletitle{A residential load forecasting method for multi-attribute adversarial learning considering multi-source uncertainties}.
\newblock \bibinfo{journal}{\emph{International Journal of Electrical Power \& Energy Systems}}  \bibinfo{volume}{154} (\bibinfo{year}{2023}), \bibinfo{pages}{109421}.
\newblock


\bibitem[Subbiah and Chinnappan(2022)]%
        {subbiah2022deep}
\bibfield{author}{\bibinfo{person}{Siva~Sankari Subbiah} {and} \bibinfo{person}{Jayakumar Chinnappan}.} \bibinfo{year}{2022}\natexlab{}.
\newblock \showarticletitle{Deep learning based short term load forecasting with hybrid feature selection}.
\newblock \bibinfo{journal}{\emph{Electric Power Systems Research}}  \bibinfo{volume}{210} (\bibinfo{year}{2022}), \bibinfo{pages}{108065}.
\newblock


\bibitem[Sun et~al\mbox{.}(2020)]%
        {sun2020short}
\bibfield{author}{\bibinfo{person}{Gaiping Sun}, \bibinfo{person}{Chuanwen Jiang}, \bibinfo{person}{Xu Wang}, {and} \bibinfo{person}{Xiu Yang}.} \bibinfo{year}{2020}\natexlab{}.
\newblock \showarticletitle{Short-term building load forecast based on a data-mining feature selection and {LSTM-RNN} method}.
\newblock \bibinfo{journal}{\emph{IEEJ Transactions on Electrical and Electronic Engineering}} \bibinfo{volume}{15}, \bibinfo{number}{7} (\bibinfo{year}{2020}), \bibinfo{pages}{1002--1010}.
\newblock


\bibitem[Sun et~al\mbox{.}(2018)]%
        {sun2018short}
\bibfield{author}{\bibinfo{person}{Hongbin Sun}, \bibinfo{person}{Xin Pan}, {and} \bibinfo{person}{Changxin Meng}.} \bibinfo{year}{2018}\natexlab{}.
\newblock \showarticletitle{A short-term power load prediction algorithm of based on power load factor deep cluster neural network}.
\newblock \bibinfo{journal}{\emph{Wireless Personal Communications}}  \bibinfo{volume}{102} (\bibinfo{year}{2018}), \bibinfo{pages}{1073--1084}.
\newblock


\bibitem[Tan et~al\mbox{.}(2022)]%
        {tan2022multi}
\bibfield{author}{\bibinfo{person}{Mao Tan}, \bibinfo{person}{Chenglin Hu}, \bibinfo{person}{Jie Chen}, \bibinfo{person}{Ling Wang}, {and} \bibinfo{person}{Zhengmao Li}.} \bibinfo{year}{2022}\natexlab{}.
\newblock \showarticletitle{Multi-node load forecasting based on multi-task learning with modal feature extraction}.
\newblock \bibinfo{journal}{\emph{Engineering Applications of Artificial Intelligence}}  \bibinfo{volume}{112} (\bibinfo{year}{2022}), \bibinfo{pages}{104856}.
\newblock


\bibitem[Tang et~al\mbox{.}(2019b)]%
        {tang2019ensemble}
\bibfield{author}{\bibinfo{person}{Lingling Tang}, \bibinfo{person}{Yulin Yi}, {and} \bibinfo{person}{Yuexing Peng}.} \bibinfo{year}{2019}\natexlab{b}.
\newblock \showarticletitle{An ensemble deep learning model for short-term load forecasting based on {ARIMA} and {LSTM}}. In \bibinfo{booktitle}{\emph{Proceedings of the 2019 IEEE International Conference on Communications, Control, and Computing Technologies for Smart Grids (SmartGridComm'19)}}. \bibinfo{pages}{1--6}.
\newblock


\bibitem[Tang et~al\mbox{.}(2022)]%
        {tang2022short}
\bibfield{author}{\bibinfo{person}{Xianlun Tang}, \bibinfo{person}{Hongxu Chen}, \bibinfo{person}{Wenhao Xiang}, \bibinfo{person}{Jingming Yang}, {and} \bibinfo{person}{Mi Zou}.} \bibinfo{year}{2022}\natexlab{}.
\newblock \showarticletitle{Short-term load forecasting using channel and temporal attention based temporal convolutional network}.
\newblock \bibinfo{journal}{\emph{Electric Power Systems Research}}  \bibinfo{volume}{205} (\bibinfo{year}{2022}), \bibinfo{pages}{107761}.
\newblock


\bibitem[Tang et~al\mbox{.}(2019a)]%
        {tang2019short}
\bibfield{author}{\bibinfo{person}{Xianlun Tang}, \bibinfo{person}{Yuyan Dai}, \bibinfo{person}{Ting Wang}, {and} \bibinfo{person}{Yingjie Chen}.} \bibinfo{year}{2019}\natexlab{a}.
\newblock \showarticletitle{Short-term power load forecasting based on multi-layer bidirectional recurrent neural network}.
\newblock \bibinfo{journal}{\emph{IET Generation, Transmission \& Distribution}} \bibinfo{volume}{13}, \bibinfo{number}{17} (\bibinfo{year}{2019}), \bibinfo{pages}{3847--3854}.
\newblock


\bibitem[Tayab et~al\mbox{.}(2020)]%
        {tayab2020short}
\bibfield{author}{\bibinfo{person}{Usman~Bashir Tayab}, \bibinfo{person}{Ali Zia}, \bibinfo{person}{Fuwen Yang}, \bibinfo{person}{Junwei Lu}, {and} \bibinfo{person}{Muhammad Kashif}.} \bibinfo{year}{2020}\natexlab{}.
\newblock \showarticletitle{Short-term load forecasting for microgrid energy management system using hybrid HHO-FNN model with best-basis stationary wavelet packet transform}.
\newblock \bibinfo{journal}{\emph{Energy}}  \bibinfo{volume}{203} (\bibinfo{year}{2020}), \bibinfo{pages}{117857}.
\newblock


\bibitem[Van Den~Oord et~al\mbox{.}(2016)]%
        {van2016wavenet}
\bibfield{author}{\bibinfo{person}{Aaron Van Den~Oord}, \bibinfo{person}{Sander Dieleman}, \bibinfo{person}{Heiga Zen}, \bibinfo{person}{Karen Simonyan}, \bibinfo{person}{Oriol Vinyals}, \bibinfo{person}{Alex Graves}, \bibinfo{person}{Nal Kalchbrenner}, \bibinfo{person}{Andrew Senior}, \bibinfo{person}{Koray Kavukcuoglu}, {et~al\mbox{.}}} \bibinfo{year}{2016}\natexlab{}.
\newblock \showarticletitle{Wavenet: {A} generative model for raw audio}.
\newblock \bibinfo{journal}{\emph{arXiv preprint arXiv:1609.03499}}  \bibinfo{volume}{12} (\bibinfo{year}{2016}).
\newblock


\bibitem[Van~der Meer et~al\mbox{.}(2018)]%
        {van2018review}
\bibfield{author}{\bibinfo{person}{Dennis~W Van~der Meer}, \bibinfo{person}{Joakim Wid{\'e}n}, {and} \bibinfo{person}{Joakim Munkhammar}.} \bibinfo{year}{2018}\natexlab{}.
\newblock \showarticletitle{Review on probabilistic forecasting of photovoltaic power production and electricity consumption}.
\newblock \bibinfo{journal}{\emph{Renewable and Sustainable Energy Reviews}}  \bibinfo{volume}{81} (\bibinfo{year}{2018}), \bibinfo{pages}{1484--1512}.
\newblock


\bibitem[Vaswani et~al\mbox{.}(2017)]%
        {vaswani2017attention}
\bibfield{author}{\bibinfo{person}{Ashish Vaswani}, \bibinfo{person}{Noam Shazeer}, \bibinfo{person}{Niki Parmar}, \bibinfo{person}{Jakob Uszkoreit}, \bibinfo{person}{Llion Jones}, \bibinfo{person}{Aidan~N Gomez}, \bibinfo{person}{{\L}ukasz Kaiser}, {and} \bibinfo{person}{Illia Polosukhin}.} \bibinfo{year}{2017}\natexlab{}.
\newblock \showarticletitle{Attention is all you need}.
\newblock \bibinfo{journal}{\emph{Advances in Neural Information Processing Systems}}  \bibinfo{volume}{30} (\bibinfo{year}{2017}).
\newblock


\bibitem[Veeramsetty et~al\mbox{.}(2022)]%
        {veeramsetty2022short}
\bibfield{author}{\bibinfo{person}{Venkataramana Veeramsetty}, \bibinfo{person}{Dongari~Rakesh Chandra}, \bibinfo{person}{Francesco Grimaccia}, {and} \bibinfo{person}{Marco Mussetta}.} \bibinfo{year}{2022}\natexlab{}.
\newblock \showarticletitle{Short term electric power load forecasting using principal component analysis and recurrent neural networks}.
\newblock \bibinfo{journal}{\emph{Forecasting}} \bibinfo{volume}{4}, \bibinfo{number}{1} (\bibinfo{year}{2022}), \bibinfo{pages}{149--164}.
\newblock


\bibitem[Vo{\ss} et~al\mbox{.}(2018)]%
        {voss2018residential}
\bibfield{author}{\bibinfo{person}{Marcus Vo{\ss}}, \bibinfo{person}{Christian Bender-Saebelkampf}, {and} \bibinfo{person}{Sahin Albayrak}.} \bibinfo{year}{2018}\natexlab{}.
\newblock \showarticletitle{Residential short-term load forecasting using convolutional neural networks}. In \bibinfo{booktitle}{\emph{Proceedings of the 2018 IEEE International Conference on Communications, Control, and Computing Technologies for Smart Grids (SmartGridComm'18)}}. \bibinfo{pages}{1--6}.
\newblock


\bibitem[Wan et~al\mbox{.}(2023)]%
        {wan2023short}
\bibfield{author}{\bibinfo{person}{Anping Wan}, \bibinfo{person}{Qing Chang}, \bibinfo{person}{AL-Bukhaiti Khalil}, {and} \bibinfo{person}{Jiabo He}.} \bibinfo{year}{2023}\natexlab{}.
\newblock \showarticletitle{Short-term power load forecasting for combined heat and power using CNN-LSTM enhanced by attention mechanism}.
\newblock \bibinfo{journal}{\emph{Energy}}  \bibinfo{volume}{282} (\bibinfo{year}{2023}), \bibinfo{pages}{128274}.
\newblock


\bibitem[Wang et~al\mbox{.}(2019b)]%
        {wang2019review}
\bibfield{author}{\bibinfo{person}{Huaizhi Wang}, \bibinfo{person}{Zhenxing Lei}, \bibinfo{person}{Xian Zhang}, \bibinfo{person}{Bin Zhou}, {and} \bibinfo{person}{Jianchun Peng}.} \bibinfo{year}{2019}\natexlab{b}.
\newblock \showarticletitle{A review of deep learning for renewable energy forecasting}.
\newblock \bibinfo{journal}{\emph{Energy Conversion and Management}}  \bibinfo{volume}{198} (\bibinfo{year}{2019}), \bibinfo{pages}{111799}.
\newblock


\bibitem[Wang et~al\mbox{.}(2023a)]%
        {wang2023electrical}
\bibfield{author}{\bibinfo{person}{Jianguo Wang}, \bibinfo{person}{Lincheng Han}, \bibinfo{person}{Xiuyu Zhang}, \bibinfo{person}{Yingzhou Wang}, {and} \bibinfo{person}{Shude Zhang}.} \bibinfo{year}{2023}\natexlab{a}.
\newblock \showarticletitle{Electrical load forecasting based on variable {T}-distribution and dual attention mechanism}.
\newblock \bibinfo{journal}{\emph{Energy}}  \bibinfo{volume}{283} (\bibinfo{year}{2023}), \bibinfo{pages}{128569}.
\newblock


\bibitem[Wang et~al\mbox{.}(2023c)]%
        {wang2023shortc}
\bibfield{author}{\bibinfo{person}{Jianzhou Wang}, \bibinfo{person}{Kang Wang}, \bibinfo{person}{Zhiwu Li}, \bibinfo{person}{Haiyan Lu}, {and} \bibinfo{person}{He Jiang}.} \bibinfo{year}{2023}\natexlab{c}.
\newblock \showarticletitle{Short-term power load forecasting system based on rough set, information granule and multi-objective optimization}.
\newblock \bibinfo{journal}{\emph{Applied Soft Computing}}  \bibinfo{volume}{146} (\bibinfo{year}{2023}), \bibinfo{pages}{110692}.
\newblock


\bibitem[Wang et~al\mbox{.}(2020b)]%
        {wang2020ensemble}
\bibfield{author}{\bibinfo{person}{Lingxiao Wang}, \bibinfo{person}{Shiwen Mao}, \bibinfo{person}{Bogdan~M Wilamowski}, {and} \bibinfo{person}{RM Nelms}.} \bibinfo{year}{2020}\natexlab{b}.
\newblock \showarticletitle{Ensemble learning for load forecasting}.
\newblock \bibinfo{journal}{\emph{IEEE Transactions on Green Communications and Networking}} \bibinfo{volume}{4}, \bibinfo{number}{2} (\bibinfo{year}{2020}), \bibinfo{pages}{616--628}.
\newblock


\bibitem[Wang et~al\mbox{.}(2023d)]%
        {wang2023short}
\bibfield{author}{\bibinfo{person}{Lingyun Wang}, \bibinfo{person}{Xiang Zhou}, \bibinfo{person}{Honglei Xu}, \bibinfo{person}{Tian Tian}, {and} \bibinfo{person}{Huamin Tong}.} \bibinfo{year}{2023}\natexlab{d}.
\newblock \showarticletitle{Short-term electrical load forecasting model based on multi-dimensional meteorological information spatio-temporal fusion and optimized variational mode decomposition}.
\newblock \bibinfo{journal}{\emph{IET Generation, Transmission \& Distribution}} \bibinfo{volume}{17}, \bibinfo{number}{20} (\bibinfo{year}{2023}), \bibinfo{pages}{4647--4663}.
\newblock


\bibitem[Wang et~al\mbox{.}(2021)]%
        {wang2021bottom}
\bibfield{author}{\bibinfo{person}{Shouxiang Wang}, \bibinfo{person}{Xinyu Deng}, \bibinfo{person}{Haiwen Chen}, \bibinfo{person}{Qingyuan Shi}, {and} \bibinfo{person}{Di Xu}.} \bibinfo{year}{2021}\natexlab{}.
\newblock \showarticletitle{A bottom-up short-term residential load forecasting approach based on appliance characteristic analysis and multi-task learning}.
\newblock \bibinfo{journal}{\emph{Electric Power Systems Research}}  \bibinfo{volume}{196} (\bibinfo{year}{2021}), \bibinfo{pages}{107233}.
\newblock


\bibitem[Wang et~al\mbox{.}(2019c)]%
        {wang2019bi}
\bibfield{author}{\bibinfo{person}{Shouxiang Wang}, \bibinfo{person}{Xuan Wang}, \bibinfo{person}{Shaomin Wang}, {and} \bibinfo{person}{Dan Wang}.} \bibinfo{year}{2019}\natexlab{c}.
\newblock \showarticletitle{Bi-directional long short-term memory method based on attention mechanism and rolling update for short-term load forecasting}.
\newblock \bibinfo{journal}{\emph{International Journal of Electrical Power \& Energy Systems}}  \bibinfo{volume}{109} (\bibinfo{year}{2019}), \bibinfo{pages}{470--479}.
\newblock


\bibitem[Wang et~al\mbox{.}(2020a)]%
        {wang2020short}
\bibfield{author}{\bibinfo{person}{Yuanyuan Wang}, \bibinfo{person}{Jun Chen}, \bibinfo{person}{Xiaoqiao Chen}, \bibinfo{person}{Xiangjun Zeng}, \bibinfo{person}{Yang Kong}, \bibinfo{person}{Shanfeng Sun}, \bibinfo{person}{Yongsheng Guo}, {and} \bibinfo{person}{Ying Liu}.} \bibinfo{year}{2020}\natexlab{a}.
\newblock \showarticletitle{Short-term load forecasting for industrial customers based on {TCN-LightGBM}}.
\newblock \bibinfo{journal}{\emph{IEEE Transactions on Power Systems}} \bibinfo{volume}{36}, \bibinfo{number}{3} (\bibinfo{year}{2020}), \bibinfo{pages}{1984--1997}.
\newblock


\bibitem[Wang et~al\mbox{.}(2019a)]%
        {wang2019probabilistic}
\bibfield{author}{\bibinfo{person}{Yi Wang}, \bibinfo{person}{Dahua Gan}, \bibinfo{person}{Mingyang Sun}, \bibinfo{person}{Ning Zhang}, \bibinfo{person}{Zongxiang Lu}, {and} \bibinfo{person}{Chongqing Kang}.} \bibinfo{year}{2019}\natexlab{a}.
\newblock \showarticletitle{Probabilistic individual load forecasting using pinball loss guided {LSTM}}.
\newblock \bibinfo{journal}{\emph{Applied Energy}}  \bibinfo{volume}{235} (\bibinfo{year}{2019}), \bibinfo{pages}{10--20}.
\newblock


\bibitem[Wang et~al\mbox{.}(2023b)]%
        {wang2023shortb}
\bibfield{author}{\bibinfo{person}{Yufeng Wang}, \bibinfo{person}{Lingxiao Rui}, \bibinfo{person}{Jianhua Ma}, {et~al\mbox{.}}} \bibinfo{year}{2023}\natexlab{b}.
\newblock \showarticletitle{A short-term residential load forecasting scheme based on the multiple correlation-temporal graph neural networks}.
\newblock \bibinfo{journal}{\emph{Applied Soft Computing}}  \bibinfo{volume}{146} (\bibinfo{year}{2023}), \bibinfo{pages}{110629}.
\newblock


\bibitem[Wu et~al\mbox{.}(2023a)]%
        {wu2023pulse}
\bibfield{author}{\bibinfo{person}{Han Wu}, \bibinfo{person}{Yan Liang}, {and} \bibinfo{person}{Jiani Heng}.} \bibinfo{year}{2023}\natexlab{a}.
\newblock \showarticletitle{Pulse-diagnosis-inspired multi-feature extraction deep network for short-term electricity load forecasting}.
\newblock \bibinfo{journal}{\emph{Applied Energy}}  \bibinfo{volume}{339} (\bibinfo{year}{2023}), \bibinfo{pages}{120995}.
\newblock


\bibitem[Wu et~al\mbox{.}(2023b)]%
        {wu2023novel}
\bibfield{author}{\bibinfo{person}{Kaitong Wu}, \bibinfo{person}{Xiangang Peng}, \bibinfo{person}{Zhiwen Chen}, \bibinfo{person}{Haokun Su}, \bibinfo{person}{Huan Quan}, {and} \bibinfo{person}{Hanyu Liu}.} \bibinfo{year}{2023}\natexlab{b}.
\newblock \showarticletitle{A novel short-term household load forecasting method combined {BiLSTM} with trend feature extraction}.
\newblock \bibinfo{journal}{\emph{Energy Reports}}  \bibinfo{volume}{9} (\bibinfo{year}{2023}), \bibinfo{pages}{1013--1022}.
\newblock


\bibitem[Wu et~al\mbox{.}(2021)]%
        {wu2021online}
\bibfield{author}{\bibinfo{person}{Xuedong Wu}, \bibinfo{person}{Yaonan Wang}, \bibinfo{person}{Yingjie Bai}, \bibinfo{person}{Zhiyu Zhu}, {and} \bibinfo{person}{Aiming Xia}.} \bibinfo{year}{2021}\natexlab{}.
\newblock \showarticletitle{Online short-term load forecasting methods using hybrids of single multiplicative neuron model, particle swarm optimization variants and nonlinear filters}.
\newblock \bibinfo{journal}{\emph{Energy Reports}}  \bibinfo{volume}{7} (\bibinfo{year}{2021}), \bibinfo{pages}{683--692}.
\newblock


\bibitem[Wu et~al\mbox{.}(2022)]%
        {wu2022spatial}
\bibfield{author}{\bibinfo{person}{Zeqing Wu}, \bibinfo{person}{Yunfei Mu}, \bibinfo{person}{Shuai Deng}, {and} \bibinfo{person}{Yang Li}.} \bibinfo{year}{2022}\natexlab{}.
\newblock \showarticletitle{Spatial--temporal short-term load forecasting framework via {K}-shape time series clustering method and graph convolutional networks}.
\newblock \bibinfo{journal}{\emph{Energy Reports}}  \bibinfo{volume}{8} (\bibinfo{year}{2022}), \bibinfo{pages}{8752--8766}.
\newblock


\bibitem[Xia et~al\mbox{.}(2023)]%
        {xia2023combined}
\bibfield{author}{\bibinfo{person}{Yurui Xia}, \bibinfo{person}{Jianzhou Wang}, \bibinfo{person}{Danxiang Wei}, {and} \bibinfo{person}{Ziyuan Zhang}.} \bibinfo{year}{2023}\natexlab{}.
\newblock \showarticletitle{Combined framework based on data preprocessing and multi-objective optimizer for electricity load forecasting}.
\newblock \bibinfo{journal}{\emph{Engineering Applications of Artificial Intelligence}}  \bibinfo{volume}{119} (\bibinfo{year}{2023}), \bibinfo{pages}{105776}.
\newblock


\bibitem[Xiang et~al\mbox{.}(2023)]%
        {xiang2023power}
\bibfield{author}{\bibinfo{person}{Siyu Xiang}, \bibinfo{person}{Cao Zhen}, \bibinfo{person}{Jian Peng}, \bibinfo{person}{Linghao Zhang}, {and} \bibinfo{person}{Zhengguo Pu}.} \bibinfo{year}{2023}\natexlab{}.
\newblock \showarticletitle{Power load prediction of smart grid based on deep learning}.
\newblock \bibinfo{journal}{\emph{Procedia Computer Science}}  \bibinfo{volume}{228} (\bibinfo{year}{2023}), \bibinfo{pages}{762--773}.
\newblock


\bibitem[Xie et~al\mbox{.}(2022)]%
        {xie2022multi}
\bibfield{author}{\bibinfo{person}{Jiangjian Xie}, \bibinfo{person}{Yujie Zhong}, \bibinfo{person}{Tong Xiao}, \bibinfo{person}{Zheng Wang}, \bibinfo{person}{Junguo Zhang}, \bibinfo{person}{Tuowai Wang}, {and} \bibinfo{person}{Bj{\"o}rn~W Schuller}.} \bibinfo{year}{2022}\natexlab{}.
\newblock \showarticletitle{A multi-information fusion model for short term load forecasting of an architectural complex considering spatio-temporal characteristics}.
\newblock \bibinfo{journal}{\emph{Energy and Buildings}}  \bibinfo{volume}{277} (\bibinfo{year}{2022}), \bibinfo{pages}{112566}.
\newblock


\bibitem[Xu et~al\mbox{.}(2022)]%
        {xu2022probabilistic}
\bibfield{author}{\bibinfo{person}{Lei Xu}, \bibinfo{person}{Maomao Hu}, {and} \bibinfo{person}{Cheng Fan}.} \bibinfo{year}{2022}\natexlab{}.
\newblock \showarticletitle{Probabilistic electrical load forecasting for buildings using Bayesian deep neural networks}.
\newblock \bibinfo{journal}{\emph{Journal of Building Engineering}}  \bibinfo{volume}{46} (\bibinfo{year}{2022}), \bibinfo{pages}{103853}.
\newblock


\bibitem[Xu et~al\mbox{.}(2018)]%
        {xu2018long}
\bibfield{author}{\bibinfo{person}{Liwen Xu}, \bibinfo{person}{Chengdong Li}, \bibinfo{person}{Xiuying Xie}, {and} \bibinfo{person}{Guiqing Zhang}.} \bibinfo{year}{2018}\natexlab{}.
\newblock \showarticletitle{Long-short-term memory network based hybrid model for short-term electrical load forecasting}.
\newblock \bibinfo{journal}{\emph{Information}} \bibinfo{volume}{9}, \bibinfo{number}{7} (\bibinfo{year}{2018}), \bibinfo{pages}{165}.
\newblock


\bibitem[Yang et~al\mbox{.}(2023)]%
        {yang2023iterative}
\bibfield{author}{\bibinfo{person}{Bo Yang}, \bibinfo{person}{Xiaohui Yuan}, {and} \bibinfo{person}{Fei Tang}.} \bibinfo{year}{2023}\natexlab{}.
\newblock \showarticletitle{Iterative memory-driven load forecast network model for accuracy improvement}.
\newblock \bibinfo{journal}{\emph{Energy Reports}}  \bibinfo{volume}{9} (\bibinfo{year}{2023}), \bibinfo{pages}{388--395}.
\newblock


\bibitem[Yang et~al\mbox{.}(2022)]%
        {yang2022combined}
\bibfield{author}{\bibinfo{person}{Wangwang Yang}, \bibinfo{person}{Jing Shi}, \bibinfo{person}{Shujian Li}, \bibinfo{person}{Zhaofang Song}, \bibinfo{person}{Zitong Zhang}, {and} \bibinfo{person}{Zexu Chen}.} \bibinfo{year}{2022}\natexlab{}.
\newblock \showarticletitle{A combined deep learning load forecasting model of single household resident user considering multi-time scale electricity consumption behavior}.
\newblock \bibinfo{journal}{\emph{Applied Energy}}  \bibinfo{volume}{307} (\bibinfo{year}{2022}), \bibinfo{pages}{118197}.
\newblock


\bibitem[Yang et~al\mbox{.}(2019)]%
        {yang2019deep}
\bibfield{author}{\bibinfo{person}{Yandong Yang}, \bibinfo{person}{Weijun Hong}, {and} \bibinfo{person}{Shufang Li}.} \bibinfo{year}{2019}\natexlab{}.
\newblock \showarticletitle{Deep ensemble learning based probabilistic load forecasting in smart grids}.
\newblock \bibinfo{journal}{\emph{Energy}}  \bibinfo{volume}{189} (\bibinfo{year}{2019}), \bibinfo{pages}{116324}.
\newblock


\bibitem[Yang et~al\mbox{.}(2018)]%
        {yang2018power}
\bibfield{author}{\bibinfo{person}{Yandong Yang}, \bibinfo{person}{Shufang Li}, \bibinfo{person}{Wenqi Li}, {and} \bibinfo{person}{Meijun Qu}.} \bibinfo{year}{2018}\natexlab{}.
\newblock \showarticletitle{Power load probability density forecasting using gaussian process quantile regression}.
\newblock \bibinfo{journal}{\emph{Applied Energy}}  \bibinfo{volume}{213} (\bibinfo{year}{2018}), \bibinfo{pages}{499--509}.
\newblock


\bibitem[Yazici et~al\mbox{.}(2022)]%
        {yazici2022deep}
\bibfield{author}{\bibinfo{person}{Ibrahim Yazici}, \bibinfo{person}{Omer~Faruk Beyca}, {and} \bibinfo{person}{Dursun Delen}.} \bibinfo{year}{2022}\natexlab{}.
\newblock \showarticletitle{Deep-learning-based short-term electricity load forecasting: {A} real case application}.
\newblock \bibinfo{journal}{\emph{Engineering Applications of Artificial Intelligence}}  \bibinfo{volume}{109} (\bibinfo{year}{2022}), \bibinfo{pages}{104645}.
\newblock


\bibitem[Yi et~al\mbox{.}(2023)]%
        {yi2023deep}
\bibfield{author}{\bibinfo{person}{Shiyan Yi}, \bibinfo{person}{Haichun Liu}, \bibinfo{person}{Tao Chen}, \bibinfo{person}{Jianwen Zhang}, {and} \bibinfo{person}{Yibo Fan}.} \bibinfo{year}{2023}\natexlab{}.
\newblock \showarticletitle{A deep {LSTM-CNN} based on self-attention mechanism with input data reduction for short-term load forecasting}.
\newblock \bibinfo{journal}{\emph{IET Generation, Transmission \& Distribution}} \bibinfo{volume}{17}, \bibinfo{number}{7} (\bibinfo{year}{2023}), \bibinfo{pages}{1538--1552}.
\newblock


\bibitem[Yin and Xie(2021)]%
        {yin2021multi}
\bibfield{author}{\bibinfo{person}{Linfei Yin} {and} \bibinfo{person}{Jiaxing Xie}.} \bibinfo{year}{2021}\natexlab{}.
\newblock \showarticletitle{Multi-temporal-spatial-scale temporal convolution network for short-term load forecasting of power systems}.
\newblock \bibinfo{journal}{\emph{Applied Energy}}  \bibinfo{volume}{283} (\bibinfo{year}{2021}), \bibinfo{pages}{116328}.
\newblock


\bibitem[Yu et~al\mbox{.}(2022)]%
        {yu2022self}
\bibfield{author}{\bibinfo{person}{Fan Yu}, \bibinfo{person}{Lei Wang}, \bibinfo{person}{Qiaoyong Jiang}, \bibinfo{person}{Qunmin Yan}, {and} \bibinfo{person}{Shi Qiao}.} \bibinfo{year}{2022}\natexlab{}.
\newblock \showarticletitle{Self-attention-based short-term load forecasting considering demand-side management}.
\newblock \bibinfo{journal}{\emph{Energies}} \bibinfo{volume}{15}, \bibinfo{number}{12} (\bibinfo{year}{2022}), \bibinfo{pages}{4198}.
\newblock


\bibitem[Yu and Li(2021)]%
        {yu2021correlated}
\bibfield{author}{\bibinfo{person}{Qun Yu} {and} \bibinfo{person}{Zhiyi Li}.} \bibinfo{year}{2021}\natexlab{}.
\newblock \showarticletitle{Correlated load forecasting in active distribution networks using spatial-temporal synchronous graph convolutional networks}.
\newblock \bibinfo{journal}{\emph{IET Energy Systems Integration}} \bibinfo{volume}{3}, \bibinfo{number}{3} (\bibinfo{year}{2021}), \bibinfo{pages}{355--366}.
\newblock


\bibitem[Yu et~al\mbox{.}(2023)]%
        {yu2023temporal}
\bibfield{author}{\bibinfo{person}{Xinli Yu}, \bibinfo{person}{Zheng Chen}, \bibinfo{person}{Yuan Ling}, \bibinfo{person}{Shujing Dong}, \bibinfo{person}{Zongyi Liu}, {and} \bibinfo{person}{Yanbin Lu}.} \bibinfo{year}{2023}\natexlab{}.
\newblock \showarticletitle{Temporal data meets {LLM}--explainable financial time series forecasting}.
\newblock \bibinfo{journal}{\emph{arXiv preprint arXiv:2306.11025}} (\bibinfo{year}{2023}).
\newblock


\bibitem[Yue et~al\mbox{.}(2022)]%
        {yue2022prediction}
\bibfield{author}{\bibinfo{person}{Weimin Yue}, \bibinfo{person}{Qingrong Liu}, \bibinfo{person}{Yingjun Ruan}, \bibinfo{person}{Fanyue Qian}, {and} \bibinfo{person}{Hua Meng}.} \bibinfo{year}{2022}\natexlab{}.
\newblock \showarticletitle{A prediction approach with mode decomposition-recombination technique for short-term load forecasting}.
\newblock \bibinfo{journal}{\emph{Sustainable Cities and Society}}  \bibinfo{volume}{85} (\bibinfo{year}{2022}), \bibinfo{pages}{104034}.
\newblock


\bibitem[Zamee et~al\mbox{.}(2021)]%
        {zamee2021online}
\bibfield{author}{\bibinfo{person}{Muhammad~Ahsan Zamee}, \bibinfo{person}{Dongjun Han}, {and} \bibinfo{person}{Dongjun Won}.} \bibinfo{year}{2021}\natexlab{}.
\newblock \showarticletitle{Online hour-ahead load forecasting using appropriate time-delay neural network based on multiple correlation--multicollinearity analysis in IoT energy network}.
\newblock \bibinfo{journal}{\emph{IEEE Internet of Things Journal}} \bibinfo{volume}{9}, \bibinfo{number}{14} (\bibinfo{year}{2021}), \bibinfo{pages}{12041--12055}.
\newblock


\bibitem[Zang et~al\mbox{.}(2021)]%
        {zang2021residential}
\bibfield{author}{\bibinfo{person}{Haixiang Zang}, \bibinfo{person}{Ruiqi Xu}, \bibinfo{person}{Lilin Cheng}, \bibinfo{person}{Tao Ding}, \bibinfo{person}{Ling Liu}, \bibinfo{person}{Zhinong Wei}, {and} \bibinfo{person}{Guoqiang Sun}.} \bibinfo{year}{2021}\natexlab{}.
\newblock \showarticletitle{Residential load forecasting based on {LSTM} fusing self-attention mechanism with pooling}.
\newblock \bibinfo{journal}{\emph{Energy}}  \bibinfo{volume}{229} (\bibinfo{year}{2021}), \bibinfo{pages}{120682}.
\newblock


\bibitem[Zhang et~al\mbox{.}(2023e)]%
        {zhang2023novel}
\bibfield{author}{\bibinfo{person}{Dongxue Zhang}, \bibinfo{person}{Shuai Wang}, \bibinfo{person}{Yuqiu Liang}, {and} \bibinfo{person}{Zhiyuan Du}.} \bibinfo{year}{2023}\natexlab{e}.
\newblock \showarticletitle{A novel combined model for probabilistic load forecasting based on deep learning and improved optimizer}.
\newblock \bibinfo{journal}{\emph{Energy}}  \bibinfo{volume}{264} (\bibinfo{year}{2023}), \bibinfo{pages}{126172}.
\newblock


\bibitem[Zhang et~al\mbox{.}(2022b)]%
        {zhang2022short}
\bibfield{author}{\bibinfo{person}{Guangqi Zhang}, \bibinfo{person}{Chuyuan Wei}, \bibinfo{person}{Changfeng Jing}, {and} \bibinfo{person}{Yanxue Wang}.} \bibinfo{year}{2022}\natexlab{b}.
\newblock \showarticletitle{Short-term electrical load forecasting based on time augmented transformer}.
\newblock \bibinfo{journal}{\emph{International Journal of Computational Intelligence Systems}} \bibinfo{volume}{15}, \bibinfo{number}{1} (\bibinfo{year}{2022}), \bibinfo{pages}{67:1--67:11}.
\newblock


\bibitem[Zhang et~al\mbox{.}(2022a)]%
        {zhang2022electricity}
\bibfield{author}{\bibinfo{person}{Han Zhang}, \bibinfo{person}{Chen Peng}, \bibinfo{person}{Jun Li}, \bibinfo{person}{Yajie Niu}, {and} \bibinfo{person}{Longxiang Li}.} \bibinfo{year}{2022}\natexlab{a}.
\newblock \showarticletitle{Electricity load forecasting based on an interpretable probsparse attention mechanism}. In \bibinfo{booktitle}{\emph{Proceedings of the 3rd International Conference on Artificial Intelligence, Information Processing and Cloud Computing (AIIPCC'22)}}. \bibinfo{pages}{1--7}.
\newblock


\bibitem[Zhang et~al\mbox{.}(2023d)]%
        {zhang2023improved}
\bibfield{author}{\bibinfo{person}{Jinliang Zhang}, \bibinfo{person}{Wang Siya}, \bibinfo{person}{Tan Zhongfu}, {and} \bibinfo{person}{Sun Anli}.} \bibinfo{year}{2023}\natexlab{d}.
\newblock \showarticletitle{An improved hybrid model for short term power load prediction}.
\newblock \bibinfo{journal}{\emph{Energy}}  \bibinfo{volume}{268} (\bibinfo{year}{2023}), \bibinfo{pages}{126561}.
\newblock


\bibitem[Zhang et~al\mbox{.}(2021)]%
        {zhang2021review}
\bibfield{author}{\bibinfo{person}{Liang Zhang}, \bibinfo{person}{Jin Wen}, \bibinfo{person}{Yanfei Li}, \bibinfo{person}{Jianli Chen}, \bibinfo{person}{Yunyang Ye}, \bibinfo{person}{Yangyang Fu}, {and} \bibinfo{person}{William Livingood}.} \bibinfo{year}{2021}\natexlab{}.
\newblock \showarticletitle{A review of machine learning in building load prediction}.
\newblock \bibinfo{journal}{\emph{Applied Energy}}  \bibinfo{volume}{285} (\bibinfo{year}{2021}), \bibinfo{pages}{116452}.
\newblock


\bibitem[Zhang et~al\mbox{.}(2023b)]%
        {zhang2023transformgraph}
\bibfield{author}{\bibinfo{person}{Qingyong Zhang}, \bibinfo{person}{Jiahua Chen}, \bibinfo{person}{Gang Xiao}, \bibinfo{person}{Shangyang He}, {and} \bibinfo{person}{Kunxiang Deng}.} \bibinfo{year}{2023}\natexlab{b}.
\newblock \showarticletitle{TransformGraph: A novel short-term electricity net load forecasting model}.
\newblock \bibinfo{journal}{\emph{Energy Reports}}  \bibinfo{volume}{9} (\bibinfo{year}{2023}), \bibinfo{pages}{2705--2717}.
\newblock


\bibitem[Zhang et~al\mbox{.}(2022c)]%
        {zhang2022similar}
\bibfield{author}{\bibinfo{person}{Ruixuan Zhang}, \bibinfo{person}{Chuyan Zhang}, {and} \bibinfo{person}{Miao Yu}.} \bibinfo{year}{2022}\natexlab{c}.
\newblock \showarticletitle{A similar day based short term load forecasting method using wavelet transform and LSTM}.
\newblock \bibinfo{journal}{\emph{IEEJ Transactions on Electrical and Electronic Engineering}} \bibinfo{volume}{17}, \bibinfo{number}{4} (\bibinfo{year}{2022}), \bibinfo{pages}{506--513}.
\newblock


\bibitem[Zhang et~al\mbox{.}(2023g)]%
        {zhang2023regional}
\bibfield{author}{\bibinfo{person}{Ruixiang Zhang}, \bibinfo{person}{Ziyu Zhu}, \bibinfo{person}{Meng Yuan}, \bibinfo{person}{Yihan Guo}, \bibinfo{person}{Jie Song}, \bibinfo{person}{Xuanxuan Shi}, \bibinfo{person}{Yu Wang}, {and} \bibinfo{person}{Yaojie Sun}.} \bibinfo{year}{2023}\natexlab{g}.
\newblock \showarticletitle{Regional residential short-term load-interval forecasting based on {SSA-LSTM} and load consumption consistency analysis}.
\newblock \bibinfo{journal}{\emph{Energies}} \bibinfo{volume}{16}, \bibinfo{number}{24} (\bibinfo{year}{2023}), \bibinfo{pages}{8062}.
\newblock


\bibitem[Zhang et~al\mbox{.}(2023a)]%
        {zhang2023cnn}
\bibfield{author}{\bibinfo{person}{Shiyun Zhang}, \bibinfo{person}{Runhuan Chen}, \bibinfo{person}{Jiacheng Cao}, {and} \bibinfo{person}{Jian Tan}.} \bibinfo{year}{2023}\natexlab{a}.
\newblock \showarticletitle{A CNN and LSTM-based multi-task learning architecture for short and medium-term electricity load forecasting}.
\newblock \bibinfo{journal}{\emph{Electric power systems research}}  \bibinfo{volume}{222} (\bibinfo{year}{2023}), \bibinfo{pages}{109507}.
\newblock


\bibitem[Zhang et~al\mbox{.}(2023f)]%
        {zhang2023highly}
\bibfield{author}{\bibinfo{person}{Tingze Zhang}, \bibinfo{person}{Xinan Zhang}, \bibinfo{person}{Tat~Kei Chau}, \bibinfo{person}{Yau Chow}, \bibinfo{person}{Tyrone Fernando}, \bibinfo{person}{Herbert Ho-Ching Iu}, {et~al\mbox{.}}} \bibinfo{year}{2023}\natexlab{f}.
\newblock \showarticletitle{Highly accurate peak and valley prediction short-term net load forecasting approach based on decomposition for power systems with high PV penetration}.
\newblock \bibinfo{journal}{\emph{Applied Energy}}  \bibinfo{volume}{333} (\bibinfo{year}{2023}), \bibinfo{pages}{120641}.
\newblock


\bibitem[Zhang et~al\mbox{.}(2023c)]%
        {zhang2023general}
\bibfield{author}{\bibinfo{person}{Zhenhao Zhang}, \bibinfo{person}{Jiefeng Liu}, \bibinfo{person}{Senshen Pang}, \bibinfo{person}{Mingchen Shi}, \bibinfo{person}{Hui~Hwang Goh}, \bibinfo{person}{Yiyi Zhang}, {and} \bibinfo{person}{Dongdong Zhang}.} \bibinfo{year}{2023}\natexlab{c}.
\newblock \showarticletitle{General short-term load forecasting based on multi-task temporal convolutional network in {COVID}-19}.
\newblock \bibinfo{journal}{\emph{International Journal of Electrical Power \& Energy Systems}}  \bibinfo{volume}{147} (\bibinfo{year}{2023}), \bibinfo{pages}{108811}.
\newblock


\bibitem[Zheng et~al\mbox{.}(2018)]%
        {zheng2018short}
\bibfield{author}{\bibinfo{person}{Jiaxiang Zheng}, \bibinfo{person}{Xingying Chen}, \bibinfo{person}{Kun Yu}, \bibinfo{person}{Lei Gan}, \bibinfo{person}{Yifan Wang}, {and} \bibinfo{person}{Ke Wang}.} \bibinfo{year}{2018}\natexlab{}.
\newblock \showarticletitle{Short-term power load forecasting of residential community based on {GRU} neural network}. In \bibinfo{booktitle}{\emph{Proceedings of the 2018 International Conference on Power System Technology (POWERCON'18)}}. \bibinfo{pages}{4862--4868}.
\newblock


\bibitem[Zhu et~al\mbox{.}(2022)]%
        {zhu2022lstm}
\bibfield{author}{\bibinfo{person}{Kedong Zhu}, \bibinfo{person}{Yaping Li}, \bibinfo{person}{Wenbo Mao}, \bibinfo{person}{Feng Li}, {and} \bibinfo{person}{Jiahao Yan}.} \bibinfo{year}{2022}\natexlab{}.
\newblock \showarticletitle{{LSTM} enhanced by dual-attention-based encoder-decoder for daily peak load forecasting}.
\newblock \bibinfo{journal}{\emph{Electric Power Systems Research}}  \bibinfo{volume}{208} (\bibinfo{year}{2022}), \bibinfo{pages}{107860}.
\newblock


\bibitem[Zhuang et~al\mbox{.}(2022)]%
        {zhuang2022reliable}
\bibfield{author}{\bibinfo{person}{Zhiyuan Zhuang}, \bibinfo{person}{Xidong Zheng}, \bibinfo{person}{Zixing Chen}, {and} \bibinfo{person}{Tao Jin}.} \bibinfo{year}{2022}\natexlab{}.
\newblock \showarticletitle{A reliable short-term power load forecasting method based on VMD-IWOA-LSTM algorithm}.
\newblock \bibinfo{journal}{\emph{IEEJ Transactions on Electrical and Electronic Engineering}} \bibinfo{volume}{17}, \bibinfo{number}{8} (\bibinfo{year}{2022}), \bibinfo{pages}{1121--1132}.
\newblock


\end{thebibliography}

\end{document}